\documentclass{article} % For LaTeX2e
\usepackage{atom-jepa,times}

\usepackage{amsmath,amsfonts,bm}

\def\eqref#1{equation~\ref{#1}}
\def\1{\bm{1}}

\DeclareMathAlphabet{\mathsfit}{\encodingdefault}{\sfdefault}{m}{sl}
\SetMathAlphabet{\mathsfit}{bold}{\encodingdefault}{\sfdefault}{bx}{n}

\DeclareMathOperator*{\argmin}{arg\,min}

\usepackage{graphicx}
\usepackage{subcaption}
\usepackage[utf8]{inputenc} % allow utf-8 input
\usepackage[T1]{fontenc}    % use 8-bit T1 fonts
\usepackage{hyperref}       % hyperlinks
\usepackage{url}            % simple URL typesetting
\usepackage{booktabs}       % professional-quality tables
\usepackage{makecell}
\usepackage{amsfonts}       % blackboard math symbols
\usepackage{nicefrac}       % compact symbols for 1/2, etc.
\usepackage{microtype}      % microtypography
\usepackage[table]{xcolor}         % colors
\usepackage{color}
\usepackage{svg}
\usepackage{booktabs}
\usepackage{wrapfig}
\usepackage[dvipsnames]{xcolor}
\usepackage{multirow}
\usepackage{rotating}
\usepackage{array}
\usepackage{float}
\usepackage{adjustbox}
\usepackage{siunitx}
\usepackage{hyperref}
\PassOptionsToPackage{hyphens}{url}
\usepackage{xurl}
\usepackage{etoc}
\usepackage{float}
\usepackage[dvipsnames]{xcolor}
\usepackage[capitalize,compress]{cleveref}

\usepackage{xfp}
\newsavebox{\matbenchbox}

\usepackage[table]{xcolor}
\definecolor{atomjepablue}{RGB}{230,242,255}

\usepackage{xcolor}

\newcommand{\err}[1]{\textcolor{gray}{\scriptstyle\pm #1}}
\newcommand{\tablescale}{0.60}
\usepackage{wrapfig}

\hypersetup{
    colorlinks=true,
    citecolor=BlueViolet,  % Colors citations (e.g., [1]) blue
    linkcolor=BlueViolet,  % Colors internal links (sections, figures) blue
    urlcolor=BlueViolet    % Colors URLs blue
}

\title{Atom-JEPA: Joint-Embedding Predictive Architecture for 3D Atomistic Systems}

\author{Kasper Helverskov Petersen$^*$, Rasmus Hannibal Tirsgaard$^*$, François R J Cornet, \\
\textbf{Mikkel Jordahn$^*$, Mikkel N. Schmidt} \\
Department of Applied Mathematics and Computer Science \\
Technical University of Denmark \\
Corresponding authors: \texttt{\{khepe,rhti,mikkjo\}@dtu.dk}
}

\iclrfinalcopy
\begin{document}
\etocdepthtag.toc{main}

\maketitle

\begingroup
\renewcommand{\thefootnote}{\fnsymbol{footnote}}
\footnotetext[1]{Core contributors.}
\endgroup

\begin{abstract}
Large-scale self-supervised pretraining has reshaped modern machine learning, substantially advancing the ability of language and vision models to generalize across downstream tasks.
While deep learning has driven considerable progress in modeling atomistic systems in recent years, self-supervised pretraining in this domain has not yet achieved comparable downstream generalization.
To address this, we introduce Atom-JEPA, a self-supervised pretraining framework that learns latent representations from unlabeled 3D structures through complementary atom-level and substructure-level objectives inspired by joint-embedding predictive architectures.
We pretrain Atom-JEPA on large-scale molecular and crystalline datasets and evaluate its transfer performance by fine-tuning on a diverse set of downstream property prediction tasks.
Atom-JEPA achieves state-of-the-art performance on molecular ADMET and quantum-chemical property prediction tasks, and is highly competitive in predicting the physical properties of crystalline materials.
These results demonstrate the potential of latent-space predictive pretraining to support broad downstream generalization from structural data alone. Code and pretrained model checkpoints are publicly available at \url{https://github.com/khelverskovp/atom-jepa}.
\end{abstract}

\section{Introduction}
In recent years, machine learning has been adopted at an accelerating rate in molecular and materials discovery. Generative models enable inverse design through candidate generation \citep{hoogeboom2022equivariantdiffusionmoleculegeneration, cornet2024equivariant,mattergen2025zeni}, structure prediction \citep{accurate2024abramson, boltz2025wohlwend, chai2025zero} and property-guided optimization, complementing traditional workflows reliant on expert-guided design, computational screening, and experimental structure determination. Machine-learning interatomic-potentials (MLIPs)~\citep{schütt2021equivariantmessagepassingprediction, mace, rhodes2025orbv3atomisticsimulationscale, esen, wood2026umafamilyuniversalmodels, kavanagh2026fastaccurateequivariantfoundation} serve as surrogates for electronic structure methods such as DFT, while predictive models are increasingly used to screen promising candidates across a wide range of properties \citep{CCR5, Keap1-Nrf2, catalyst_ammonia,antibotics_deep_learning, catalyst_methane, catalyst_Na-S} before running costly computational or experimental evaluation. Historically, many of these approaches have relied on tailored models trained from scratch on individual datasets, often using molecular fingerprints \citep{rogers2010extended} or string-based representations \citep{weininger1988smiles}. Recently, increasing computational resources and growing availability of structural data has enabled the training of deep neural networks directly on 3D atomistic structures \citep{schutt2017schnet,gilmer2017neural}, offering the potential to learn representations that capture chemically relevant information directly from atomistic structure. However, a central question that remains unanswered is how to effectively leverage \textit{massive unlabelled} datasets to improve performance on \textit{sparsely labeled} downstream tasks, where less complex approaches such as simple fingerprint baselines often remain competitive. 

Foundation models, i.e. neural networks pretrained on large-scale datasets and subsequently adapted to downstream tasks, have been central to recent advances in deep learning. Many modern computer vision models \citep{siméoni2025dinov3}, large-language models \citep{chiang-etal-2022-recent}, and tabular data models \citep{Hollmann2022TabPFNAT, qu2025tabicl} increasingly rely on large-scale pretraining. In machine learning for atomistic systems, there is a similar growing interest in developing foundation models \citep{pyzer-knapp2025foundation, yuan2025foundationmodelsatomisticsimulation} that can exhibit strong generalization across different chemical systems and tasks. Existing approaches either pretrain via supervised regression on quantum-chemical properties such as energies, forces, stress, and electronic characteristics, or through self-supervised structural objectives such as denoising, masking, and environment prediction. Based on the success of self-supervised learning (SSL) in computer vision and language models, we posit that SSL is the way forward for strong foundation models for atomistic systems as well.

\begin{figure}[t]
\begin{center}
\includegraphics[width=\linewidth]{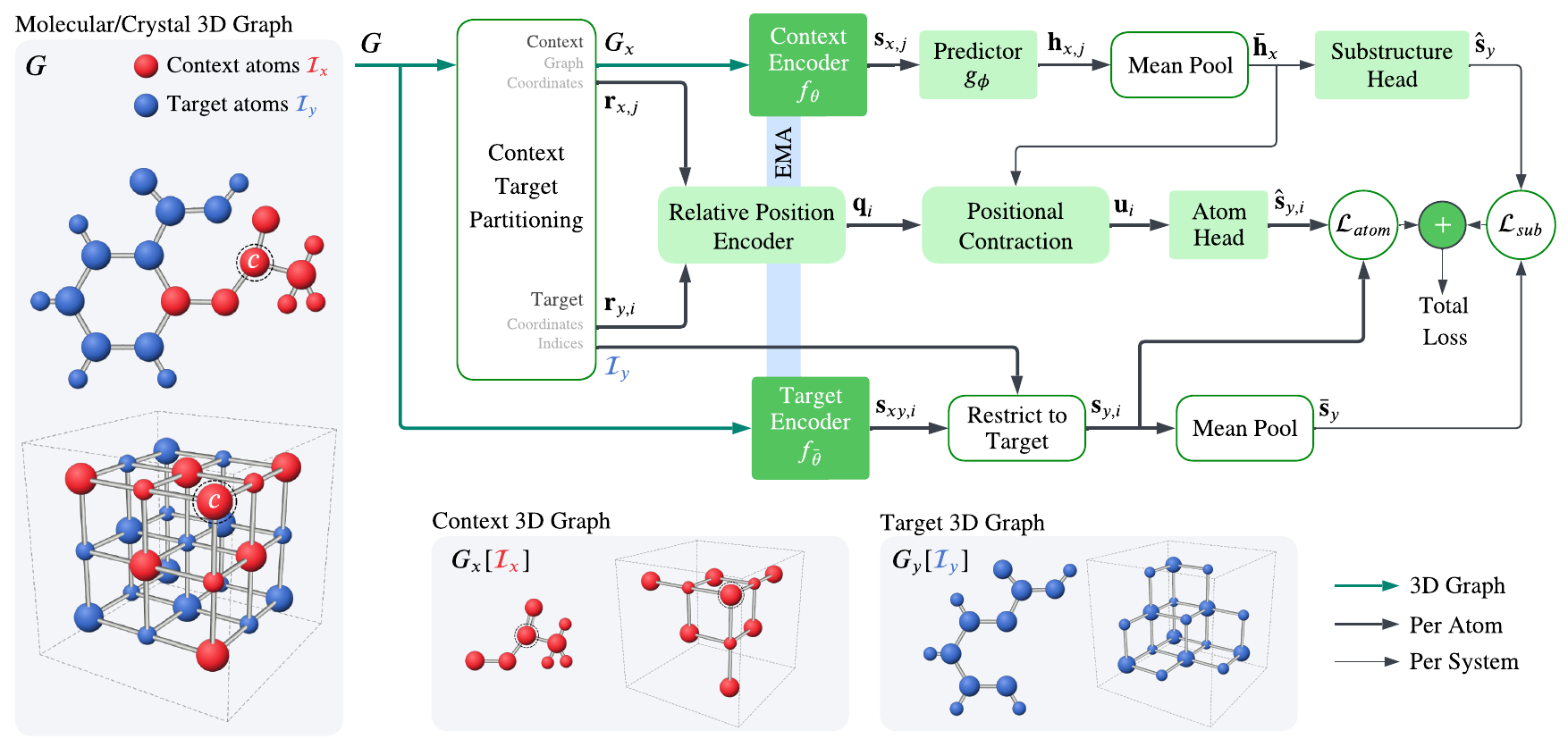}
\end{center}
\caption{\textbf{Overview of Atom-JEPA.} Molecular and crystalline structures are partitioned into complementary context (red) and target (blue) subgraphs by sampling a $k$-hop neighborhood around an anchor atom. A 3D equivariant graph neural network encodes the context, which is then used to predict target representations at two scales: individual atoms and mean-pooled substructures. Target representations are extracted from the full structure after using a target encoder updated as an exponential moving average of the context encoder. Pretraining minimizes prediction errors in latent space, with the two subgraphs alternating as context and target. The pretrained context encoder is subsequently fine-tuned for downstream property prediction.}
\label{fig:framework}
\end{figure}

In this work, we propose and demonstrate Atom-JEPA, a performant adaptation of the joint-embedding predictive architecture framework \citep{LeCun2022APT} to 3D atomistic systems. Specifically, Atom-JEPA combines complementary atom-level and substructure-level predictive objectives in latent space, using only 3D positions and atom types. In contrast, existing approaches based on 3D atomistic structure primarily formulate self-supervision through prediction or reconstruction in the input space, combining 3D information with other modalities, or employing latent-space predictive objectives only at the substructure level. For an overview of Atom-JEPA, we refer to \cref{fig:framework}. We train Atom-JEPA models on both molecular and crystalline structures, analyse their learned embedding spaces, and finally evaluate their performance on supervised downstream tasks spanning ADMET, quantum-chemical, and crystalline material properties. To summarise:

\begin{enumerate}
    \item We propose Atom-JEPA, to the best of our knowledge, the first JEPA-style loss that operates both on the atom- and substructure level of 3D atomistic systems alone.
    \item We show that Atom-JEPA outperforms most existing open-source models and methods across ADMET, quantum-chemical and crystalline material property predictions.
    \item We analyse the learned embeddings of Atom-JEPA trained models, illustrating where they do (and do not) separate atomistic systems by higher-order structural and functional groups.
\end{enumerate}

\section{Related Work}

\textbf{Supervised pretraining}~~~Supervised pretraining trains a neural network on a large \textit{labeled} dataset, before finetuning or adapting it to downstream tasks \citep{jmp,IvkoviÄ2024transfer,kong2025mattertune,kim2026large}. For atomistic systems, models are typically pretrained on energies and forces computed using density functional theory (DFT), and while these labels provide rich supervision about the underlying potential-energy surface, this pretraining objective may not capture all information relevant to diverse downstream properties. Another limitation is scalability: generating high-quality labels for molecules or crystals is computationally expensive. OMol25, for example, contains labels for 150 million configurations of atomistic systems at the $\omega$B97M-V/def2-TZVPD level of theory and required approximately 6.6 \textit{billion} CPU hours to generate \citep{levine2026openmolecules2025omol25}. In contrast, SSL provides a route to exploiting structural data without requiring expensive label simulations.

\textbf{Self-supervised pretraining}~~~Self-supervised learning as pretraining has also seen increasing attention in recent years. One of the earliest methods used VAEs, where the authors learn a latent representation of molecules by training a VAE on SMILES strings \citep{bombarelli2018automatic}. Since then, the development has been wide and varied. Several works have done SSL on 2D graphs, using both masking and contrastive learning losses \citep{hu2020strategiespretraininggraphneural, li2022kpgt,wang2023batmannet,xia2023molebert, DBLP:journals/corr/abs-2606-11508} and more recently, similar losses have been proposed for 3D molecular graphs, with the expectation that features incorporating geometry should allow better downstream performance \citep{zhu2022unified,zhou2023unimol,wu2025dgsrd}. Finally, a recent surge of papers has proposed using positional 3D denoising, sometimes in combination with masking losses, as a pretraining objective \citep{zaidi2023pretraining, pmlr-v202-feng23c, liu2025denoiseVAE, perez2026selfconditioneddenoisingatomisticrepresentation, morehead2026zatom1}.

\textbf{Joint-Embedding Predictive Architecture}~~~Joint-embedding predictive architectures (JEPA) \citep{LeCun2022APT} predict directly in \textit{latent space} rather than in the input domain, and rely on generating \textit{views} of data which teaches the model to focus on higher, more abstract level features of datapoints by using a predictor network. Graph-JEPA \citep{skenderi2025graphlevel} divides 2D graphs into sub-graphs, and uses these sub-graphs as context and target views, but without a node-level objective. Polymer-JEPA \citep{piccoli2026joint} extends the Graph-JEPA framework to 2D polymers by changing the target encoder to encode the full polymer, while only computing the loss over a target subgraph, by pooling the embeddings of the nodes in that subgraph that are not contained within the context subgraph, again with no node-level objective, and on 2D graphs. Crys-JEPA \citep{liu2026crysjepaacceleratingcrystaldiscovery} operates on sequences (created from the 3D crystal graph) using a Transformer and generates contexts by rotating and translating original target graphs, whilst using formation energy in the loss to produce energy-aware embeddings. C-Free \citep{cfree} is most similar to our method, but with significant differences: C-Free operates on 2D and 3D graphs jointly and generates contexts via k-hops along covalent bonds, where the targets are the complements to the context subgraph. In comparison, we sidestep the need for multiple modalities by introducing an atom-level JEPA objective, and generating contexts via k-hops on a cutoff radius based graph. Most recently, Mol-JEPA \citep{rottach2026moljepamultimodaljointembedding} is a framework in which representations from multiple pretrained models, computed descriptors, and experimental measurements define different molecular modalities, with masked modalities predicted in latent space from the remaining modalities. To summarise, our method is to the best of our knowledge, the first JEPA model for atomistic systems to learn atomic representations solely through latent-space prediction between 3D structural views.

\section{Method}
The core idea of Atom-JEPA is analogous to I-JEPA for images~\citep{i-jepa}, but tailored to atomistic systems in 3D space: given part of a molecule or crystal (context view), predict the representation of the remaining part (target view) using context and target encoders coupled through an exponential moving average (EMA). We consider two prediction objectives: (1) an atom-level objective that predicts the representation of each atom in the target view, and (2) a substructure-level objective that predicts the mean-pooled representation of all target atoms. For the context encoder, target encoder, and predictor, we use an SE(3)-equivariant graph neural network (GNN). 

We restrict our training data to structures that are close to local minima of the potential energy surface, as including off-equilibrium structures without information about their relative stability (e.g. total energy), could make the learning problem ambiguous. We discuss this choice further in \cref{app:data_curation}.  

\subsection{3D Graph Representation and Partitioning}
We represent an atomistic system with $N$ atoms by a 3D graph $G=(V,E)$. The node set $V=\{(z_i,\mathbf{r}_i)\}^N_{i=1}$ represents atoms by their species $z_i$ and coordinates $\mathbf{r}_i\in \mathbb{R}^3$, while the edge set $E=\{(i,j)\mid i\!\neq\!j, \|\mathbf{r}_i\!-\!\mathbf{r}_j\|\!<\!r_c\}$ encodes spatial proximity defined by a cutoff radius $r_c$. For crystalline systems, the edge set is constructed under periodic boundary conditions, including edges to periodic images within the cutoff radius.

Given an atomistic system, we partition atom indices into a $k$-hop EgoNet $\mathcal{I}_\mathrm{a}$ and its complement $\mathcal{I}_\mathrm{b}$: we  sample an anchor atom $c\sim\operatorname{Unif}\{1,\dots,N\}$ and a hop count $k\sim\operatorname{Unif}\mathcal{K}$ from a predefined set of hop counts $\mathcal{K}$, and define the EgoNet centered at $c$ and its complement as
\begin{equation}
%\mathcal{I}_x = \mathcal{N}_k(c)=\{i\in \{1,\dots,N\}\mid d(c,i)\leq k\},
\mathcal{I}_\mathrm{a} =\{i\in \{1,\dots,N\}\mid d(c,i)\leq k\},\quad \mathcal{I}_\mathrm{b} = \{1,\dots,N\}\setminus \mathcal{I}_\mathrm{a},
\end{equation}
where $d(c,i)$ denotes the shortest-path distance between atoms $c$ and $i$, measured in number of edges. 
The corresponding induced subgraphs are $G_\mathrm{a}=G[\mathcal{I}_\mathrm{a}]$ and $G_\mathrm{b}=G[\mathcal{I}_\mathrm{b}]$ where
\begin{equation}
G[\mathcal{I}] = (V_\mathcal{I}, E_\mathcal{I}), \quad V_\mathcal{I} = \{(z_i,\mathbf{r}_i)\mid i\in \mathcal{I}\}, \quad E_\mathcal{I} = \{(i,j)\in E\mid i,j\in \mathcal{I}\}.
\end{equation}
For each partition, we train in both prediction directions, defining the context and target graphs $(G_\mathrm{x}, G_\mathrm{y})$ as either $(G_\mathrm{a},G_\mathrm{b})$ or $(G_\mathrm{b},G_\mathrm{a})$.

\subsection{Context, Target and Predictor Models}
The context and target encoders, $f_\theta$ and $f_{\bar{\theta}}$ respectively, share the same 3D equivariant GNN architecture, taking 3D graphs as inputs and producing atom-level features. The parameters of the context encoder, $\theta$, are updated via gradients, whereas the target encoder parameters, $\bar{\theta}$, are an exponential moving average of $\theta$. The context encoder is only given context graph $G_x$, whilst the target encoder is given the full graph $G$; the JEPA loss is only computed over the target graph $G_y$. We label the context encoder output as $\mathbf{s}_x$ and the target encoder output on the $G_y$ graph as $\mathbf{s}_y$.

Given the context representations $\mathbf{s}_x$, the predictor, $g_\phi$, predicts representations produced by the target encoder. The predictor consists of a shallow 3D equivariant GNN backbone followed by objective-specific readout heads. 
We consider two prediction objectives: (1) a substructure objective, $\mathcal{L}_{\mathrm{sub}}$, that targets a global representation of the target subgraph, and (2) an atom-level objective, $\mathcal{L}_{\mathrm{atom}}$, that targets the representation of each target atom. We write the atom features produced by the predictor as $\mathbf{h}_x$ and the mean-pooled features as $\bar{\mathbf{h}}_x$.

\subsection{Prediction Objectives \& Atom-JEPA Loss}

\textbf{Substructure Objective}~~~In the substructure objective, the target is the mean-pooled invariant scalar features produced by the target encoder indexed over the target subgraph, whereas the prediction is constructed by passing the invariant component of the pooled context through a two-layer MLP mapping into the feature space of the target encoder,
\begin{align}
   \bar{\mathbf{s}}_\mathrm{y}=\frac{1}{|\mathcal{I}_\mathrm{y}|}\sum_{i\in \mathcal{I}_\mathrm{y}}\mathbf{s}_{\mathrm{y},i}^{(0)} \quad \text{and} \quad \hat{\bar{\mathbf{s}}}_{\mathrm{y}}=R^{\text{sub}}_\phi(\mathbf{\bar{h}}_\mathrm{x})=\text{MLP}(\bar{\mathbf{h}}_\mathrm{x}^{(0)}), 
\end{align}
where $(\cdot)^{(0)}$ denotes the zero-th degree (invariant) features alone.

\textbf{Atom-level Objective}~~~The atom-level objective instead produces one prediction $\hat{\mathbf{s}}_{\mathrm{y},i}$ for every target atom $i\in \mathcal{I}_\mathrm{y}$, which is compared against the corresponding target encoder representation $\mathbf{s}_{\mathrm{y},i}$. In addition to the context features, the atom-level readout head of the predictor is provided with an equivariant relative positional encoding $\mathbf{q}_i$ of the target atom’s position relative to the context subgraph,
\begin{align}
\hat{\mathbf{s}}_{\mathrm{y},i}=R^\text{atom}_\phi(\mathbf{\bar{h}}_\mathrm{x},\mathbf{q}_i)=\text{MLP}(\mathbf{u}_i), \quad i\in \mathcal{I}_\mathrm{y},
\end{align}

where $\mathbf{u}_i$ is constructed by computing the inner product between each feature channel in $\mathbf{q}_i$ and $\mathbf{\bar{h}}_\mathrm{x}$ and concatenating the individual inner products. We refer to Appendix \cref{app:atom_representation_construction,app:equivariant_relative_positions} for details on the construction of $\mathbf{u}_i$ and $\mathbf{q}_i$ respectively.

\textbf{Atom-JEPA Loss}~~~We use mean squared error between predicted and target representations, $\ell(\mathbf{a},\mathbf{b}) = \frac{1}{C}\|\mathbf{a}-\mathbf{b}\|_2^2,$ where $C$ is the feature dimension as the loss. The substructure and atom-level objectives are thus defined as 
\begin{align}
    \mathcal{L}_{\mathrm{sub}}^{x\rightarrow y} = \frac{1}{B} \sum_{g=1}^{B} \ell\left( \hat{\bar{\mathbf{s}}}_{\mathrm{y}}^{(g)}, \bar{\mathbf{s}}_{\mathrm{y}}^{(g)} \right) \quad \text{and} \quad 
\mathcal{L}_{\mathrm{atom}}^{x\rightarrow y} = \frac{1}{\sum_{g=1}^{B}|\mathcal{I}_\mathrm{y}^{(g)}|} \sum_{g=1}^{B} \sum_{i\in\mathcal{I}_\mathrm{y}^{(g)}} \ell\left( \hat{\mathbf{s}}_{\mathrm{y},i}, \mathbf{s}_{\mathrm{y},i} \right),
\end{align}
where $x\rightarrow y$ denotes the prediction direction in which $G_x$ is used as the context graph and $G_y$ as the target graph and $B$ is the batch size. The subgraph loss is a mean over the $B$ graphs in a batch, whilst the atom-level loss is a mean over all the target atoms in a batch $B$. The final loss is then,
\begin{equation}
    \mathcal{L}_{\text{Atom-JEPA}}=\lambda_{\text{atom}}\mathcal{L}_{\text{atom}}+\lambda_{\text{sub}}\mathcal{L}_{\text{sub}},
\end{equation}
where $\lambda_{\{\mathrm{atom},  \mathrm{sub}\}}$ control the weight of each objective.

\section{Experiments}
%\subsection{Pretraining and Downstream Evaluation}
We use EquiformerV3~\citep{eqv3} as the SE(3) equivariant GNN architecture for the context encoder, target encoder, and predictor. For molecular pretraining, we use $\sim$19 million molecules from Uni-Mol~\citep{zhou2023unimol}; for crystal pretraining, we use $\sim$1.7 million structures from Alexandria~\citep{Cavignac2026Alexandria} with energies less than $50~\mathrm{meV/atom}$ above the convex hull. Details about datasets and preprocessing are provided in App. \ref{app:datasets}. Pretraining details are provided in App. \ref{app:pretrain_details}.
After pretraining, we finetune the context encoder for downstream property prediction across both domains. For molecules, we consider ADMET prediction and quantum-chemical property prediction on QM9~\citep{qm9}. For crystals, we evaluate materials-property prediction using the Matbench benchmark suite~\citep{matbench}.
We further conduct ablation studies to assess the effects of pretraining, varying the pretraining prediction objective, and the scale and composition of the pretraining data (\cref{sec:ablation_studies}). 

\begin{figure}[t]
    \centering
    \includegraphics[width=0.99\linewidth]{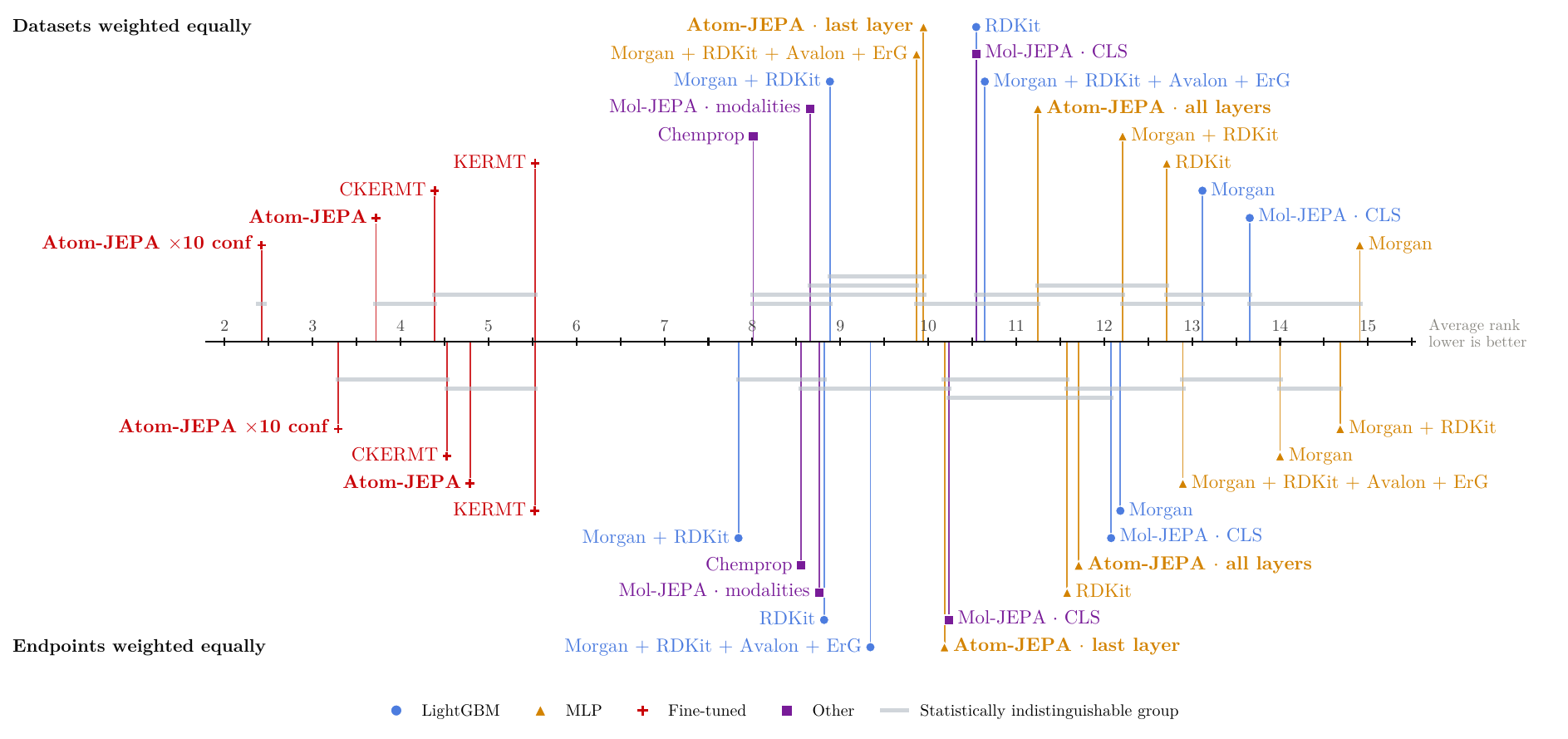}
    \caption{Average rank across 38 endpoints from three benchmarks (Biogen 4 endpoints, ChEMBL-MT 25 endpoints, ExpansionRx 9 endpoints). Top part weights each dataset equally, whereas bottom part weighs each endpoint equally. Bars join models whose paired rank difference could not be distinguished from zero by a stratified seed/fold bootstrap over endpoints ($10^6$ replicates, resampled within dataset, $95\%$ BCa interval). Tables with individual results can be found in \cref{app:fulladmet}.}
    \label{fig:admet-combined-rank}
\end{figure}

\subsection{Experimental Results}

\textbf{Results on ADMET}~~~We extensively investigate the transfer learning for ADMET tasks on three different datasets, spanning various amounts of data, label sparsity, and splitting methods:
\begin{enumerate}
    \item ChEMBL-MT~\citep{DBLP:journals/corr/abs-2510-12719}: Multi-task, multiple sources of data, 25 targets, Taylor-Butina clustering on Morgan fingerprints with pre-defined splits.
    \item ExpansionRx~\citep{macdermott_opeskin_2026_21784075}: Multi-task, single source for collection, 10 targets, temporal split of training and test data.
    \item Biogen ADME~\citep{10.1021/acs.jcim.3c00160}: Multi-task, internal source, 6 targets, Bemis–Murcko scaffold and Taylor-Butina clustering on Morgan fingerprints.
\end{enumerate}
For completeness, we also include results on TDC-ADMET~\citep{DBLP:conf/nips/HuangFG0RLCXSZ21} in \cref{app:fulladmet}, but exclude them from the statistical analysis, due to reported data acquisition and leaderboard inconsistencies ~\citep{koleiev2026end}. As Atom-JEPA requires 3D structures, we use the Uni-Mol pipeline to generate conformers (described in \cref{appendix:conformer_generation}). 
 To ensure a fair comparison, we tune the hyperparameters using the training data from a cluster split on Biogen from \citet{DBLP:journals/corr/abs-2510-12719}. We use the average validation MAE over endpoints as our selection criteria (details in \cref{appendix:admet_finetune}). We follow the transfer learning setup of multi-head multi-target training strategy used by CKERMT~\citep{DBLP:journals/corr/abs-2606-11508} with some minor changes (see \cref{app:finetune_details}).

 % Table
\begin{table}[t]
\caption{\textbf{ADMET results.} For each endpoint, the model with the lowest mean error across folds is designated as \emph{Best}, and an unpaired Tukey's test is used to identify additional models that are statistically indistinguishable (\emph{Indist.}) from the best model at the $95\%$ confidence level (see Appendix \ref{appendix:statistical_test}). \emph{Sum} denotes the total best or indistinguishable endpoints. $^*$ reports only the mean error across splits; these values are treated as constant in the statistical tests. $^\dag$ did not report Biogen results for two targets and is therefore excluded from the statistical tests for these targets.}
    \label{table:admet_comparison}
    \centering
    \normalsize
    \scalebox{\tablescale}{%
        \begin{tabular}{lccccccc}
            \toprule
            & \multicolumn{2}{c}{Biogen (6)} & \multicolumn{2}{c}{ChEMBL MT (25)} & \multicolumn{2}{c}{ExpansionRx (9)} \\
            \cmidrule(lr){2-3} \cmidrule(lr){4-5} \cmidrule(lr){6-7}
            \textbf{Model} & \textbf{Best\,/\,Indist.} & \textbf{Sum} & \textbf{Best\,/\,Indist.} & \textbf{Sum} & \textbf{Best\,/\,Indist.} & \textbf{Sum} & \textbf{Total} (40) \\
            \midrule
            \rowcolor{atomjepablue} Atom-JEPA $\times$10 conf & 3\,/\,3 & 6 & 6\,/\,10 & 16 & 7\,/\,2 & 9 & 31 \\
            \rowcolor{atomjepablue} Atom-JEPA & 0\,/\,6 & 6 & 0\,/\,15 & 15 & 0\,/\,9 & 9 & 30 \\
            CKERMT~\citep{DBLP:journals/corr/abs-2606-11508} & 0\,/\,6 & 6 & 11\,/\,8 & 19 & 1\,/\,3 & 4 & 29 \\
            KERMT\textsuperscript{*}~\citep{DBLP:journals/corr/abs-2510-12719} & 1\,/\,1\textsuperscript{\dag} & 2 & 3\,/\,9 & 12 & 0\,/\,1 & 1 & 15\textsuperscript{\dag} \\
            Chemprop~\citep{graff2026chemprop} & 1\,/\,3 & 4 & 0\,/\,12 & 12 & 0\,/\,1 & 1 & 17 \\
            Mol-JEPA $\cdot$ modalities~\citep{rottach2026moljepamultimodaljointembedding} & 1\,/\,2 & 3 & 2\,/\,5 & 7 & 0\,/\,3 & 3 & 13 \\
            LightBGM Morgan + RDKit & 0\,/\,2 & 2 & 3\,/\,9 & 12 & 1\,/\,2 & 3 & 17 \\
            \bottomrule
        \end{tabular}%
    }
\end{table}

    We present the aggregated results from the three main ADMET datasets in \cref{fig:admet-combined-rank,table:admet_comparison}. Atom-JEPA reaches on par or significantly better performance than the next best model, CKERMT, and outperforms fingerprint-based models. Notably, Atom-JEPA has no explicitly coded expert chemical knowledge in comparison to all other models, including the self-supervised models of KERMT and CKERMT. This also explains why we find that Atom-JEPA's predictions are the least correlated with other models tested, see Appendix \ref{appendix:prediction_correlation}. This is especially useful for the common ADMET practice of ensembling models. We additionally observe significantly improved prediction accuracy when averaging predictions over our 10 generated conformers (Atom-JEPA $\times10$ conf) for each molecule, suggesting that different conformers can provide different information to the prediction. 

Interestingly, Atom-JEPA performs especially well on ExpansionRX, a dataset that differs notably from other benchmarks~\citep{DBLP:journals/corr/abs-2606-11508}, including our pretraining data Uni-Mol, where only a single molecule shares the same scaffold (see \cref{tab:unimol-evaluation-overlap}). This suggests that the features learned by Atom-JEPA are more robust at transferring, compared to existing models.

\textbf{Results on QM9}~~~QM9 consists of $\sim$130k small organic molecules with 12 quantum-chemical target properties. We use the random 110k/10k/11k split most commonly used in the literature; finetuning details are given in \cref{app:finetune-qm9}. The same optimization protocol and hyperparameters are used for all targets, and all readout heads use sum aggregation over per-atom features.

\cref{tab:qm9_finetune_baselines} compares Atom-JEPA with previous models. 3D-EMGP, DenoiseVAE, TriForces and JMP-L use different random data partitions; their results are shown in grey for reference and excluded from highlighting. Atom-JEPA outperforms all SSL pretraining methods on 11/12 targets and ties with SliDe on the remaining target, $C_\nu$. Overall, Atom-JEPA achieves the best results on 8/12 targets with tied performance on $\alpha$ and $C_\nu$. On $G$ and $U_0$ it is within 0.3\% of the best result, making it best or within 0.3\% of the best on 10/12 targets. Relative to Equiformer$_{\text{V}2}$, the predecessor of our Equiformer$_{\text{V}3}$ backbone, Atom-JEPA reduces MAE on all 12 targets, with a mean relative reduction of 36\%.

GotenNet$_L$, trained from scratch, remains strong, achieving the best performance on 4/12 targets, which suggests that this architecture is particularly well suited to QM9. As GotenNet is a 3D equivariant GNN, Atom-JEPA pretraining could in principle be applied to its backbone. Beyond using a different random split with a smaller test set and correspondingly larger training set, JMP-L is not directly comparable, as it is pretrained with supervised DFT labels aligned with several QM9 targets, on molecules that overlap with QM9. Even so, Atom-JEPA achieves lower error on $\alpha$, $\mu$, and $R^2$, three targets not aligned with JMP-L's supervised pretraining labels, suggesting stronger transfer.

\begin{table}[t]
\caption{\textbf{QM9 results.} The evaluation metric is MAE $(\downarrow)$.
\textbf{Bold:} best result in each column; \underline{underline}: within 10\%
of the best. \textcolor{gray}{Grey} rows with $\dagger$ use a different random
train/validation/test split from 110k/10k/11k and are shown
for reference only, excluded from highlighting. ``--'' denotes unreported targets.}
\label{tab:qm9_finetune_baselines}
\centering
\normalsize
\scalebox{\tablescale}{%
\begin{tabular}{lcccccccccccc}
\toprule
Task
& $\alpha$
& $\Delta\varepsilon$
& $\varepsilon_{\text{HOMO}}$
& $\varepsilon_{\text{LUMO}}$
& $\mu$
& $C_\nu$
& $G$
& $H$
& $R^2$
& $U$
& $U_0$
& ZPVE \\
Units
& $ma_0^3$
& meV
& meV
& meV
& mD
& $\frac{\text{mcal}}{\text{mol K}}$
& meV
& meV
& $ma_0^2$
& meV
& meV
& meV \\
\midrule

\multicolumn{13}{l}{\textit{No pretraining}} \\
\quad MACE~\citep{mace}
& 38 & 42 & 22 & 19 & 15 & 21
& 5.5 & 4.7 & 210 & 4.1 & 4.1 & 1.23 \\

\quad Equiformer$_{\text{V}2}$~\citep{equiformerv2}
& 47 & 29.0 & 14.4 & 13.3 & 9.9 & 23
& 7.57 & 6.22 & 186 & 6.49 & 6.17 & 1.47 \\

\quad GotenNet$_L$~\citep{gotennet}
& \textbf{28} & \textbf{19.8} & \underline{13.4} & 12.2
& \underline{6.7} & \textbf{19}
& \textbf{4.98} & \underline{3.30} & \textbf{24}
& \underline{3.41} & \textbf{3.37} & \underline{1.08} \\

\addlinespace
\multicolumn{13}{l}{\textit{Supervised pretraining}} \\
\quad Transformer-M~\citep{transformer-m}
& 41 & 27.4 & 17.5 & 16.2 & 37 & 22
& 9.63 & 9.39 & 75 & 9.41 & 9.37 & 1.18 \\

\quad \textcolor{gray}{JMP-L$^\dagger$~\citep{jmp}}
& \textcolor{gray}{32}
& \textcolor{gray}{19.1}
& \textcolor{gray}{8.8}
& \textcolor{gray}{8.6}
& \textcolor{gray}{8.0}
& \textcolor{gray}{17}
& \textcolor{gray}{4.3}
& \textcolor{gray}{2.8}
& \textcolor{gray}{163}
& \textcolor{gray}{2.8}
& \textcolor{gray}{2.9}
& \textcolor{gray}{0.9} \\

\addlinespace
\multicolumn{13}{l}{\textit{Self-supervised pretraining only}} \\
\quad \textcolor{gray}{3D-EMGP$^\dagger$~\citep{3demgp}}
& \textcolor{gray}{57}
& \textcolor{gray}{37.1}
& \textcolor{gray}{21.3}
& \textcolor{gray}{18.2}
& \textcolor{gray}{20}
& \textcolor{gray}{26}
& \textcolor{gray}{9.30}
& \textcolor{gray}{8.70}
& \textcolor{gray}{92}
& \textcolor{gray}{8.60}
& \textcolor{gray}{8.60}
& \textcolor{gray}{1.38} \\

\quad GeoSSL-DDM~\citep{geossl}
& 46 & 40.2 & 23.5 & 19.4 & 15 & 24
& 7.65 & 7.09 & 122 & 6.99 & 6.92 & 1.31 \\

\quad Coord~\citep{zaidi2023pretraining,pmlr-v202-feng23c}
& 52 & 31.8 & 17.7 & 14.3 & 12 & \underline{20}
& 6.91 & 6.45 & 450 & 6.11 & 6.57 & 1.71 \\

\quad Frad (VRN)~\citep{ni2024pre}
& 42 & 27.7 & 17.9 & 13.8 & 11 & 21
& 6.03 & 6.01 & 354 & 5.35 & 5.41 & 1.63 \\

\quad Frad (RN)~\citep{pmlr-v202-feng23c}
& 37 & 27.8 & 15.3 & 13.7 & 10 & \underline{20}
& 6.19 & 5.55 & 342 & 5.62 & 5.33 & 1.42 \\

\quad \textcolor{gray}{DenoiseVAE$^\dagger$~\citep{liu2025denoiseVAE}}
& \textcolor{gray}{65}
& \textcolor{gray}{26.0}
& \textcolor{gray}{14.2}
& \textcolor{gray}{11.9}
& \textcolor{gray}{7.9}
& \textcolor{gray}{15}
& \textcolor{gray}{5.35}
& \textcolor{gray}{4.19}
& \textcolor{gray}{62}
& \textcolor{gray}{4.03}
& \textcolor{gray}{4.31}
& \textcolor{gray}{1.03} \\

\quad SliDe~\citep{slide}
& 37 & 26.2 & 13.6 & 12.3 & 8.7 & \textbf{19}
& \underline{5.37} & 4.26 & 341 & 4.29 & 4.28 & 1.52 \\

\quad \textcolor{gray}{TriForces$^\dagger$~\citep{triforces}}
& \textcolor{gray}{75}
& \textcolor{gray}{38.8}
& \textcolor{gray}{20.2}
& \textcolor{gray}{20.4}
& \textcolor{gray}{18.0}
& \textcolor{gray}{31}
& \textcolor{gray}{--}
& \textcolor{gray}{--}
& \textcolor{gray}{--}
& \textcolor{gray}{--}
& \textcolor{gray}{8.9}
& \textcolor{gray}{--} \\

\rowcolor{atomjepablue}
\quad Atom-JEPA
& \textbf{28} & 24.6 & \textbf{12.2} & \textbf{10.9}
& \textbf{6.1} & \textbf{19}
& \underline{4.99} & \textbf{3.28} & 29
& \textbf{3.38} & \underline{3.38} & \textbf{1.06} \\

\bottomrule
\end{tabular}%
}
\end{table}

\textbf{Results on Matbench}~~~\cref{tab:matbench} reports means and standard deviations across the five official folds for 8 Matbench tasks; training details are in \cref{app:finetune-matbench}. Baseline results are taken from the official leaderboard for CGCNN, ALIGNN, and coGN; from the original papers for Crystal Twins (CT) and JMP-L; and from the Triforces paper for MACE, Orb, eSEN, and both TriForces variants.

\begin{table*}[h]
\caption{\textbf{Matbench results.}
The evaluation metric is MAE $(\downarrow)$ for all tasks except
MP Is Metal, which uses F1 $(\uparrow)$. Values are means over
five folds, with standard deviations where available.
Pretrained models are ordered by increasing pretraining dataset size,
shown in parentheses after each method.
\textbf{Bold:} best result in each task. ``--'' denotes targets not
reported by the original authors. * results from ~\citep{triforces}.}
\label{tab:matbench}
\centering
\normalsize
\scalebox{\tablescale}{%
\begin{tabular}{lcccccccc}
\toprule
Tasks
& Phonons & Dielectric & Log GVRH & Log KVRH
& Perovskites & MP Gap & MP E Form & MP Is Metal \\

Units
& cm$^{-1}$ & -- & $\log_{10}$(GPa) & $\log_{10}$(GPa)
& meV & eV & meV/atom & F1 \\

\#Samples
& 1{,}265 & 4{,}764 & 10{,}987 & 10{,}987
& 18{,}928 & 106{,}113 & 132{,}752 & 106{,}113 \\
\midrule

\multicolumn{9}{l}{\textit{No pretraining}} \\
\quad CGCNN~\citep{cgcnn}
& $57.8 \err{12.3}$ & $.599 \err{.083}$
& $.090 \err{.002}$ & $.071 \err{.003}$
& $45.2 \err{.7}$ & $.297 \err{.004}$
& $33.7 \err{.6}$ & $\mathbf{.946} \err{.008}$ \\

\quad ALIGNN~\citep{alignn}
& $29.5 \err{2.1}$ & $.345 \err{.087}$
& $.072 \err{.001}$ & $.057 \err{.003}$
& $28.8 \err{.9}$ & $.186 \err{.003}$
& $21.5 \err{.5}$ & $.902 \err{.002}$ \\

\quad coGN~\citep{cogn}
& $29.7 \err{2.0}$ & $.309 \err{.086}$
& $.069 \err{.001}$ & $.054 \err{.003}$
& $26.9 \err{.8}$ & $.156 \err{.002}$
& $17.0 \err{.3}$ & $.901 \err{.003}$ \\

\addlinespace
\multicolumn{9}{l}{\textit{With pretraining}} \\
\quad CT$_{\mathrm{SimSiam}}$~\citep{crystaltwins} (0.43M)
& $48.9 \err{7.7}$ & $.417 \err{.079}$
& $.087 \err{.003}$ & $.067 \err{.003}$
& $42.0 \err{1.0}$ & $.281 \err{.025}$
& $37.0 \err{.0}$ & -- \\

\rowcolor{atomjepablue}
\quad Atom-JEPA (1.7M)
& $23.3 \err{4.5}$ & $.268 \err{.074}$
& $.067 \err{.001}$ & $.047 \err{.002}$
& $29.1 \err{1.0}$ & $.162 \err{.002}$
& $20.3 \err{.2}$ & $.908 \err{.002}$ \\

\quad Orb$^*$~\citep{orb} (4M)
& 26.2 & .251 & .063 & .051
& 30.7 & .194 & 21.1 & .895 \\

\quad MACE$^*$~\citep{mace}(4M)
& 36.7 & .279 & .082 & .055
& 61.4 & .370 & 40.8 & .858 \\

\quad eSEN$^*$~\citep{esen} (4M)
& 57.8 & .205 & .093 & .072
& 40.1 & .392 & 83.5 & .811 \\

\quad TriForces MACE~\citep{triforces} (5M)
& 27.6 & \textbf{.159} & .073
& .050 & 35.1 & .250 & 34.4 & .876 \\

\quad TriForces eSEN~\citep{triforces} (5M)
& \textbf{19.5} & .232 & \textbf{.058}
& \textbf{.043} & \textbf{25.6} & .139 & 20.2 & .888 \\

\quad JMP-L~\citep{jmp} (120M)
& 20.6 & .249 & .059 & .045 & 26.0
& \textbf{.091} & \textbf{10.1} & -- \\

\bottomrule
\end{tabular}
}
\end{table*}

Compared with models trained from scratch, Atom-JEPA achieves the lowest error on the four smaller tasks, with improvements of up to 21\%. On the larger MP regression tasks, coGN remains stronger, although pretraining improves our encoder by 12--15\% on these tasks relative to training from scratch (\cref{sec:ablation_studies}). Among pretrained models with at most 4M pretraining structures, Atom-JEPA achieves the lowest error on 6/8 tasks, and outperforms Crystal Twins on all tasks. This holds even though MACE, Orb, and eSEN are pretrained on 4M structures with energy and force labels, while Atom-JEPA is pretrained on 1.7M structures without any labels.

Comparison to TriForces (5M structures) highlights the importance of the backbone architecture: Atom-JEPA outperforms TriForces MACE on 7/8 tasks, whereas compared to TriForces eSEN it performs worse, with the exception of MP Is Metal (.908 versus .888 F1) and MP E Form, which is essentially tied (20.3 versus 20.2 meV/atom). JMP-L, pretrained with supervision on 120M structures, achieves lower MAE than Atom-JEPA on all seven regression tasks.

\subsection{Ablation Studies}
\label{sec:ablation_studies}
\textbf{Effect of Atom-JEPA pretraining}~~~We compare the Atom-JEPA pretrained encoder to a randomly initialized counterpart on both ADMET and Matbench, using the training procedures described in \cref{appendix:admet_finetune,app:finetune-matbench}. Atom-JEPA pretraining improves downstream performance for both molecular and crystalline tasks (\cref{tab:ablation_pretraining}). On the molecular ADMET benchmarks, pretraining reduces error by 9.6\% on Biogen and 8.5\% on ExpansionRx. ChEMBL-MT does not see an improvement, but we suspect that it is confounded by hyperparameters not generalizing well from Biogen to the much larger dataset size. On Matbench, pretraining improves all eight tasks and all 40 task--fold combinations, yielding mean foldwise relative MAE reductions of 10.1--14.5\% across the seven regression tasks and increasing mean classification F1 from 0.9010 to 0.9083. These gains span datasets ranging from 1,265 to 132,752 samples and persist on the largest regression benchmarks: MP Gap and MP E Form improve by 14.5\% and 12.0\%, respectively. Even without pretraining, the encoder is competitive with established from-scratch models, matching or exceeding their performance on several tasks (\cref{tab:matbench} \& \cref{tab:ablation_pretraining}) . This confirms that the gains from pretraining do not simply stem from a weak baseline.

\begin{figure}[t]
    \centering
    \includegraphics[width=\linewidth]{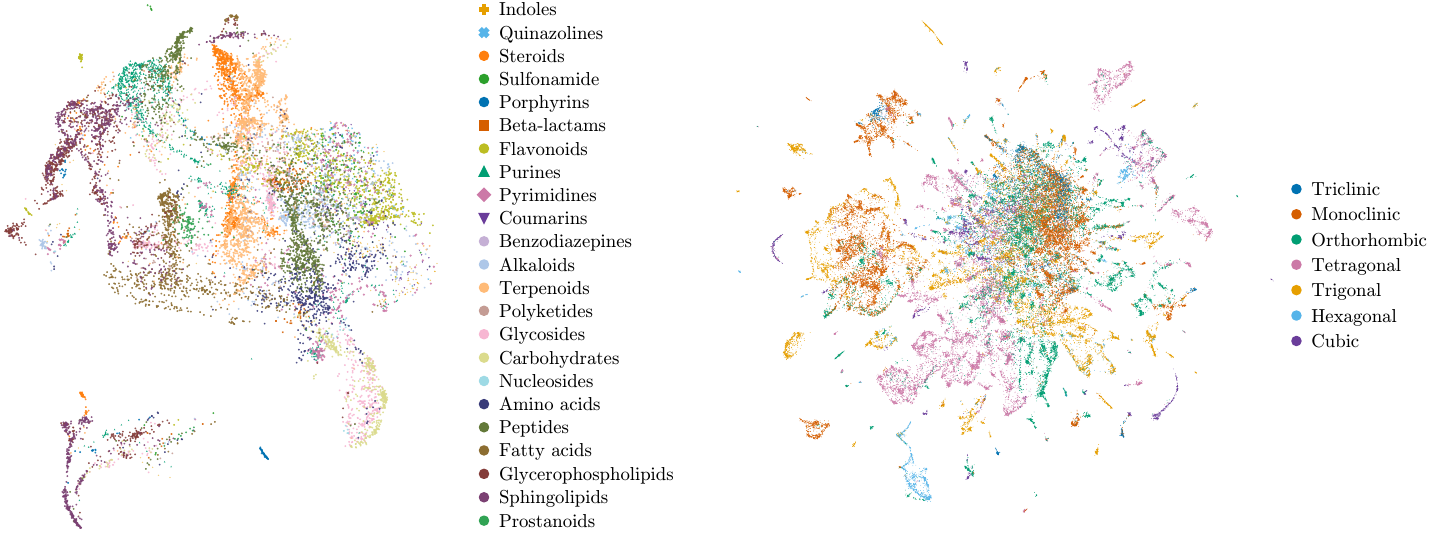}
    \caption{
        UMAP projections of Atom-JEPA representations for molecules annotated as belonging to one chemical family from the 32,252 molecules in the manually annotated three-star subset of ChEBI~\citep{chebi}, colored by chemical family, and 50,000 randomly sampled crystals from the Alexandria pretraining dataset, colored by crystal system.
    }
    \label{fig:umap-molecules-crystals}
\end{figure}

\begin{table*}[h]
\caption{\textbf{Effect of Atom-JEPA pretraining.} Identical encoder architecture is compared with and without pretraining on identical official folds; values are
means and standard deviations across the five folds or endpoints for ADMET tasks.
$\Delta$ (rel.) is the mean fold-wise or endpoint relative improvement.
``Impr.'' counts folds or endpoints on which pretraining wins;
$p$ is a two-sided paired $t$-test over the five Matbench folds or mean MAE over the folds in the endpoints for ADMET tasks.}
\label{tab:ablation_pretraining}
\centering
\normalsize
\scalebox{\tablescale}{%
\begin{tabular}{lcc>{\columncolor{atomjepablue}}cccc}
\toprule
Task (Units) & \#Samples & No pretraining & Atom-JEPA
& $\Delta$ (rel.) & Impr..& $p$ \\
\midrule
\multicolumn{7}{l}{\textbf{ADMET datasets}} \\
\midrule
Biogen
& 3,521
& $.3825$ & $\mathbf{.3458}$
& $9.6\%$ & 6/6 & $.011$ \\
ExpansionRX
& 7,608
& $.3673$ & $\mathbf{.3361}$
& $8.5\%$ & 9/9 & $<.001$ \\
ChEMBL-MT & 114,112 & $\mathbf{.4543}$ & .4567 & $-.5\%$ & 12/25 & .57 \\
\midrule
\multicolumn{7}{l}{\textbf{Matbench}} \\
\midrule
Phonons (cm$^{-1}$)
& 1{,}265
& $26.9 \err{4.0}$ & $\mathbf{23.3} \err{4.5}$
& $13.5\%$ & 5/5 & $.015$ \\
Dielectric (--)
& 4{,}764
& $.298 \err{.084}$ & $\mathbf{.268} \err{.074}$
& $10.1\%$ & 5/5 & $.011$ \\
Log GVRH ($\log_{10}$(GPa))
& 10{,}987
& $.0773 \err{.0010}$ & $\mathbf{.0670} \err{.0010}$
& $13.3\%$ & 5/5 & $<.001$ \\
Log KVRH ($\log_{10}$(GPa))
& 10{,}987
& $.0539 \err{.0034}$ & $\mathbf{.0473} \err{.0021}$
& $12.3\%$ & 5/5 & $.001$ \\
Perovskites (meV)
& 18{,}928
& $32.5 \err{.6}$ & $\mathbf{29.1} \err{1.0}$
& $10.5\%$ & 5/5 & $<.001$ \\
MP Gap (eV)
& 106{,}113
& $.1896 \err{.0022}$ & $\mathbf{.1622} \err{.0022}$
& $14.5\%$ & 5/5 & $<.001$ \\
MP E Form (meV/atom)
& 132{,}752
& $23.03 \err{.29}$ & $\mathbf{20.26} \err{.19}$
& $12.0\%$ & 5/5 & $<.001$ \\
MP Is Metal (F1)
& 106{,}113
& $.9010 \err{.0021}$ & $\mathbf{.9083} \err{.0024}$
& $.8\%$ & 5/5 & $<.001$ \\
\bottomrule
\end{tabular}%
}
\end{table*}

We hypothesize that these performance gains arise because Atom-JEPA pretraining encourages representations to organize according to high-level features that transfer well across downstream tasks in both domains. The UMAP visualizations in \cref{fig:umap-molecules-crystals} suggest such organization may be by chemical family or crystal system, providing a qualitative indication consistent with this hypothesis. In contrast, an encoder trained from scratch starts from representations shaped by the inductive biases of the backbone architecture, which primarily organize systems according to size and composition initially. For a deeper analysis of the learned representations of Atom-JEPA in \cref{app:learned-representations}.

\begin{figure*}[t]
    \centering
    \includegraphics[width=\textwidth]{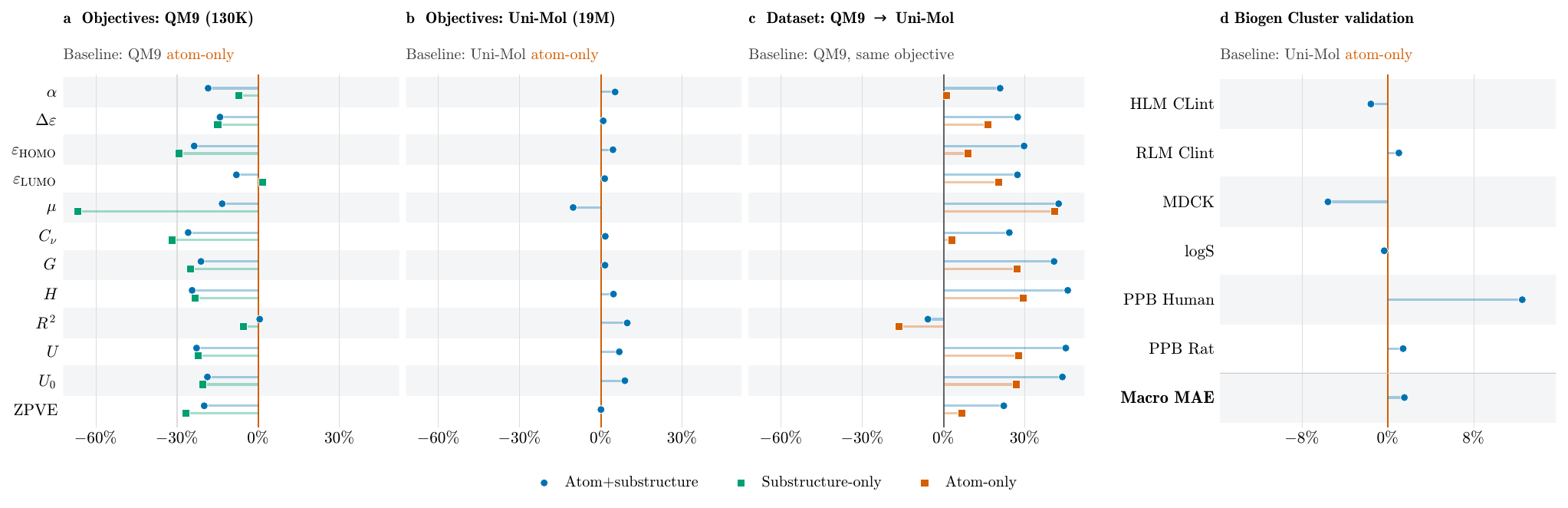}
    \caption{
        \textbf{Effect of pretraining objectives and dataset.} Frozen-feature
        MLP probes across 12 QM9 targets.
        The horizontal axis shows relative MAE reduction to the baseline of each panel.
        Positive values indicate improvement.
        \textbf{(a)} Joint (Atom+substructure) and substructure-only
        objectives relative to atom-only pretraining on QM9.
        \textbf{(b)} The joint objective relative to atom-only
        pretraining on Uni-Mol.
        \textbf{(c)} Uni-Mol pretraining relative to QM9 pretraining,
        with the objective held fixed.
        \textbf{(d)} The joint objective relative to atom-only
        pretraining on Uni-Mol on Biogen ADME validation objective.
    }
    \label{fig:qm9_pretraining}
\end{figure*}

\textbf{Pretraining objective}~~~Next, we examine how the choice of pretraining prediction objective affects downstream performance. In addition to the joint-objective encoder pretrained on the 19M molecules from Uni-Mol, we pretrain an encoder using only the atom-level objective on the same data. We also pretrain three encoders on the 130K molecules from the QM9 dataset using atom-only, substructure-only, and joint objectives. All encoders are evaluated using all-layer MLP probes on frozen features, on the 12 QM9 properties. With QM9 pretraining, atom-only achieves the lowest MAE on 10/12 properties, while substructure-only performs best on $\varepsilon_{\mathrm{LUMO}}$ and the joint objective performs best on $R^2$ (\cref{fig:qm9_pretraining}a). With Uni-Mol pretraining, the joint objective improves over atom-only on 10/12 properties, ties on ZPVE, and performs worse only on $\mu$ (\cref{fig:qm9_pretraining}b). We additionally compare the two Uni-Mol-pretrained encoders on the six Biogen ADME tasks. The joint objective performs better on average, although atom-only remains competitive, performing better or similarly on 3/6 tasks (\cref{fig:qm9_pretraining}d). These results suggest that the benefit of combining atom-level and substructure-level objectives depends on both the pretraining dataset and the downstream task.

\textbf{Data scaling}~~~Increasing the pretraining dataset from 130k (QM9) to 19M molecules (Uni-Mol) improves both atom-only and joint-objective pretraining on 11/12 QM9 properties, with $R^2$ as the sole exception (\cref{fig:qm9_pretraining}c). The gains are particularly pronounced for the joint objective, including MAE reductions of 27\% for $\Delta\varepsilon$, 42.5\% for $\mu$, and 45.9\% for $H$. Notably, Uni-Mol pretraining substantially improves performance even though QM9 pretraining already exactly matches the molecular domain of the downstream evaluation. While this comparison changes both dataset size and composition, the results suggest that scaling Atom-JEPA pretraining to larger and more chemically and structurally diverse datasets could further improve downstream performance.

\section{Conclusion}
In this work, we proposed Atom-JEPA, a self-supervised pretraining framework for 3D atomic systems that combines atom-level and substructure-level prediction in latent space. Using structural data alone without expensive property labels nor molecular attributes as the pretraining signal, Atom-JEPA learns representations that transfer effectively to molecular ADMET, quantum-chemical, and crystalline materials property prediction. Ablation studies demonstrate consistent improvement over training the same encoder from scratch, and indicate potential for further increasing downstream performance by scaling to larger and more diverse pretraining data. Overall, Atom-JEPA establishes latent-space prediction from 3D structure as a promising approach to learning transferable representations for molecules and materials, reducing reliance on property-labeled pretraining data.

\subsection*{Acknowledgements}
This work was supported by the Novo Nordisk Foundation under grant no NNF25OC0095542 by funding KHS, FRJC and MNS (AI-driven materials optimization for light trapping in thin-film solar cells). This work was supported by the Novo Nordisk Foundation (Grant ID: NNF25OC0105141) by funding RHT, MJ and MNS and via access to the Gefion AI Supercomputer, operated by the Danish Centre for AI Innovation (DCAI).

\subsection*{Reproducibility statement}
To facilitate full reproducibility, we report all hyperparameters used in our pretraining and downstream fine-tuning experiments. The source code for pretraining and all finetuning experiments can be found at \url{https://github.com/khelverskovp/atom-jepa} together with experimental results across all folds. Pretrained model checkpoints can be found at \url{https://huggingface.co/atom-jepa/atom-jepa}.

\bibliography{atom-jepa}
\bibliographystyle{atom-jepa}

\appendix
\etocdepthtag.toc{appendix}

\begingroup
  \etocsettagdepth{main}{none}
  \etocsettagdepth{appendix}{subsection}
  \etocsettocdepth{subsection}
  \renewcommand{\contentsname}{Appendix Contents}
  \tableofcontents
\endgroup

\newpage

\section{Detailed Methodology}
\label{app:method}
The overall idea of Atom-JEPA is analogous to I-JEPA for images~\citep{i-jepa}, but redesigned for atomic systems in 3-dimensional space: Given part of a molecule or crystal (context view), predict the representation of the remaining part (target view). We consider two prediction objectives: (1) an atom-level objective that predicts the feature representation of each atom in the target view, and (2) a substructure-level objective that predicts the mean-pooled representation of all target atoms. For the context encoder, target encoder and predictor, we use a 3D geometric graph neural network (GNN). We only use data that are either in ground-state or close to local minima of the potential energy surface. We posit that using data in off-equilibrium states (either rattled systems or from MD trajectories) without explicitly telling the model something about the stability of the system (e.g. via total energy of the system) could make the learning problem much more ambigious. 

\subsection{Atomic 3D Graph Representation and Partitioning}
\label{app:part_strat}
We represent atomic systems as 3D graphs and partition each system into complementary context and target subgraphs. An atomic system containing $N$ atoms can be represented by a 3D graph $G=(V,E)$ with the node set $V=\{(z_i,\mathbf{r}_i)\}^N_{i=1}$, where $z_i$ denotes the atom type of atom $i$ and $\mathbf{r}_i\in \mathbb{R}^3$ its 3D position. The edge set $E$ encodes spatial proximity between atoms within a cutoff radius $r_c$. Given a partition of the atomic indices into a context set $\mathcal{I}_x$ and a target set $\mathcal{I}_y$, we define the corresponding induced subgraphs as
$$
G_x=G[\mathcal{I}_x]=(V_x,E_x), \quad G_y=G[\mathcal{I}_y]=(V_y,E_y),
$$
where $V_x=\{(z_i,\mathbf{r}_i):i\in \mathcal{I}_x\}$ and $V_y=\{(z_i,\mathbf{r}_i):i\in \mathcal{I}_y\}.$ The two index sets form a partition of the atoms in the atomic system, such that $\mathcal{I}_x \cap \mathcal{I}_y = \emptyset$ and $\mathcal{I}_x\cup\mathcal{I}_y=\{1,\dots,N\}$.

\paragraph{Molecular Graphs. } For a molecule containing $N$ atoms, the neighborhood of atom $i$ is defined by all atoms within the cutoff radius $r_c$,
$$
\mathcal{N}(i)=\{j\in \{1,\dots,N\}\setminus \{i\} \mid \|\mathbf{r}_j-\mathbf{r}_i\|<r_c\}.
$$
The edge set $E$ is obtained from these neighborhoods. For each neighboring pair $(i,j)$, we define the relative displacement vector $\mathbf{r}_{ij}=\mathbf{r}_j-\mathbf{r}_i$, together with the interatomic distance $\|\mathbf{r}_{ij}\|$ and unit direction $\hat{\mathbf{r}}_{ij}=\frac{\mathbf{r}_{ij}}{\|\mathbf{r}_{ij}\|}.$

\paragraph{Crystal Graphs. }
Crystalline systems are represented analogously, with the additional requirement that periodic images of atoms in neighboring unit cells are taken into account. Let the lattice be specified by three lattice vectors $\mathbf{l}_1,\mathbf{l}_2,\mathbf{l}_3$, collected in the lattice matrix $\mathbf{L}$. A periodic image of atom $i$, indexed by the integer translation vector $\mathbf{n}\in \mathbb{Z}^3$, has position $\mathbf{r}_i+\mathbf{n}\mathbf{L}.$ The periodic neighborhood of atom $i$ is therefore
$$
 \mathcal{N}(i)=\{(j,\mathbf{n})\in \{1,\dots,N\}\times \mathbb{Z}^3 \;\bigm|\; 0<\|\mathbf{r}_j+\mathbf{n}\mathbf{l}-\mathbf{r}_i\|<r_c\}.
$$
Although the underlying crystal is infinite, the graph contains only the $N$ atoms of the reference unit cell as nodes. Periodicity is represented through edges connecting these atoms to neighboring periodic images within the cutoff radius $r_c$. Similarly to molecules, for each such edge, we define the periodic displacement vector $\mathbf{r}_{ij\mathbf{n}}=\mathbf{r}_j+\mathbf{n}\mathbf{L}-\mathbf{r}_i,$ along with the distance $\|\mathbf{r}_{ij\mathbf{n}}\|$ and unit direction $\hat{\mathbf{r}}_{ij\mathbf{n}}$.

\paragraph{Context and Target Partitioning. }
Given an atomic system, we partition it into the context and target index sets $\mathcal{I}_x$ and $\mathcal{I}_y$ introduced above by partitioning the atoms using a $k$-EgoNet subgraph. We first sample an anchor atom $c\sim\text{Unif}(\{1,\dots,N\})$ and a hop count $k\sim\text{Unit}(K)$ from a predefined set of hop counts $K$. The $k$-hop EgoNet centered at $c$ is then defined by the index set
$$
\mathcal{N}_k(c)=\{i\in \{1,\dots,N\}\mid d(c,i)\leq k\},
$$
where $d(c,i)$ denotes the shortest-path distance between atoms $c$ and $i$, measured in number of edges. We take the atoms of the sampled EgoNet $\mathcal{N}_k(c)$ as the context graph $G_x$, and the atoms in the complementary part as the target graph $G_y$. If the sampled value of $k$ produces an EgoNet containing all atoms in the system, we instead use the largest value in $K$ for which the target graph remains non-empty. During training, the two subgraphs are used symmetrically: each subgraph acts once as the context $G_x$ and once as the target $G_y$, and the loss is evaluated in both directions. The anchor atom $c$ and hop count $k$ are resampled at each training epoch, yielding different partitions of the atomic system throughout training. To encourage spatially local partitions, the graph used to determine the $k$-hop EgoNet is constructed using a smaller cutoff radius \(r_k \ll r_c.\) This auxillary graph is used only to determine the partition $(\mathcal{I}_x,\mathcal{I}_y)$. Once the partition has been obtained, the full graph $G$, context graph $G_x$, and target graph $G_y$ are constructed using the larger interaction cutoff $r_c$. Thus, $r_k$ controls the locality of the sampled partition, whereas $r_c$ determines the atomic interactions represented in the graphs provided to the models.

\subsection{Context and Target Encoders}
The context encoder $f_\theta$ and target encoder $f_{\bar{\theta}}$ are both graph neural networks that take a 3D graph as input and produce atom-level feature representations. We use an equivariant architecture and write $\mathbf{s}_i=\{\mathbf{s}_i^{(l)}\}^{L_\text{max}}_{l=0}$ to denote atomic features, where $\mathbf{s}^{(0)}_i$ denote the invariant scalar features and $\mathbf{s}^{(l)}_i, l>0$ denote the higher-order equivariant features from $l=1,\dots,L_\text{max}$. The parameters of the context encoder $\theta$ are updated via gradient based optimization during training, while the parameters of the target encoder $\bar{\theta}$ are an exponential moving average of the context encoder parameters: $\bar{\theta}\leftarrow\tau\bar{\theta}+(1-\tau)\theta$, where $\tau\in[0,1]$ controls the rate of the exponential moving average. The context encoder is given only the context graph $G_x$ and produces a representation for each context atom,
$$
\mathbf{s}_x=f_\theta(G_x)=\{\mathbf{s}_{x_i}\}_{i\in \mathcal{I}_x}.
$$
In contrast, the target encoder operates on the full atomic graph $G$. It therefore produces representations for all $N$ atoms,
$$
\tilde{\mathbf{s}}_y=f_{\bar{\theta}}(G)=\{\tilde{\mathbf{s}}_{y_i}\}_{i=1}^{N}.
$$
After encoding the full graph, only the representations corresponding to atoms in the target subgraph are retained as prediction targets, 
$$
\mathbf{s}_y=\{\tilde{\mathbf{s}}_{y_i}\}_{i\in \mathcal{I}_y}.
$$
Thus, although the learning objective is defined only over atoms belonging to the target graph $G_y$, their target representations are computed from the full atomic system. Each target representation can therefore incorporate information from the surrounding atomic environment through the message-passing operations of the target encoder, rather than being restricted to information contained within $G_y$ alone. We argue that this provides richer target representations that encode the target atoms in the context of the complete atomic system.

\subsection{Prediction}
Given the context representations $\mathbf{s}_x$, a shallow graph neural network $g_\phi$ predicts representations produced by the target encoder. The predictor consists of a GNN body followed by objective-specific readout heads. Since the GNN body of the predictor operates on the atom features produced by the context encoder it does not contain an embedding layer. We consider two prediction objectives: a substructure objective that predicts a global representation of the target subgraph and an atom-level objective that predicts the representation of each individual target atom. First, the GNN body of the predictor produces further refined context-atom features $\mathbf{h}_x$, which are then mean-pooled and passed to the respective readout heads:
$$
\mathbf{h}_x=g_\phi^{\text{GNN}}(G_x,\mathbf{s}_x), \quad \bar{\mathbf{h}}_x=\frac{1}{|\mathcal{I}_x|}\sum_{i\in \mathcal{I}_x}\mathbf{h}_{x_i}.
$$
\subsubsection{Substructure Target Prediction}
In the substructure objective, the target is the mean-pooled invariant scalar features produced by the target encoder indexed over the target subgraph

$$
\bar{\mathbf{s}}_y=\frac{1}{|\mathcal{I}_y|}\sum_{i\in \mathcal{I}_y}\mathbf{s}_{y_i}^{(0)},
$$
and the substructure readout head is a two-layer MLP with SiLU activation applied to the invariant component of the pooled context
$$
\hat{\mathbf{s}}_{y}=R^{\text{sub}}_\phi(\mathbf{\bar{h}}_x)=\text{MLP}(\bar{\mathbf{h}}_x^{(0)}),
$$
mapping into the feature space of the target encoder.

\subsubsection{Atom-level Target Prediction}
\label{app:atom_representation_construction}
The atom-level objective instead produces one prediction $\hat{\mathbf{s}}_{y_i}$ for every atom $i\in \mathcal{I}_y$, which is compared against the corresponding target encoder representation $\mathbf{s}_{y_i}$. In addition to the context features, the atom-level readout head is provided with a positional encoding $\mathbf{q}_i$ of the target atom’s position relative to the context subgraph:
$$
\hat{\mathbf{s}}_{y_i}=R^\text{atom}_\phi(\mathbf{h}_x,\mathbf{q}_i), \quad i\in \mathcal{I}_y.
$$
We construct the positional encoding so that it has the same degree structure as the atom features, with each $\mathbf{q}^{(l)}_i$ transforming under rotation exactly as $\mathbf{h}_x^{(l)}$ does. For each degree $l\ge1$ we then take the inner product of the two over the $2l+1$ components seperately in each feature channel and concatenate them with the invariant parts to form
$$
\mathbf{u}_i=\left[\bar{\mathbf{h}}_x^{(0)}\|\mathbf{q}_i^{(0)}\|\langle\bar{\mathbf{h}}^{(1)}_x,\mathbf{q}^{(1)}_i\rangle\|\cdots\|\langle\bar{\mathbf{h}}^{(L_\text{max})}_x,\mathbf{q}^{(L_\text{max})}_i\rangle\right] \in \mathbb{R}^{C(L_\text{max}+2)},
$$
where $\cdot\|\cdot$ denotes concatenation and $\langle\bar{\mathbf{h}}_x^{(l)},\mathbf{q}_i^{(l)}\rangle\in \mathbb{R}^C$ denote the channel-wise inner product, with $c$-th entry
$$
\langle\bar{\mathbf{h}}_x^{(l)},\mathbf{q}^{(l)}_i\rangle_c=\sum_{m=-l}^{l}\bar{h}^{(l)}_{x,mc}q_{i,mc}^{(l)}.
$$
The invariant term $\bar{\mathbf{h}}^{(0)}_i$ carries only the content of the context, the term $\mathbf{q}^{(0)}_i$ depends only on the distance to the target atom, and the remaining terms couple the directional components of the context to the direction of the query. A two-layer MLP maps this feature into the target encoder’s feature space
$$
\hat{\mathbf{s}}_{y_i}=\text{MLP}(\mathbf{u}_i).
$$
\subsubsection{Equivariant Relative Positional Encoding}
\label{app:equivariant_relative_positions}
We encode target positions relative to the context using a radial-spherical-harmonic construction following Tensor Field Networks~\citep{thomas2018tensor}, with distances expanded in a Gaussian radial basis as in SchNet~\citep{schutt2017schnet}.

For molecules, we define the position of each target atom $i$ relative to the context graph. First, the centroid of the context atoms is computed as $\mathbf{r}_x=\frac{1}{|\mathcal{I}_x|}\sum_{j\in \mathcal{I}_x}\mathbf{r}_j.$ 
The relative displacement of target atom $i$ from the context centroid is then $\mathbf{r}_{xi}=\mathbf{r}_i-\mathbf{r}_x$. We expand the distance $\|\mathbf{r}_{xi}\|$ in a Gaussian radial basis $\mathbf{\varphi}(\|\mathbf{r}_{xi}\|)\in \mathbb{R}^K$, with 
$$
\varphi_k(\|\mathbf{r}_{xi}\|)=\exp\left(\frac{-(\|\mathbf{r}_{xi}\|-\mu_k)^2}{2\sigma^2}\right),
$$
where the centers $\mu_k$ are spaced over $[0,r_q]$. The direction $\hat{\mathbf{r}}_{xi}=\mathbf{r}_{xi}/\|\mathbf{r}_{xi}\|$ is encoded using spherical harmonics, giving
$$
q^{(l)}_{i,mc}=Y^{(l)}_m(\hat{\mathbf{r}}_{xi})(\mathbf{W}^{(l)}\varphi(\|\mathbf{r}_{xi}\|)+\mathbf{b}^{(l)})_c,
$$
where $l=0,\dots,L_\text{max}, |m|\le l, c=1,\dots C$. The parameters $\mathbf{W}^{(l)}\in \mathbb{R}^{C\times K}, \mathbf{b}^{(l)}\in \mathbb{R}^C$ are a separate learned linear layer for each degree. 

For crystals, the relative displacement $\mathbf{r}_{xi}$ is defined differently to account for periodic boundary conditions. The centroid of the context atoms is not well defined from the coordinates of a unit cell, since atoms that are spatially close may lie on opposite sides of a periodic boundary. We therefore define a reference atom $c_x\in \mathcal{I}_x$ for the context graph. When the context graph is the sampled $k$-EgoNet $c_x=c$ , while when the complementary subgraph forms the context, $c_x$ is sampled uniformly from $\mathcal{I}_x$. The relative displacement of target atom $i$ is then defined using its closest periodic image with respect to $c_x$
$$
\mathbf{r}_{xi}=\mathbf{r}_i+\mathbf{n}^*\mathbf{L}-\mathbf{r}_{c_x}, \quad \mathbf{n}^*=\argmin_{n\in \mathbb{Z}^3}\|\mathbf{r}_i+\mathbf{n}\mathbf{L}-\mathbf{r}_{c_x}\|.
$$
Thus, for crystals $\mathbf{r}_{xi}$ is the minimum-image displacement from the context reference atom $c_x$ to target atom $i$. The resulting displacement is then encoded in the same way as in the molecular case.

\subsubsection{Loss}
We use mean squared error between predicted and target representations, $\ell(\mathbf{a},\mathbf{b}) = \frac{1}{C}\|\mathbf{a}-\mathbf{b}\|_2^2,$ where $C$ is the feature dimension. For the substructure objective, the loss is averaged over the $B$ graphs in the batch
$$
\mathcal{L}_{\mathrm{sub}}^{x\rightarrow y} = \frac{1}{B} \sum_{g=1}^{B} \ell\left( \hat{\mathbf{s}}_{y}^{(g)}, \bar{\mathbf{s}}_{y}^{(g)} \right),
$$
where $x\rightarrow y$ denotes the prediction direction in which $G_x$ is used as the context graph and $G_y$ as the target graph. For the atom-level objective, the loss is averaged over all target atoms across all graphs in the batch
$$
\mathcal{L}_{\mathrm{atom}}^{x\rightarrow y} = \frac{1}{\sum_{g=1}^{B}|\mathcal{I}_y^{(g)}|} \sum_{g=1}^{B} \sum_{i\in\mathcal{I}_y^{(g)}} \ell\left( \hat{\mathbf{s}}_{y_i}, \mathbf{s}_{y_i} \right).
$$
Both objectives are evaluated symmetrically in both directions, with the two complementary subgraphs alternating as context and target. Thus, for $o\in\{\mathrm{sub},\mathrm{atom}\}$,
$$
\mathcal{L}_o = \frac{1}{2} \left( \mathcal{L}_o^{x\rightarrow y} + \mathcal{L}_o^{y\rightarrow x} \right),
$$
and the complete loss is $\mathcal{L} = \lambda_{\mathrm{sub}}\mathcal{L}_{\mathrm{sub}} + \lambda_{\mathrm{atom}}\mathcal{L}_{\mathrm{atom}}$. We train models with each objective individually and jointly, setting $\lambda_{\mathrm{pool}}=\lambda_{\mathrm{atom}}=1$.

\subsection{Data Curation}
\label{app:data_curation}
We use structures at or near local energy minima because we expect their geometry to help constrain the identity of the target sub-structure and atom. The intuition is that, in a relaxed structure, the atomic arrangement must be compatible with the interactions between the constituent atom types. Strong distortions of the system can weaken this connection, making the target embedding of a specific atom \textit{significantly} harder to predict from the context $\mathcal{I}_x$. Restricting the data to near-minimum structures should reduce such ambiguity, even though several atom types could still be compatible with the same context. An alternative to this approach would be to include off-equilibrium structures and then condition the model on the system’s potential energy, as this would likely also help constrain the learning problem. 

\section{Limitations}
Throughout our experiments, we have used the Equiformer V3 architecture for all tasks. While we do observe a strong effect from using our SSL objective, we do not have a direct comparison to other SSL objectives with the same architecture. For ADMET datasets, other SSL objectives are not directly translatable due to their reliance on chemical bond encoding, whereas our SSL objective relies on the 3D structure of molecules. An additional nuance to this, is that Atom-JEPA is the only ADMET method tested that explicitly uses a 3D structure as input from conformers. We hypothesise that in addition to the Atom-JEPA loss, the ADMET performance could be driven by the strong implicit biases present in Equiformer V3 such as rotational equivariance or the data augmentation benefit of generating multiple conformers. 

We found a strong influence from the conformer choices as highlighted in Appendix \ref{appendix:conformer_influence}. Notably, the number of conformers used during training and averaging the model prediction over multiple conformers improve prediction accuracy. At the same time the quality of the conformer (as judged by the energy), seems to not have an impact on the performance. This suggest that it is rather the diversity of the conformers that improve the model not better conformers. This could be a problem when using weak conformer generation methods such as ours (see Appendix \ref{appendix:conformer_generation}), as their strong biases restrict the true conformer space. We have not investigated more accurate conformer generation strategies through higher order theory here.

ADMET tasks are notoriously difficult to benchmark correctly, due to their heavy generalization requirements from non-uniformly random splits, small test samples, and noisy labels. To asses optimal performance for each model, their hyperparameters would have been tuned through 2-fold cross-validation. It would have been computationally infeasible to do this for finetuning of Atom-JEPA and costly to do for all the MLP baselines. While it would be possible to do 2-fold cross-validation for the LGBT models, we opted to use our shared hyperparameter setup consistent across all models to make cross-model comparisons more fair.

\section{Datasets}
\label{app:datasets}
Table \ref{tab:datasets} summarizes the datasets used for large-scale pretraining and downstream evaluation. In addition to its use as a downstream benchmark, we use QM9 for a separate set of pretraining ablations of the loss objectives. 

We subsample the Alexandria PBE dataset by retaining only near-stable relaxed crystal structures with an energy above the convex hull of at most 0.05 eV/atom and containing between 2 and 100 atoms per unit cell. These filtering criteria reduce the dataset from 5,952,515 to 1,696,344 structures.

The Uni-Mol dataset contains 10 3D conformers per molecule. We use only a single conformer per molecule (the first), resulting in 18,837,320 training samples. 

\begin{table}[H]
\centering
\caption{
    \textbf{Summary of datasets.}
    Uni-Mol and Alexandria are used for large-scale pretraining.
    The remaining datasets are used for downstream evaluation.
    QM9 is additionally used for pretraining in the separate ablation experiments.
}
\label{tab:datasets}
\small
\setlength{\tabcolsep}{5pt}

\resizebox{\columnwidth}{!}{%
\begin{tabular}{lrrl}
\toprule
\textbf{Dataset} & \textbf{Samples} & \textbf{Tasks} & \textbf{Description} \\
\midrule
\multicolumn{4}{l}{Large-scale pretraining datasets} \\
\midrule
Uni-Mol~\citep{zhou2023unimol}
    & 18,837,320 & -- & 3D molecular conformers \\
Alexandria PBE~\citep{Cavignac2026Alexandria}
    & 1,696,344 & -- & DFT-relaxed inorganic crystals \\

\midrule
\multicolumn{4}{l}{Downstream evaluation datasets} \\
\midrule
ChEMBL-MT~\citep{DBLP:journals/corr/abs-2510-12719}
    & 114,000 & 25 & Multi-task ADMET \\
ExpansionRx~\citep{macdermott_opeskin_2026_21784075}
    & 7,608 & 10 & Multi-task ADMET \\
Biogen ADME~\citep{10.1021/acs.jcim.3c00160}
    & 3,521 & 6 & Multi-task ADME \\
TDC-ADMET~\citep{DBLP:conf/nips/HuangFG0RLCXSZ21}
    & 475--13,130 & 22 & Individual-task ADMET \\
QM9~\citep{qm9}
    & 130,831 & 12 & Quantum-chemical properties \\
Matbench~\citep{matbench}
    & 1,265--132,752 & 8 & Materials-property prediction \\
\bottomrule
\end{tabular}%
}
\end{table}

\subsection{Structural Overlap}
\label{app:data_leakage}
Due to the size of Unimol and Alexandria, some of the structures used for pretraining of Atom-JEPA are also present in datasets of ADMET, QM9, and Matbench datasets. While self-supervised training has not been observed to bias accuracy noticeably~\citep{DBLP:journals/corr/abs-2606-11508}, we still report the overlaps for transparency of the test sets for ADMET tasks in \cref{tab:unimol-evaluation-overlap}, full datasets of QM9 in \cref{tab:qm9-unimol-overlap}, and full dataset of Matbench in \cref{tab:matbench-alexandria-overlap}. For most ADMET endpoints, Unimol contains somewhere around $35\%-75\%$ of the structures of molecules seen in the endpoints. Due to the temporal split of ExpansionRX, only 1 test point is found in Unimol and at most $2\%$ of the molecules have the same scaffold. We found that Atom-JEPA's improvement over  other models is the largest on specifically ExpansionRX, but that the accuracy cannot be attributed to structural overlap between the pretraining data and ExpansionRX. For QM9 the overlap is also quite small at $4\%$. For Matbench the overlap to Alexandria varies for the endpoints. 

\begin{table}
\centering
\small
\caption{%
  \textbf{Uni-Mol pretraining-set overlap in downstream evaluation data.} Dataset rows are unions over their evaluation test membership. "Unique molecules" and "Found in Uni-Mol" use the Exact definition and with an observed label for that endpoint. Exact denotes identical RDKit canonical isomeric molecular graphs and retains stereochemistry, isotopes, formal charge, and disconnected fragments. Parent compares cleaned fragment parents, ignoring salts and solvents while retaining stereochemistry. Broad parent also normalizes charge and isotopes, removes stereochemistry, and uses a heteroatom-tautomer hash. Murcko scaffold is the percentage of unique Bemis-Murcko scaffolds observed in Uni-Mol and measures structural relatedness. Percentage rounded to nearest percentage.
}
\label{tab:unimol-evaluation-overlap}
\setlength{\tabcolsep}{5pt}
\resizebox{\columnwidth}{!}{%
\begin{tabular}{lrrrrrr}
\toprule
\textbf{Evaluation dataset} & \textbf{Unique molecules} & \textbf{Found in Uni-Mol} & \multicolumn{4}{c}{\textbf{Overlap (\%)}} \\
\cmidrule(lr){4-7}
 & & & \textbf{Exact} & \textbf{Parent} & \textbf{Broad parent} & \textbf{Murcko scaffold} \\
\midrule
\textbf{ChEMBL-MT} & 23,213 & 11,964 & 52 & 52 & 73 & 90 \\
\quad CL Microsome Human & 968 & 632 & 65 & 67 & 73 & 88 \\
\quad CL Microsome Mouse & 136 & 84 & 62 & 66 & 69 & 92 \\
\quad CL Microsome Rat & 341 & 226 & 66 & 66 & 69 & 87 \\
\quad CL Total Dog & 90 & 54 & 60 & 71 & 94 & 97 \\
\quad CL Total Human & 241 & 150 & 62 & 71 & 94 & 97 \\
\quad CL Total Monkey & 42 & 21 & 50 & 55 & 95 & 97 \\
\quad CL Total Rat & 124 & 72 & 58 & 65 & 94 & 95 \\
\quad CYP2C8 Inhibition & 53 & 28 & 53 & 55 & 60 & 77 \\
\quad CYP2C9 Inhibition & 515 & 264 & 51 & 53 & 60 & 81 \\
\quad CYP2D6 Inhibition & 449 & 252 & 56 & 60 & 67 & 88 \\
\quad CYP3A4 Inhibition & 1,084 & 615 & 57 & 58 & 64 & 79 \\
\quad Fraction unbound plasma (Dog) & 50 & 23 & 46 & 62 & 96 & 95 \\
\quad Fraction unbound plasma (Human) & 603 & 274 & 45 & 49 & 83 & 74 \\
\quad Fraction unbound plasma (Monkey) & 24 & 10 & 42 & 50 & 96 & 95 \\
\quad Fraction unbound plasma (Rat) & 70 & 40 & 57 & 67 & 96 & 95 \\
\quad Papp Caco2 & 1,216 & 495 & 41 & 42 & 74 & 90 \\
\quad Pgp (Human) & 419 & 296 & 71 & 73 & 77 & 93 \\
\quad hERG binding & 757 & 322 & 43 & 44 & 75 & 90 \\
\quad LogD pH 7.4 & 763 & 516 & 68 & 72 & 79 & 94 \\
\quad LogSaq Kinetic  & 11,350 & 6,528 & 58 & 58 & 82 & 94 \\
\quad LogSaq Thermo & 6,750 & 2,835 & 42 & 43 & 59 & 86 \\
\quad VDss Dog & 87 & 53 & 61 & 71 & 94 & 97 \\
\quad VDss Human & 245 & 152 & 62 & 71 & 94 & 97 \\
\quad VDss Monkey & 41 & 20 & 49 & 54 & 95 & 97 \\
\quad VDss Rat & 111 & 61 & 55 & 63 & 95 & 95 \\
\addlinespace[3pt]
\textbf{ExpansionRx} & 2,282 & 1 & 0 & 0 & 0 & 2 \\
\quad LogD & 2,270 & 1 & 0 & 0 & 0 & 2 \\
\quad KSOL & 2,170 & 1 & 0 & 0 & 0 & 1 \\
\quad HLM CLint & 782 & 0 & 0 & 0 & 0 & 1 \\
\quad MLM CLint & 1,170 & 0 & 0 & 0 & 0 & 1 \\
\quad Caco-2 Permeability Papp A\textgreater{}B & 1,616 & 0 & 0 & 0 & 0 & 2 \\
\quad Caco-2 Permeability Efflux & 1,616 & 0 & 0 & 0 & 0 & 2 \\
\quad MPPB & 454 & 0 & 0 & 0 & 0 & 1 \\
\quad MBPB & 451 & 0 & 0 & 0 & 0 & 1 \\
\quad MGMB & 209 & 0 & 0 & 0 & 0 & 0 \\
\addlinespace[3pt]
\textbf{Biogen ADME} & 1,830 & 907 & 50 & 50 & 63 & 83 \\
\quad Log HLM CLint (mL/min/kg) & 1,620 & 809 & 50 & 50 & 63 & 83 \\
\quad Log MDR1-MDCK & 1,384 & 703 & 51 & 51 & 63 & 83 \\
\quad Log SOLUBILITY PH 6.8 (ug/mL) & 1,097 & 508 & 46 & 47 & 60 & 83 \\
\quad Log Plasma Protein Binding Human & 120 & 53 & 44 & 44 & 64 & 75 \\
\quad Log Plasma Protein Binding Rat & 93 & 42 & 45 & 45 & 67 & 81 \\
\quad Log RLM CLint (mL/min/kg) & 1,596 & 804 & 50 & 51 & 64 & 84 \\
\addlinespace[3pt]
\textbf{TDC-ADMET} & 13,816 & 7,406 & 54 & 56 & 71 & 89 \\
\quad AMES & 1,453 & 513 & 35 & 36 & 51 & 73 \\
\quad BBB Martins & 394 & 248 & 63 & 72 & 84 & 93 \\
\quad Bioavailability Ma & 128 & 89 & 70 & 83 & 94 & 99 \\
\quad Caco2 Wang & 181 & 67 & 37 & 39 & 66 & 86 \\
\quad Clearance Hepatocyte AZ & 203 & 142 & 70 & 73 & 82 & 94 \\
\quad Clearance Microsome AZ & 221 & 155 & 70 & 72 & 76 & 94 \\
\quad CYP2C9 Substrate Carbon-Mangels & 135 & 89 & 66 & 69 & 88 & 96 \\
\quad CYP2C9 Veith & 2,419 & 1,541 & 64 & 66 & 82 & 93 \\
\quad CYP2D6 Substrate Carbon-Mangels & 134 & 93 & 69 & 77 & 92 & 100 \\
\quad CYP2D6 Veith & 2,626 & 1,707 & 65 & 67 & 84 & 94 \\
\quad CYP3A4 Substrate Carbon-Mangels & 134 & 90 & 67 & 75 & 92 & 97 \\
\quad CYP3A4 Veith & 2,467 & 1,603 & 65 & 67 & 82 & 93 \\
\quad DILI & 96 & 59 & 61 & 70 & 96 & 99 \\
\quad Half-life Obach & 134 & 84 & 63 & 78 & 93 & 96 \\
\quad hERG & 128 & 35 & 27 & 32 & 73 & 74 \\
\quad HIA Hou & 117 & 60 & 51 & 55 & 84 & 97 \\
\quad LD50 Zhu & 1,466 & 538 & 37 & 38 & 53 & 84 \\
\quad Lipophilicity AstraZeneca & 840 & 601 & 72 & 74 & 78 & 90 \\
\quad Pgb Broccatelli & 245 & 137 & 56 & 60 & 82 & 90 \\
\quad PPBR AZ & 343 & 241 & 70 & 71 & 79 & 96 \\
\quad Solubility AqSolDB & 1,995 & 762 & 38 & 41 & 57 & 84 \\
\quad VDss Lombardo & 221 & 45 & 20 & 25 & 89 & 67 \\
\bottomrule
\end{tabular}%
}
\end{table}
\begin{table}[H]
\centering
\small
\caption{%
  \textbf{Full QM9 overlap with Uni-Mol pretraining data.} Exact denotes an identical canonical isomeric molecular graph. Parent removes salts and solvents while retaining stereochemistry. Broad parent additionally normalizes charge, isotopes, stereochemistry, and common heteroatom tautomers. Murcko scaffold measures structural relatedness rather than molecule identity and excludes acyclic molecules from its denominator. All QM9 properties share this molecular inventory. The graph-based profiles exclude 1,819 PyG QM9 entries whose explicit-hydrogen SMILES cannot be sanitized by RDKit; Unique molecules therefore denotes unique valid graph identities.
}
\setlength{\tabcolsep}{5pt}

\resizebox{\columnwidth}{!}{%
\begin{tabular}{lrrrrrr}
\toprule
\textbf{Evaluation dataset} & \textbf{Unique molecules} & \textbf{Found in Uni-Mol} & \multicolumn{4}{c}{\textbf{Overlap (\%)}} \\
\cmidrule(lr){4-7}
 & & & \textbf{Exact} & \textbf{Parent} & \textbf{Broad parent} & \textbf{Murcko scaffold} \\
\midrule
\textbf{QM9} & 128,822 & 4,742 & 4 & 4 & 5 & 9 \\
\bottomrule
\end{tabular}%
}
\label{tab:qm9-unimol-overlap}
\end{table}

\begin{table}[H]
\centering
\small
\setlength{\tabcolsep}{5pt}
\caption{%
  \textbf{Matbench overlap with Alexandria PBE pretraining data.} Strict requires the same elemental composition and a tight primitive-cell match without lattice rescaling. Relaxed uses lattice rescaling and moderate geometric tolerances to accommodate relaxation differences. Prototype permits element substitution but requires the same canonical anonymous formula, Pearson symbol, space group, and Wyckoff sequence. Composition requires only the same reduced elemental composition. The latter two columns measure structural or chemical relatedness, not identical materials.
}
\resizebox{\columnwidth}{!}{%
\begin{tabular}{lrrrrrr}
\toprule
\textbf{Evaluation dataset} & \textbf{Unique structures} & \textbf{Found in Alexandria} & \multicolumn{4}{c}{\textbf{Overlap (\%)}} \\
\cmidrule(lr){4-7}
 & & & \textbf{Strict} & \textbf{Relaxed} & \textbf{Prototype} & \textbf{Composition} \\
\midrule
\textbf{Matbench} & 151,731 & 59,032 & 39 & 44 & 76 & 59 \\
\quad Phonons & 1,265 & 1,190 & 94 & 96 & 100 & 99 \\
\quad Dielectric & 4,764 & 3,399 & 71 & 80 & 93 & 92 \\
\quad Log GVRH & 10,987 & 8,697 & 79 & 83 & 99 & 93 \\
\quad Log KVRH & 10,987 & 8,697 & 79 & 83 & 99 & 93 \\
\quad Perovskites & 18,928 & 26 & 0 & 0 & 64 & 5 \\
\quad MP Gap & 106,113 & 58,659 & 55 & 63 & 84 & 78 \\
\quad MP E Form & 132,752 & 58,959 & 44 & 51 & 78 & 67 \\
\quad MP Is Metal & 106,113 & 58,659 & 55 & 63 & 84 & 78 \\
\bottomrule
\end{tabular}%
}
\label{tab:matbench-alexandria-overlap}
\end{table}

\newpage
\section{Analysis of Learned Representations}
\label{app:learned-representations} 
Here we examine the learned representation space of Atom-JEPA using visualizations, nearest-neighbor analysis and probes on frozen features. First, for each of the two main models trained for molecules and crystals respectively, we randomly sample 50,000 structures from the corresponding pretraining dataset and mean-pool over the final-layer scalar atom features. For all embedding visualizations, we standardize the features and apply principal component analysis (PCA), retaining components that explain at least 90\% of the variance, followed by UMAP with 60 neighbors and a minimum distance of 0.4 to obtain a two-dimensional projection. We color crystals by crystallographic symmetry and molecules by physicochemical properties and structural features. We further assess structural similarity within embedding neighborhoods and use probes to quantify the structural, geometric and physical-property information captured beyond what can be predicted from atomic composition, system size and geometry descriptor baselines. For the molecular model, we extend the analysis to 32,252 molecules from the manually annotated three-star subset of ChEBI (Chemical Entities of Biological Interest)~\citep{chebi}, coloring their embeddings by chemical family. Finally, on QM9 molecules, we compare the pretrained molecular model with a randomly initialized counterpart through visualizations and probes of each QM9 target using final-layer and all-layer features, assessing Atom-JEPA pretraining’s contribution beyond the 3D equivariant GNN backbone’s inductive biases.

\subsection{Structure and Property Organization in the Embedding Space}
\label{app:structural-organization}

\subsubsection{Crystal Representations}
\begin{figure}[htbp]
    \centering
    \begin{subfigure}{\textwidth}
        \centering
        \includegraphics[
            width=0.75\linewidth,
            height=0.36\textheight,
            keepaspectratio
        ]{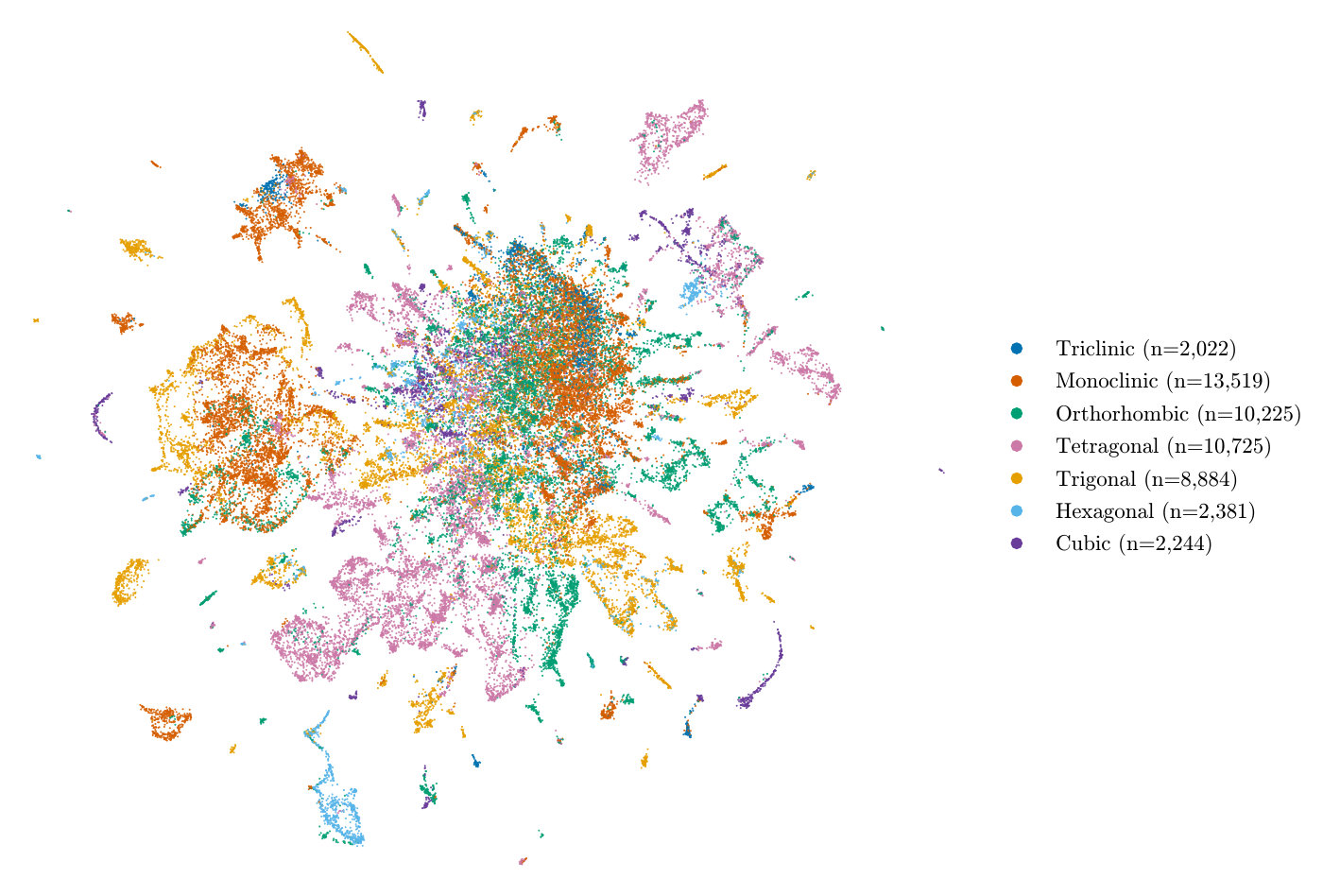}
        \caption{Combined crystal systems.}
        \label{fig:crystal-systems-combined}
    \end{subfigure}

    \medskip

    \begin{subfigure}{\textwidth}
        \centering
        \includegraphics[
            width=\linewidth,
            height=0.36\textheight,
            keepaspectratio
        ]{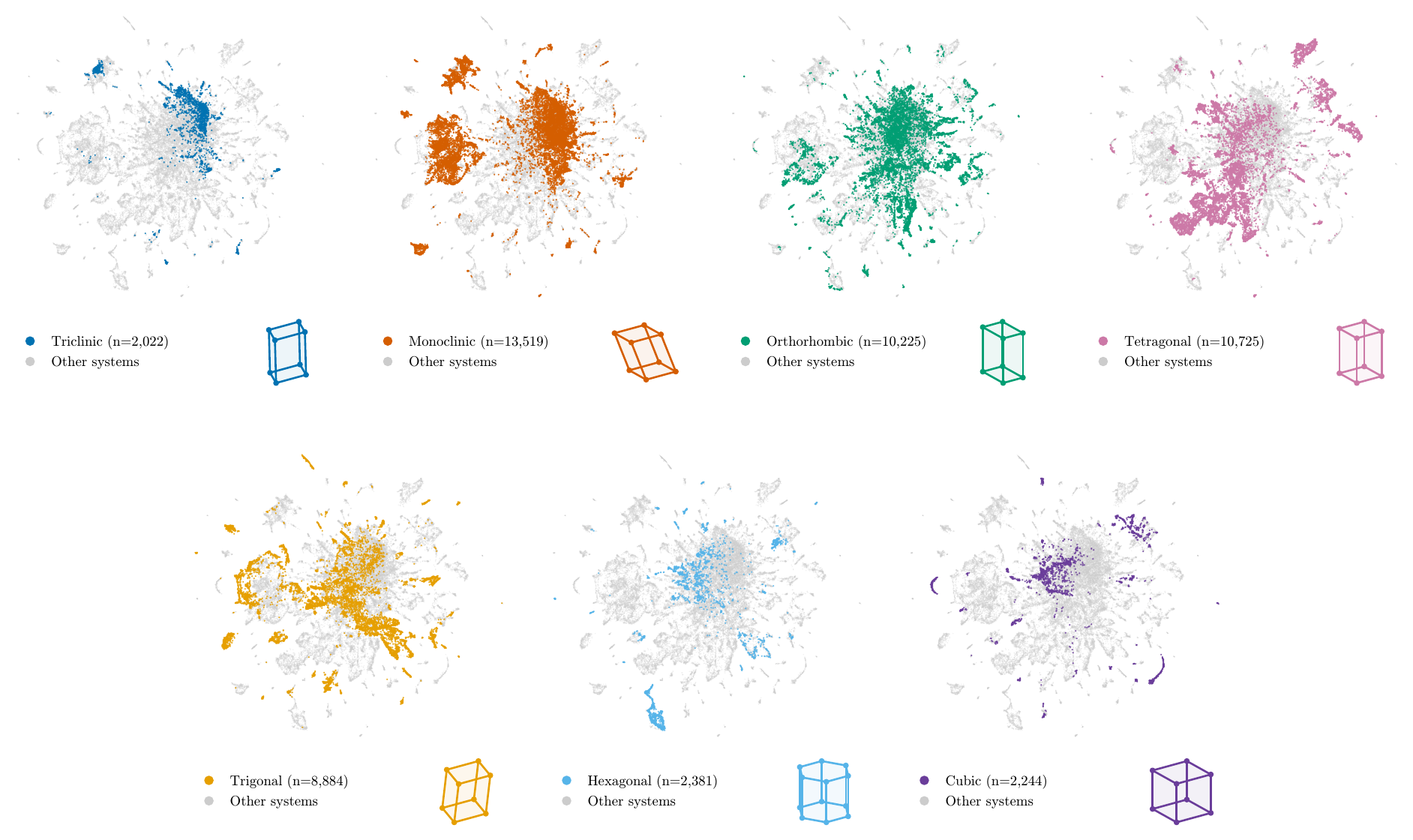}
        \caption{Each crystal system highlighted against the others in gray.}
        \label{fig:crystal-systems-individual}
    \end{subfigure}
    \caption{UMAP visualization of crystal representations by crystal system.}
    \label{fig:crystal-systems}
\end{figure}

\begin{figure}[htbp]
    \centering
    \begin{subfigure}{\textwidth}
        \centering
        \includegraphics[
            width=0.75\linewidth,
            height=0.36\textheight,
            keepaspectratio
        ]{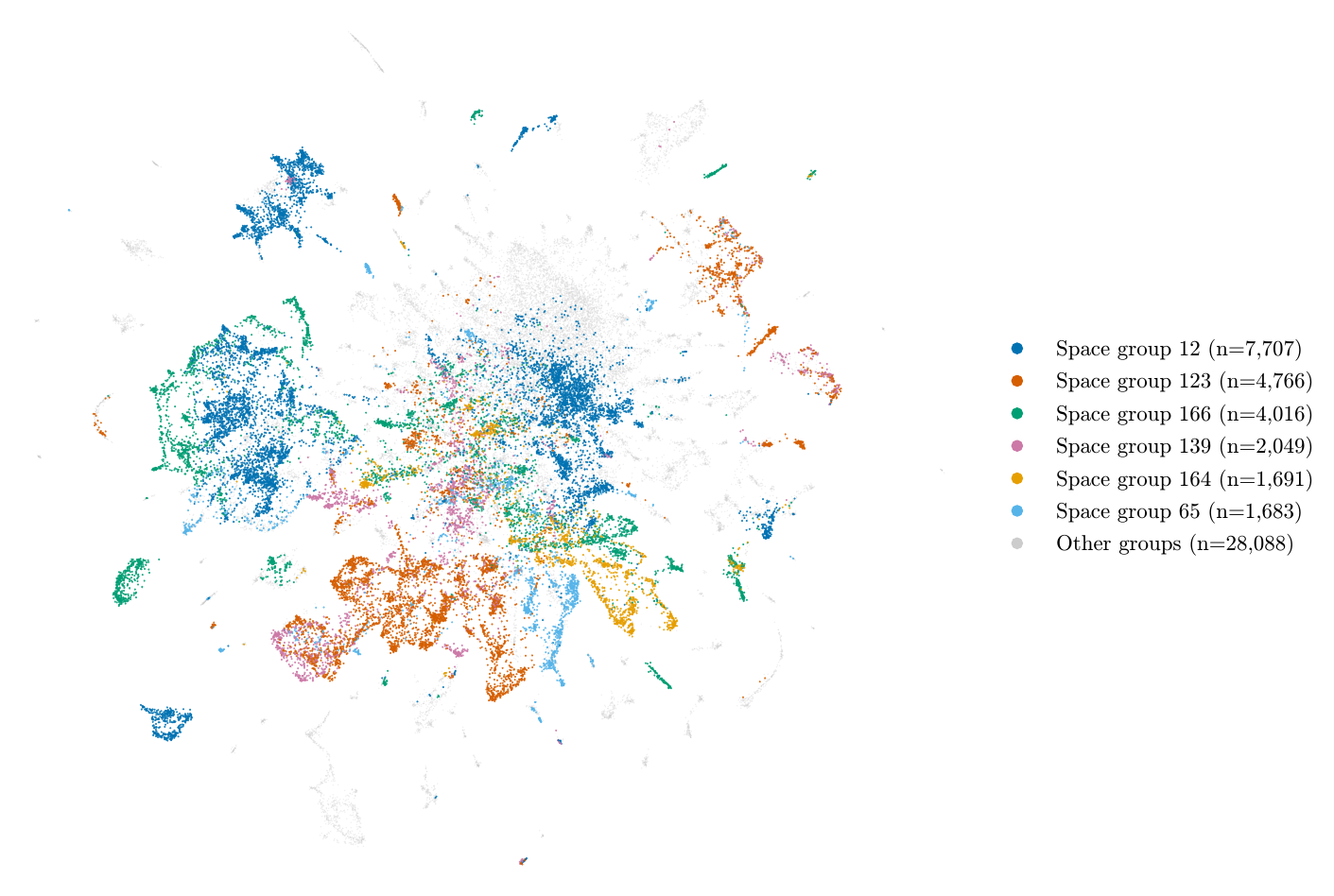}
        \caption{Combined space groups.}
        \label{fig:space-groups-combined}
    \end{subfigure}

    \medskip

    \begin{subfigure}{\textwidth}
        \centering
        \includegraphics[
            width=\linewidth,
            height=0.36\textheight,
            keepaspectratio
        ]{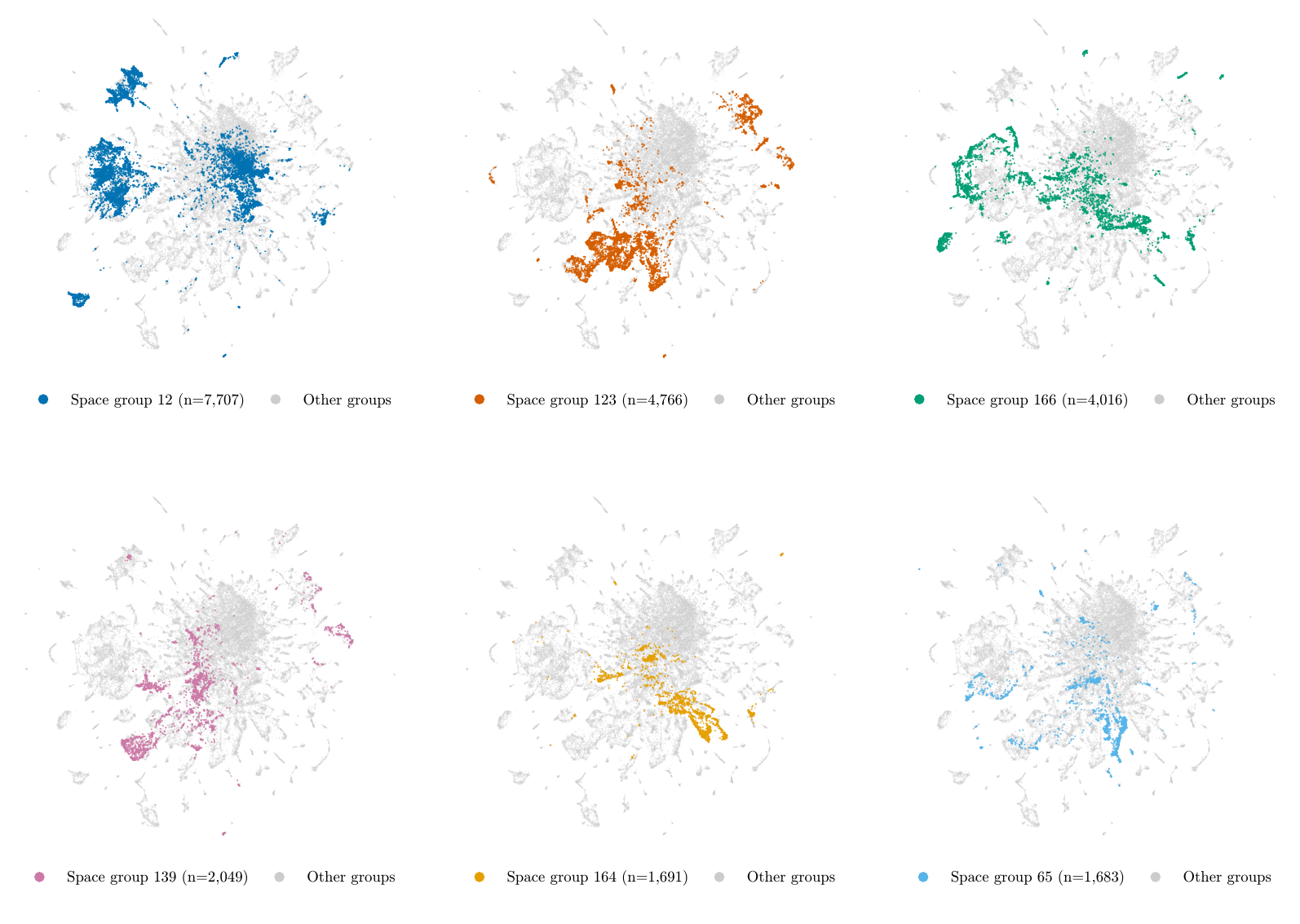}
        \caption{Each of the six most frequent space groups highlighted
        against the others in gray.}
        \label{fig:space-groups-individual}
    \end{subfigure}
    \caption{UMAP visualization of crystal representations by space group.}
    \label{fig:space-groups}
\end{figure}

Figures~\ref{fig:crystal-systems} and~\ref{fig:space-groups} visualize the Atom-JEPA features of 50,000 random samples from the Alexandria pretraining dataset. Figure~\ref{fig:crystal-systems} colors each crystal by its corresponding crystal system and Figure~\ref{fig:space-groups} colors by the six most frequent space groups in the sample.

To quantify structural organization in the embedding space, we evaluate the 10 nearest neighbors of 1,000 randomly selected samples using Euclidean distance in Table~\ref{tab:crystal-neighborhoods}. We compare these neighborhoods with nearest neighbors measured by composition (118-dimensional vector of elemental fractions), random crystals with matching atom count, and unrestricted random crystals. Embedding neighborhoods exhibit substantially greater space-group agreement than all three controls. Only 0.05\% of embedding-neighbor pairs in the sample share the same reduced composition (elemental counts divided by their greatest common divisor), indicating that this symmetry organization extends across different chemical compositions.

\begin{table}[htbp]
    \centering
    \small
    \caption{Structural neighborhood agreement.}
    \label{tab:crystal-neighborhoods}
    \begin{tabular}{@{}lr@{}}
        \toprule
        Neighbor selection & Same space group \\
        \midrule
        Embedding neighbors                       & 69.7\% \\
        Composition-only neighbors                & 14.4\% \\
        Random crystals with matching atom count  & 12.4\% \\
        Unmatched random crystals                 & 4.9\%  \\
        \bottomrule
    \end{tabular}
\end{table}

Next, we assess the information accessible through linear probes on the frozen features by predicting geometric quantities and physical properties. We partition the 50,000 sample into training, validation, and test sets containing 70\%, 15\%, and 15\% of the structures, respectively, with reduced compositions kept disjoint across partitions. We then fit ridge-regression probes with regularization selected by validation mean squared error. Features are standardized using training-set statistics, and performance is reported as $R^2$, the coefficient of determination on the test set. 

Table~\ref{tab:crystal-geometry} compares embeddings with a composition-and-size baseline for predicting log-volume per atom $\ln(V/N)$, where $V$ is the unit-cell volume and $N$ is the number of atoms in the cell as well as predicting mean coordination within 3 Å. The baseline represents each crystal using 122 features: 118 elemental fractions and four size features given by $N$, $\ln N$, $N^2$ and $N^3$ allowing the regression to model nonlinear relationships between cell size and the target. Coordination is calculated using periodic neighbours, including qualifying self-images. Combining embeddings with composition and size improves prediction of both quantities, indicating that the learned features include complementary geometric information that can be accessed by linear probes. While the embeddings alone slightly outperform the baseline for coordination, they perform worse for log-volume per atom. 

\begin{table}[htbp]
    \centering
    \small
    \caption{Test-set coefficient of determination $R^2$ for geometric quantities.}
    \label{tab:crystal-geometry}
    \begin{tabular}{@{}lcc@{}}
        \toprule
        Predictor
        & \shortstack{Log-volume-per-atom}
        & Coordination \\
        \midrule
        Composition + atom count & .942 & .794 \\
        Embeddings               & .854 & .814 \\
        Both together            & \textbf{.973} & \textbf{.894} \\
        \bottomrule
    \end{tabular}
\end{table}

Table~\ref{tab:phys-properties} reports linear-probe performance for energy above hull, formation energy, and indirect band gap with targets taken from the corresponding Alexandria record. We compare embeddings with composition-and-size baselines, additionally incorporating nine geometric descriptors that summarize cell geometry and local atomic environments. Embeddings alone outperform these baselines for energy above hull but underperform for formation energy and band gap. However, combining embeddings with composition, size, and geometry yields the highest test $R^2$ for all three targets. This suggests that the embeddings provide complementary predictive features, rather than consistently replacing explicit descriptors in the setting of linear probing. 

Formation energy is predicted most strongly, with the combined model improving $R^2$ from 0.861 to 0.908 over the baseline including geometry. Energy-above-hull prediction remains weak, despite achieving the largest relative improvement over that baseline: the combined model reaches $R^2=0.136$, which corresponds to an MAE of 11.7 meV/atom, compared with 13.2 meV/atom for a training-mean predictor. These results concern the near-stable subset used for pretraining, restricted to relaxed structures with energy above hull at most 0.05 eV/atom. The probe therefore assesses fine differences within the narrow 0–50 meV/atom interval, rather than discrimination across a broad range of crystal stabilities.

The nine geometric descriptors that we use are log-volume per atom; the ratio of the longest to shortest lattice-vector length; the root-mean-square cosine of the cell angles; the mean, standard deviation, minimum, and maximum coordination within 3 Å; and the mean and standard deviation of nearest-neighbor distances.

\begin{table}[htbp]
    \centering
    \small
    \caption{Test-set coefficient of determination $R^2$ for physical properties.}
    \label{tab:phys-properties}
    \begin{tabular}{@{}lrrr@{}}
        \toprule
        Predictor & Energy above hull & Formation energy & Indirect band gap \\
        \midrule
        Composition + size
            & .072 & .821 & .493 \\
        Composition + size + geometry
            & .079 & .861 & .507 \\
        Embeddings
            & .109 & .712 & .446 \\
        Embeddings + composition + size + geometry
            & \textbf{.136} & \textbf{.908} & \textbf{.542} \\
        \bottomrule
    \end{tabular}
\end{table}

\subsubsection{Molecular Representations}
\begin{figure}[htbp]
    \centering
    \includegraphics[
        width=0.65\linewidth,
        height=0.65\textheight,
        keepaspectratio
    ]{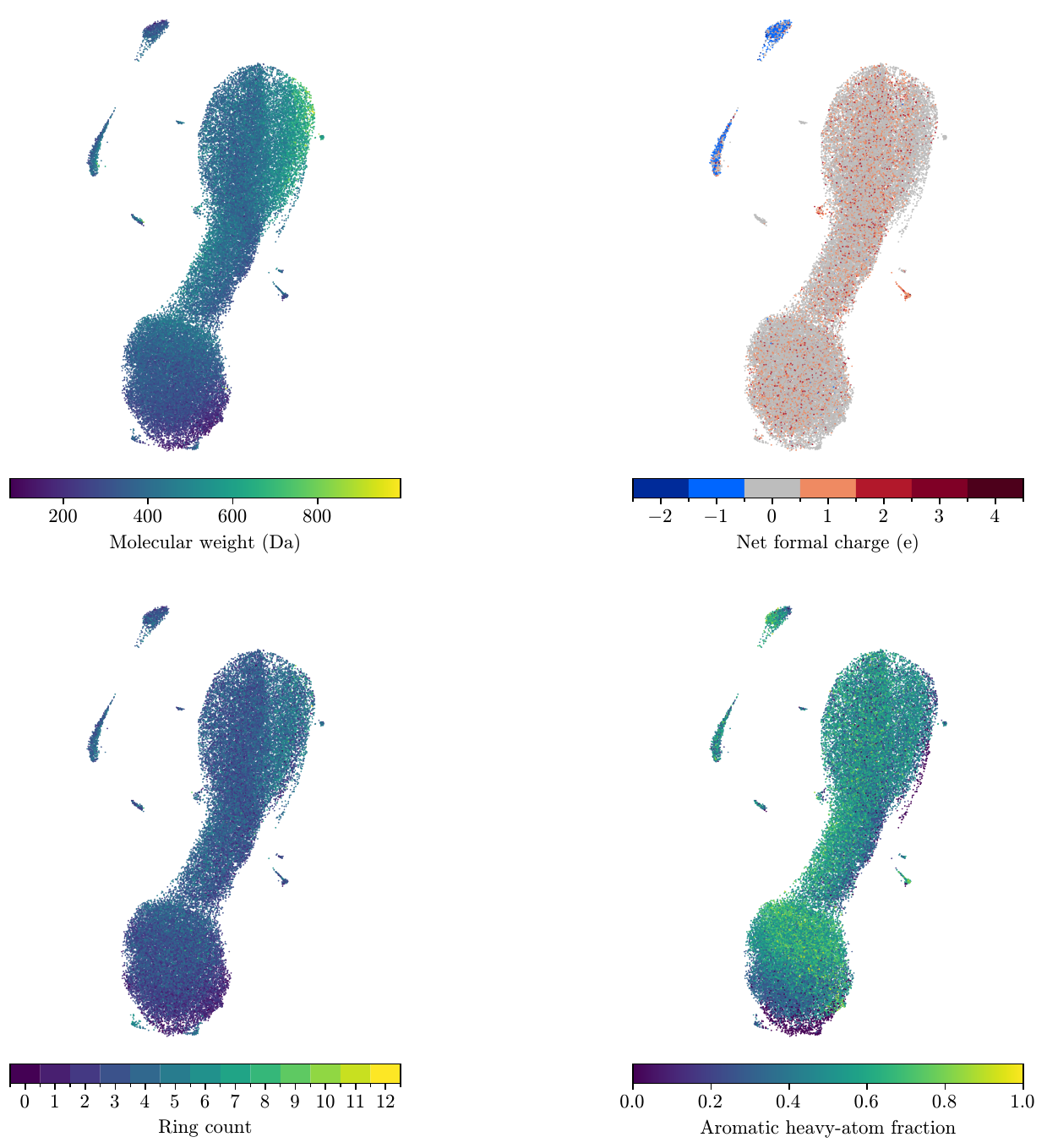}
    \caption{Uni-Mol representations colored
    by molecular weight, formal charge, ring count, and aromatic
    heavy-atom fraction.}
    \label{fig:unimol-descriptors}
\end{figure}

\begin{figure}[htbp]
    \centering
    \includegraphics[
        width=\linewidth,
        height=0.65\textheight,
        keepaspectratio
    ]{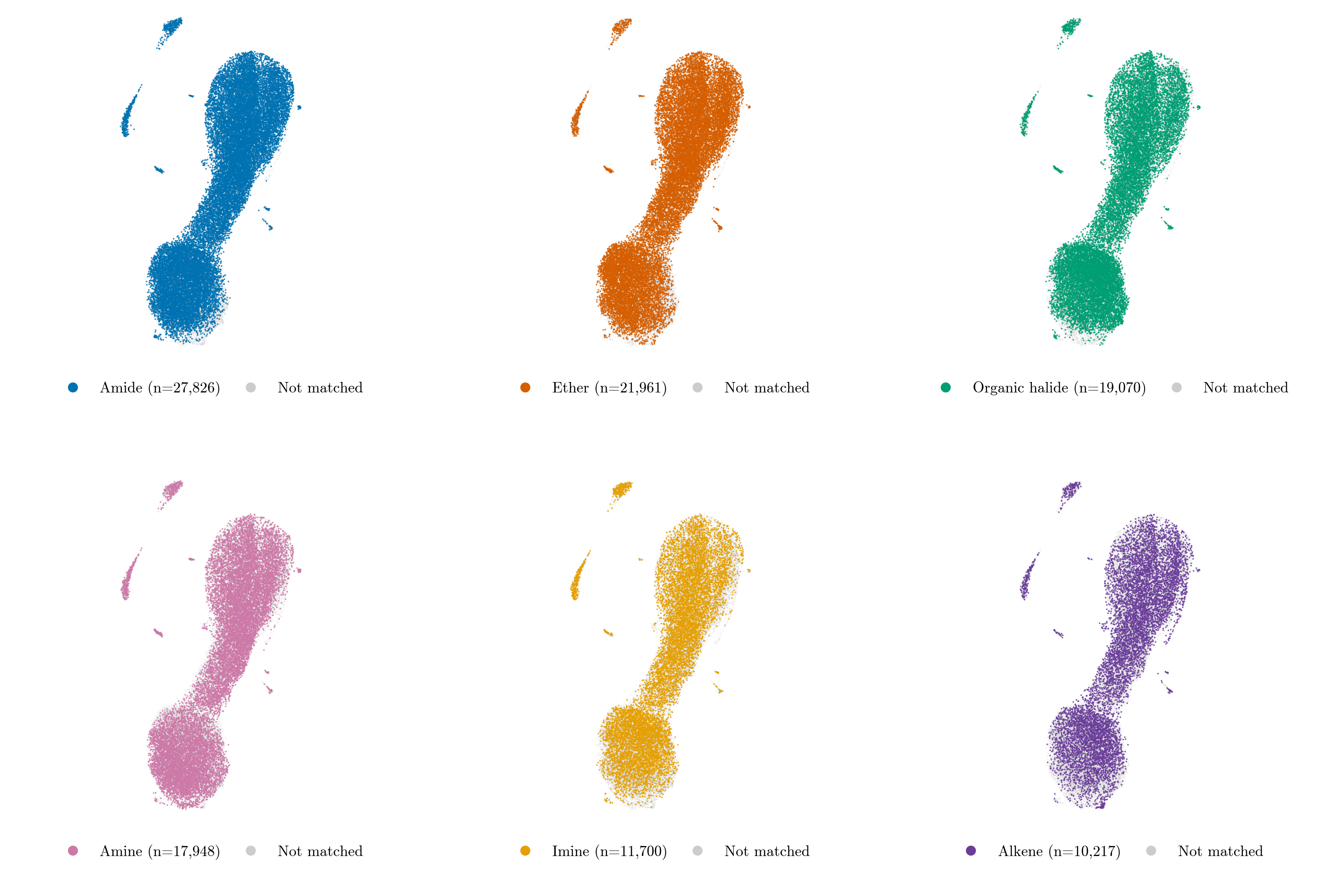}
    \caption{Uni-Mol representations with six functional groups
    highlighted individually.}
    \label{fig:unimol-functional-groups}
\end{figure}

\begin{figure}[htbp]
    \centering
    \begin{subfigure}[b]{\linewidth}
        \centering
        \includegraphics[
            width=\linewidth,
            height=0.35\textheight,
            keepaspectratio
        ]{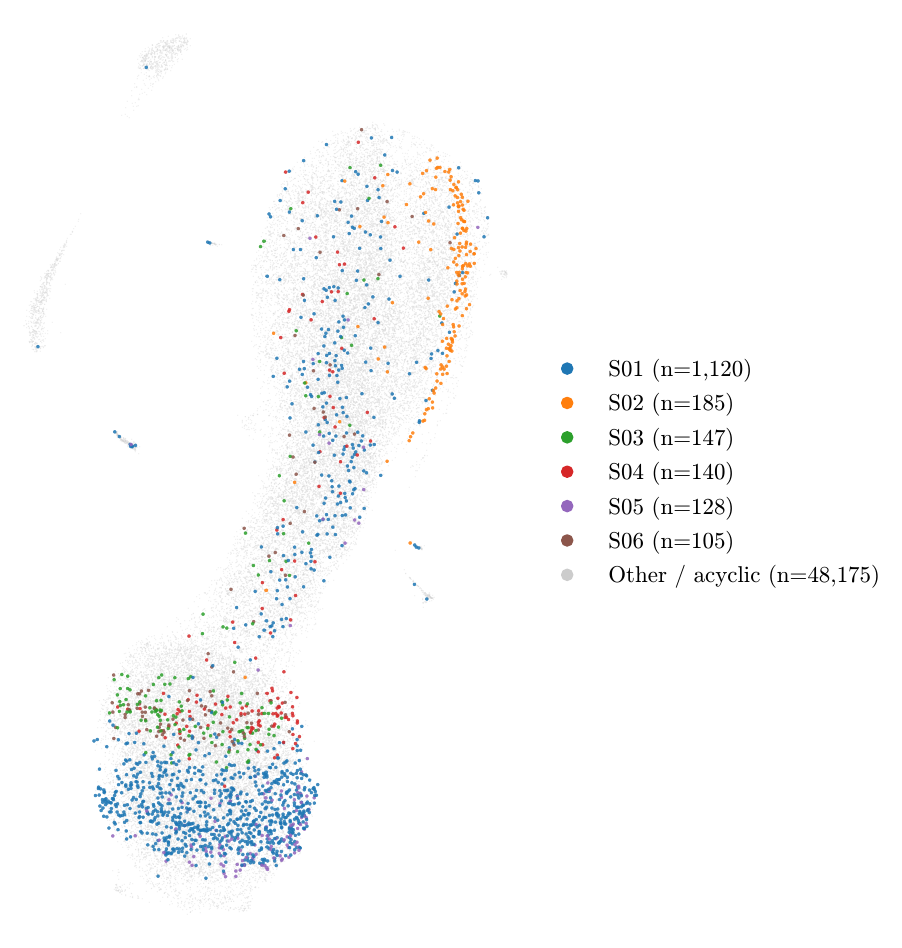}
        \caption{6 most frequent scaffolds combined}
        \label{fig:unimol_scaffold_combined}
    \end{subfigure}

    \vspace{1em}

    \begin{subfigure}[b]{\linewidth}
        \centering
        \includegraphics[
            width=\linewidth,
            height=0.35\textheight,
            keepaspectratio
        ]{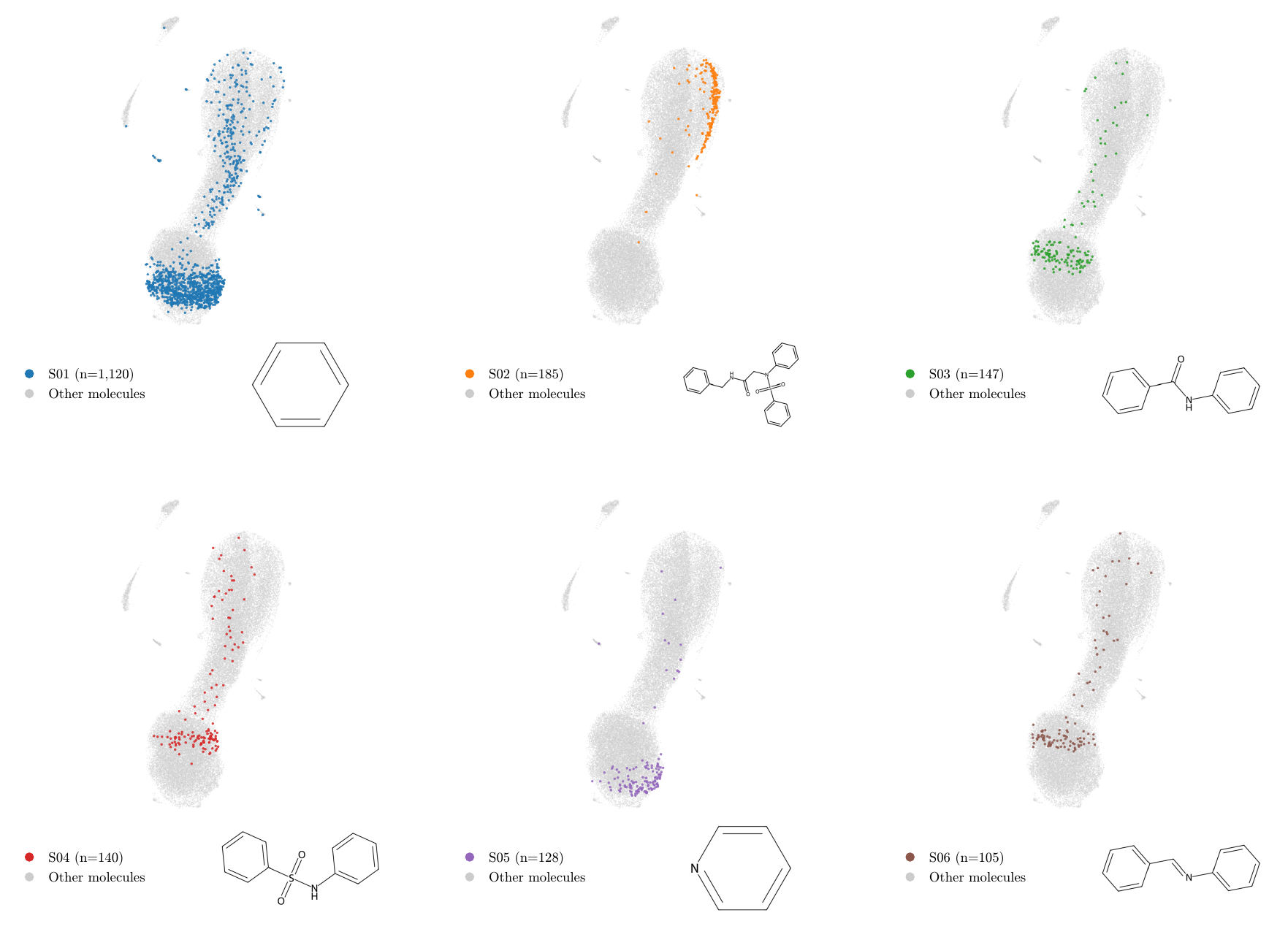}
        \caption{6 most frequent scaffolds highlighted individually}
        \label{fig:unimol_scaffold_individual}
    \end{subfigure}

    \caption{Uni-Mol representations with the 6 most frequent molecular scaffolds highlighted, shown combined and individually (remaining points in grey).}
    \label{fig:unimol_scaffold}
\end{figure}

Next, we examine molecular representations using 50,000 randomly sampled molecules from the Uni-mol pretraining dataset. Figures~\ref{fig:unimol-descriptors}, ~\ref{fig:unimol-functional-groups}, and ~\ref{fig:unimol_scaffold} shows visualizations colored by molecular weight, net formal charge, ring count, aromatic heavy-atom fraction, the six most frequent functional groups, and molecular scaffolds. At the low level, we observe limited separation according to molecular properties and no clear clustering by functional group. A small exception is negatively charged molecules, which are concentrated around two localized islands, whereas neutral and positively charged molecules are distributed broadly. At the mid-level, molecular scaffolds show clearer separation, with some distinguishable local clusters, although many scaffolds remain scattered. Thus, scaffold identity is more apparent in the organization of the representations than individual low-level features, but does not fully determine it.

To assess whether low-level information remains accessible despite this limited visual separation, table~\ref{tab:unimol-structural-probes} evaluates linear ridge-regression probes for ring count, aromatic heavy atom fraction and radius of gyration, and compare to a composition-and-size baseline. We use a 70/15/15\% train/val/test split of scaffold groups defined by Murcko scaffold, with acyclic molecules grouped by canonical SMILES. Again features are standardized using training-set statistics, and regularization is selected for each target by validation mean squared error, and we report the test-set coefficient of determination. Embeddings outperform composition and size for all three targets, with the largest improvement for radius of gyration. Combined features achieve the highest scores, although gains over embeddings alone are small for aromatic fraction and radius of gyration. Thus, despite limited visual separation in the UMAP projections, the representations make molecular structural and geometric properties accessible to linear prediction across scaffold-disjoint partitions.

\begin{table}[htbp]
    \centering
    \small
    \caption{Uni-Mol structural probe results.}
    \label{tab:unimol-structural-probes}
    \begin{tabular}{@{}lccc@{}}
        \toprule
        Input & Ring count & Aromatic fraction & Radius of gyration \\
        \midrule
        Composition + size & .706 & .642 & .615 \\
        Embeddings         & .909 & .848 & .964 \\
        Combined           & \textbf{.958} & \textbf{.856} & \textbf{.969} \\
        \bottomrule
    \end{tabular}
\end{table}

\begin{figure}[htbp]
    \centering
    \includegraphics[
        width=\linewidth,
        height=0.85\textheight,
        keepaspectratio
    ]{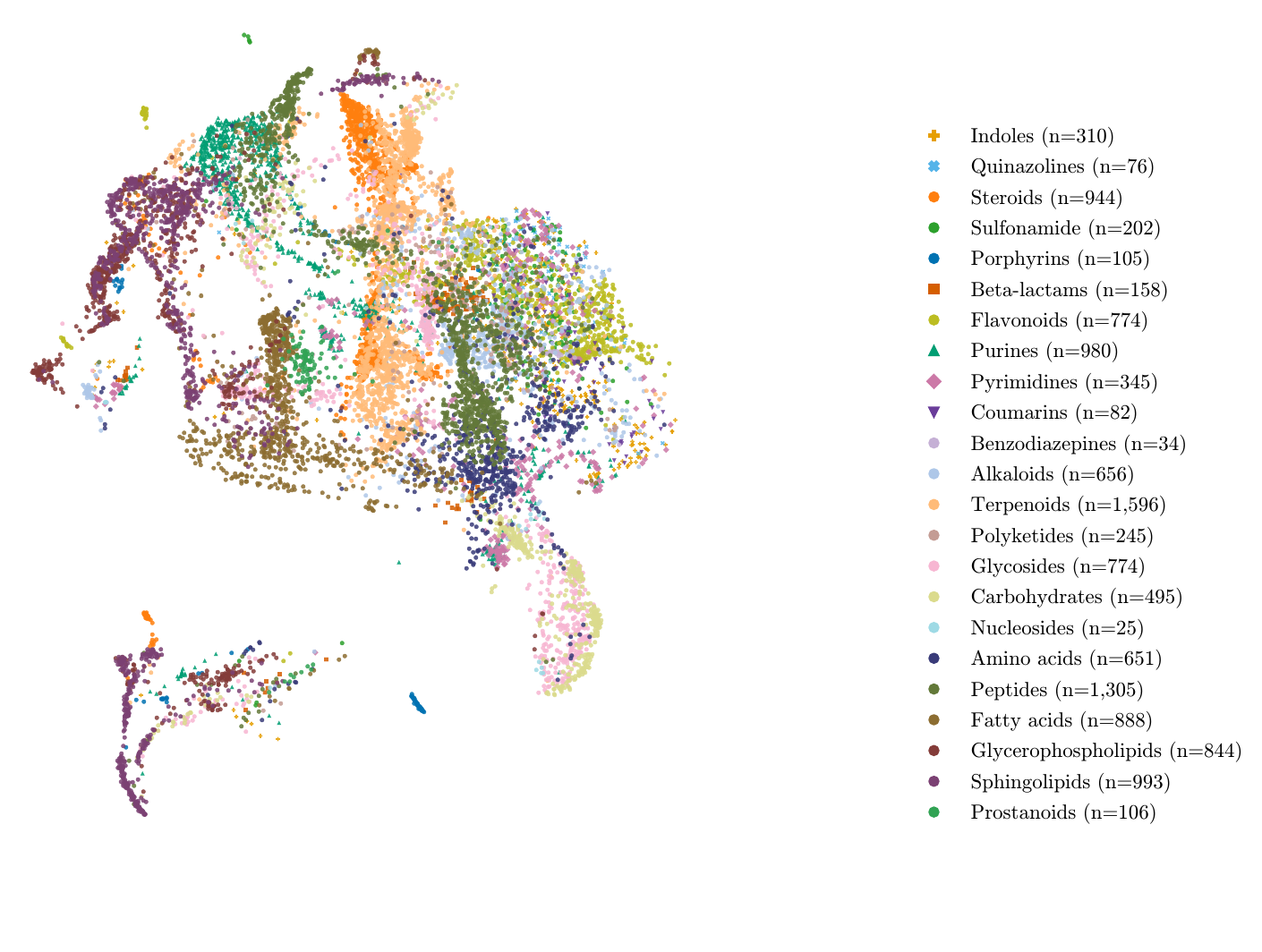}
    \caption{UMAP visualization of ChEBI representations, with exclusive chemical families highlighted.}
    \label{fig:chebi-families-combined}
\end{figure}

\begin{figure}[htbp]
    \centering
    \includegraphics[
        width=\linewidth,
        height=0.85\textheight,
        keepaspectratio
    ]{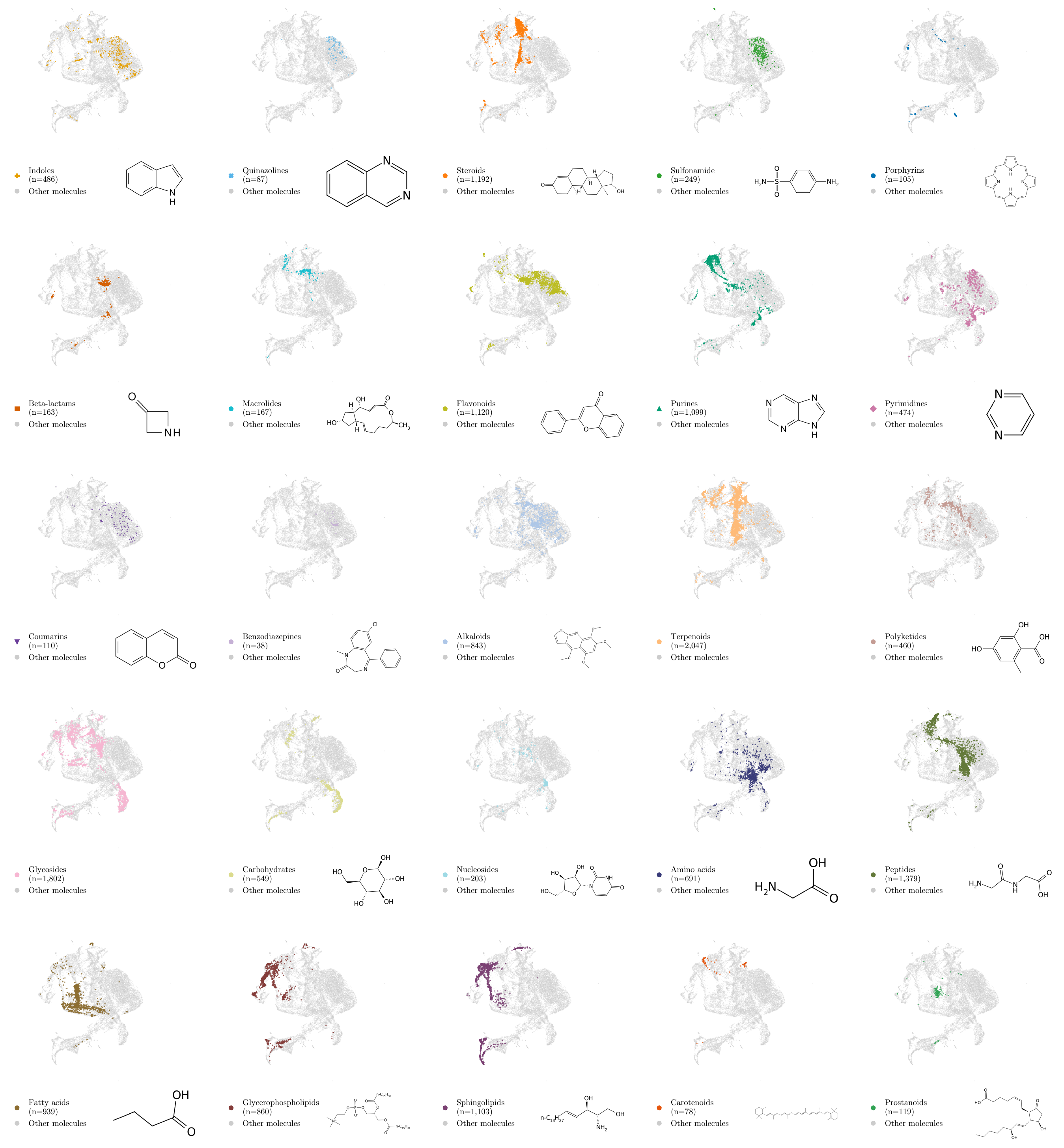}
    \caption{Visualization of ChEBI representations, with
    chemical families highlighted individually and the rest in grey.}
    \label{fig:chebi-families}
\end{figure}

Next, we encode and plot approximately 32,252 molecules from the manually annotated three-star subset of ChEBI~\cite{chebi} and color them by chemical family in figures~\ref{fig:chebi-families-combined} and ~\ref{fig:chebi-families}. Here, the projections exhibit clearer clustering than the low-level property and mid-level scaffold colorings of the Uni-Mol sample. Table~\ref{tab:chebi-family-probes} quantifies this using linear ridge scores fitted to binary membership labels. We use the same scaffold-group splitting procedure and training-set standardization, and regularization as above, selecting regularization separately for each family by validation average precision. Evaluation includes families with at least 50 annotated positives in training, 20 in validation, and 20 in testing, together with at least 20 unannotated examples in each of validation and testing. We report average precision and AUROC averaged equally across eligible families. The outputs are ranking scores, and unannotated membership is treated as the negative class for evaluation. Embeddings substantially outperform composition and size in recovering annotated chemical-family membership, while their combination yields only small further gains. Together, these results suggest a hierarchy in the organization of the learned representations: low-level properties and functional groups show no or very limited visual separation, mid-level scaffolds exhibit partial clustering, and high-level chemical families show the clearest organization. This pattern is consistent with the hypothesized effect of Atom-JEPA pretraining, with latent-space prediction encouraging representations to prioritize higher-level chemical features while reducing the influence of low-level details. The strong structural probe results further suggest that this emphasis does not require discarding low-level information entirely; such information remains recoverable. 

\begin{table}[htbp]
    \centering
    \small
    \caption{ChEBI family probe results.}
    \label{tab:chebi-family-probes}
    \begin{tabular}{@{}lcc@{}}
        \toprule
        Input & Mean average precision & Mean AUROC \\
        \midrule
        Composition + size & .186 & .874 \\
        Embeddings         & .547 & .964 \\
        Combined           & \textbf{.554} & \textbf{.966} \\
        \bottomrule
    \end{tabular}
\end{table}

\newpage
\subsection{Effect of Pretraining on Molecular Representations}
\label{app:effect-of-pretraining}
Finally, to assess how Atom-JEPA pretraining shapes molecular representations beyond the inductive biases of the GNN backbone, we compare pretrained and randomly initialized encoders with identical architectures on the QM9 dataset. For each, we train frozen-backbone probes using the dataset split specified in~\ref{app:finetune_details} and visualize representations of the test set molecules to relate their organization to predictive performance. 

\begin{figure}[htbp]
    \centering
    \includegraphics[
        width=\linewidth,
        height=0.65\textheight,
        keepaspectratio
    ]{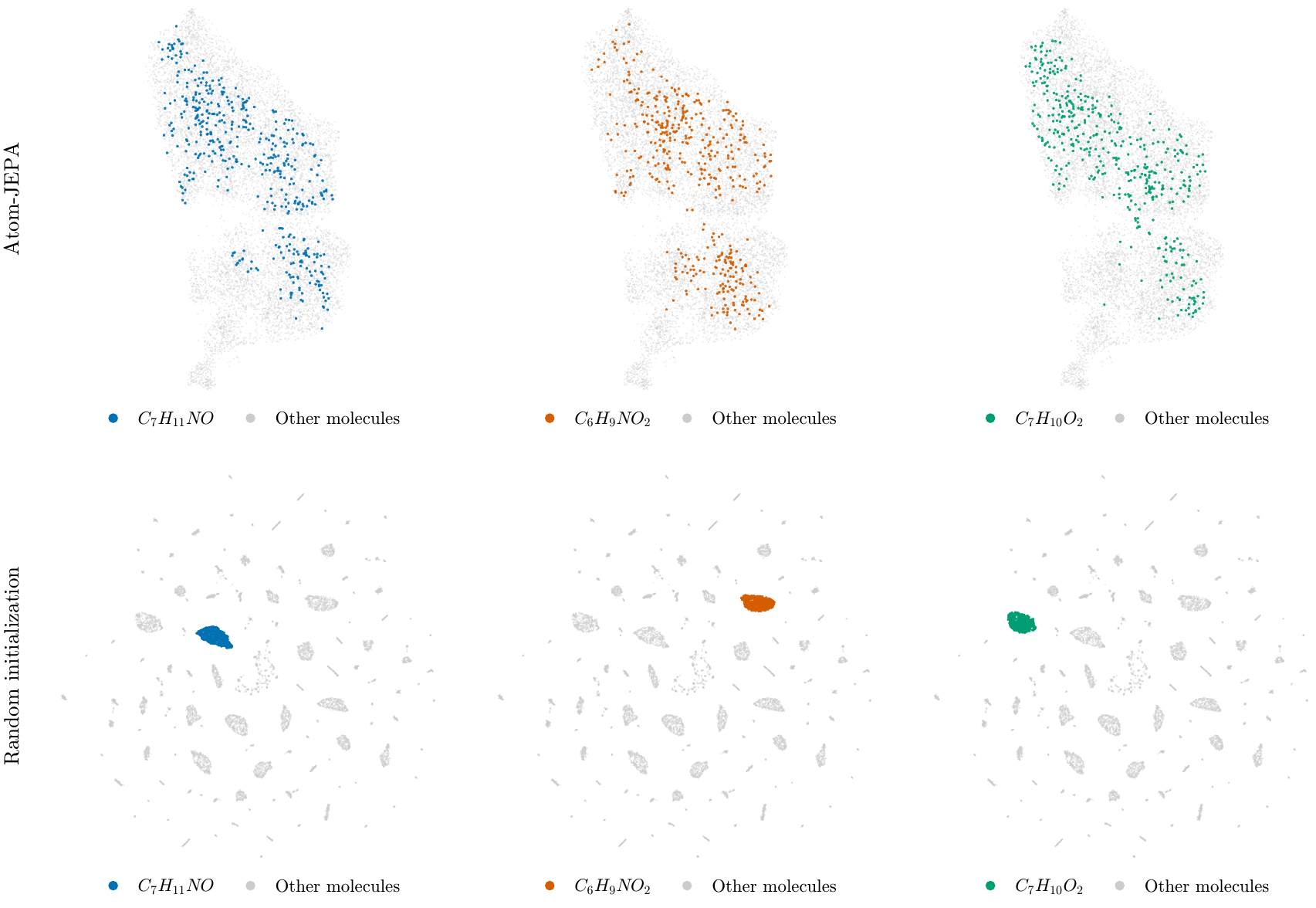}
    \caption{The three most frequent size/composition profiles
    highlighted in the randomly initialized and the Atom-JEPA pretrained QM9 representation, respectively}
    \label{fig:qm9-random-profiles}
\end{figure}

\begin{figure}[htbp]
    \centering
    \includegraphics[
        width=\linewidth,
        height=0.88\textheight,
        keepaspectratio
    ]{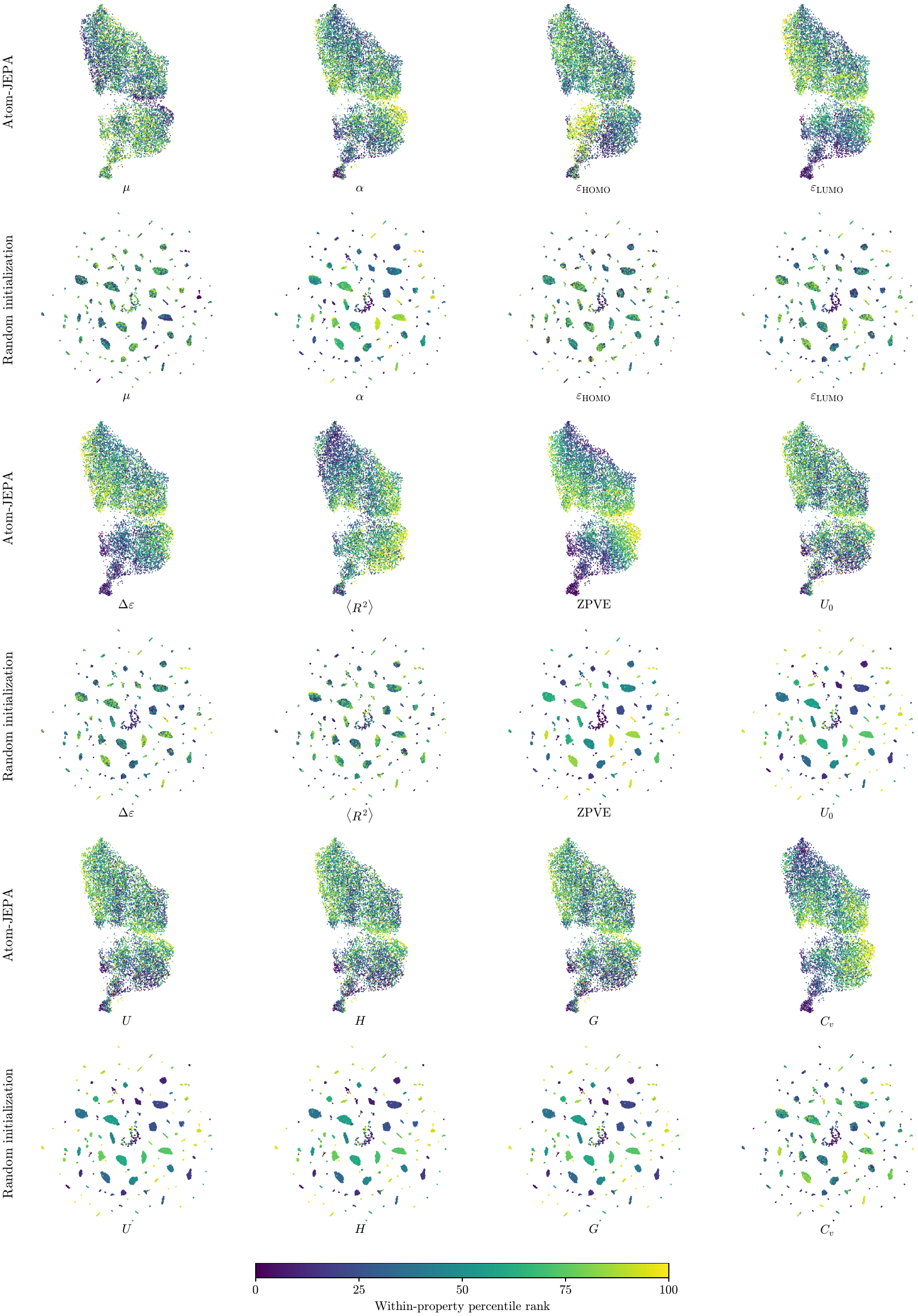}
    \caption{Pretrained and randomly initialized QM9 representations
    colored by within-property percentile rank for all twelve targets.}
    \label{fig:qm9-all-properties}
\end{figure}

Figure~\ref{fig:qm9-random-profiles} highlights the three most frequent chemical formulas in the test set. The randomly initialized encoder separates these formulas into 3 distinct clusters, suggesting that without pretraining representations are organized entirely by molecular size and elemental composition. In contrast, as found above Atom-JEPA representations are not separated by low-level features such as size and composition. Coloring the embeddings by the twelve QM9 properties (Figure~\ref{fig:qm9-all-properties}) also reveals clear patterns for several targets in the random representations. These patterns are consistent with a simple organization by size and composition, which already accounts for substantial variation in several properties. Again, the Atom-JEPA representations do not exhibit the same apparent patterns. 

To assess how these visual differences translate to downstream performance, we train all-layer MLP probes on frozen per-atom features from both encoders and report results in Table~\ref{tab:qm9_mlp_probes}. All probes outperform the training-set mean predictor on every target, confirming that the backbone's inductive biases alone provide useful predictive features. Nevertheless, Atom-JEPA achieves substantially lower MAE across all twelve targets with both probe types. With all-layer features, pretraining reduces MAE by approximately three- to fourfold for polarizability, the HOMO--LUMO gap, HOMO and LUMO energies, and heat capacity, and by seven- to eightfold for the atom-referenced energies $G$, $H$, $U$, and $U_0$. Gains are smaller for dipole moment, where random features already perform comparatively well. An additional finding is that all-layer probes outperform last-layer probes on every target for Atom-JEPA and on eleven of twelve targets for the random encoder, with particularly pronounced gains for Atom-JEPA energy predictions.

These results show that clear visual separation by target value does not indicate greater predictive utility. The patterns in the random representations reflect a simple organization by molecular size and elemental composition, whereas Atom-JEPA supports substantially more accurate prediction without comparable formula-based clustering. Together with the chemical-family analysis, these findings are consistent with latent-space predictive pretraining shifting representations beyond these basic attributes toward higher-level, property-relevant molecular information that is not readily apparent in two-dimensional projections, but nevertheless improve downstream performance.

\begin{table}[htbp]
\centering
\caption{QM9 frozen-backbone MLP probe test MAE results using all-layer or last-layer features for a pretrained encoder vs random initialization, with a training-set mean predictor as a baseline (atom-referenced for $U$, $U_0$, $H$, $G$).}
\label{tab:qm9_mlp_probes}
\centering
\normalsize
\scalebox{\tablescale}{%
\begin{tabular}{lcccccccccccc}
\toprule
Model & $\alpha$ & $\Delta\varepsilon$ & $\varepsilon_{\text{HOMO}}$ & $\varepsilon_{\text{LUMO}}$ & $\mu$ & $C_v$ & $G$ & $H$ & $\langle R^2 \rangle$ & $U$ & $U_0$ & ZPVE \\
 & $a_0^3$ & eV & eV & eV & D & $\frac{\text{cal}}{\text{mol K}}$ & eV & eV & $a_0^2$ & eV & eV & eV \\
\midrule
Training-set mean & 6.343 & 1.077 & .434 & 1.055 & 1.165 & 3.195 & 7.511 & 8.304 & 203.364 & 8.244 & 8.170 & .720 \\
\midrule
\multicolumn{13}{l}{All layers} \\
\midrule
Atom-JEPA & \textbf{.091} & \textbf{.095} & \textbf{.061} & \textbf{.064} & \textbf{.032} & \textbf{.031} & \textbf{.013} & \textbf{.012} & \textbf{.419} & \textbf{.012} & \textbf{.012} & \textbf{.001} \\
Random initialization & .371 & .279 & .171 & .221 & .039 & .108 & .091 & .089 & 1.850 & .089 & .091 & .004 \\
\midrule
\multicolumn{13}{l}{Last layer} \\
\midrule
Atom-JEPA & .217 & .141 & .094 & .099 & .034 & .071 & .061 & .070 & 1.579 & .068 & .072 & .004 \\
Random initialization & .479 & .327 & .192 & .246 & .038 & .135 & .114 & .110 & 2.444 & .114 & .110 & .005 \\
\bottomrule
\end{tabular}%
}
\end{table}

\newpage
~\newpage
\section{ADMET Full Results}
\label{app:fulladmet}

\subsection{Statistical Tests Setup}
\label{appendix:statistical_test}
To compute statistical indistinguishable models in Table \ref{table:admet_comparison}, we compute an ANOVA test over each endpoint. For Biogen and ChEMBL the design is blocked across folds, whereas ExpansionRX is unpaired due to using the same test set each run. The best model that is compared against is the model with the lowest mean error over the respective endpoint folds. For runs where we have the individual fold errors, we check if each models computed $95\%$ confidence intervals do not overlap using a pairwise Tukey test. Because we only have the mean run error for KERMT, we instead compare using a single-sided t-test with Holm–Bonferroni correction.

\subsection{Model Prediction Correlation} \label{appendix:prediction_correlation}
We present the full analysis of the differences between the ADMET models. We quantify difference between the prediction of models by their spearman correlation $(r)$. We use residuals instead of direct predictions to reduce model strength as a confounder.  In the interest of ensembling model predictions, we also control against the case where one model happens to be weakly correlated with other models, but strongly correlated with the sum of the other models. For completeness we calculate the uniqueness score as follows:
For endpoint $t$ and run $s$, let $e_m\in\mathbb{R}^{n}$ denote model $m$'s residual vector $\hat{y}-y$ over the $n$ test molecules with an observed label for $t$, and let $\mathcal{P}(m)$ be one representative of each of the four model families other than $m$'s own. We regress $e_m$ on $\{e_k\}_{k\in\mathcal{P}(m)}$ with an intercept by ordinary least squares and define 
\begin{equation} 
U_m(t,s) = 1-R^2_{\mathrm{adj}} = \bigl(1-R^2\bigr)\,\frac{n-1}{n-p-1} , 
\end{equation} where $p=\lvert\mathcal{P}(m)\rvert=4$ for our for model comparisons, $R^2$ is the coefficient of determination of that fit. $U_m$ is the share of model $m$'s squared error that no linear combination of the other families reproduces,. We do not include endpoints with less than 30 test datapoints for robustness. We only compare other models to Atom-JEPA averaged over 10 conformers to not bias the results by also including the the non-conformer averaged Atom-JEPA model. The single and $10\times$ conformer predictions correlate with over $0.99$.
The results can be seen in \cref{tab:model_distinctiveness}.
\begin{table*}[h]
\caption{\textbf{Model distinctiveness.} How reproducible each model's errors are from the other four model families, measured two ways on the same per-molecule test residuals. \emph{Residual correlation} is the averaged \emph{pairwise} correlation between other models, averaged over test data-points, \emph{Uniqueness} is the $1-R^2_{\mathrm{adj}}$ from regressing a model's residuals on all four peers' residuals at once. Per-benchmark columns are means over each benchmark's endpoints. \emph{Mean rank} ranks the families within each benchmark and endpoint cell and averages with equal weight per benchmark, with rank 1 = most distinct. The brackets are a stratified cluster bootstrap over endpoints. \emph{Group} shows the group of models that cannot separated by the paired 95\% bootstrap interval on their mean-rank difference.}
\label{tab:model_distinctiveness}
\centering
\normalsize
\scalebox{\tablescale}{%
\begin{tabular}{lcccc c}
\toprule
Model & Biogen & ChEMBL-MT & ExpansionRx & Mean rank across endpoints & Group \\
\midrule
\multicolumn{6}{l}{\textbf{Residual correlation} $\bar{r}$ to peers \emph{(lower is more distinct)}} \\
\midrule
\rowcolor{atomjepablue} Atom-JEPA $\times$10 conf & $.723$ & $\mathbf{.791}$ & $.730$ & \shortstack{$\mathbf{1.98}$\\{\scriptsize [1.66, 2.34]}} & a \\
Chemprop & $\mathbf{.719}$ & $.803$ & $\mathbf{.681}$ & \shortstack{$2.53$\\{\scriptsize [2.02, 2.88]}} & ab \\
Mol-JEPA $\cdot$ modalities & $.738$ & $.801$ & $.749$ & \shortstack{$2.78$\\{\scriptsize [2.42, 3.14]}} & b \\
Morgan + RDKit $\cdot$ LGBM & $.724$ & $.814$ & $.761$ & \shortstack{$2.98$\\{\scriptsize [2.59, 3.31]}} & b \\
CKERMT & $.779$ & $.829$ & $.787$ & \shortstack{$4.73$\\{\scriptsize [4.48, 4.86]}} & c \\
\midrule
\multicolumn{6}{l}{\textbf{Uniqueness} $1-R^2_{\mathrm{adj}}$ \emph{(higher is more distinct)}} \\
\midrule
\rowcolor{atomjepablue} Atom-JEPA $\times$10 conf & $\mathbf{.327}$ & $\mathbf{.253}$ & $.283$ & \shortstack{$\mathbf{1.97}$\\{\scriptsize [1.65, 2.32]}} & a \\
Chemprop & $.325$ & $.221$ & $\mathbf{.392}$ & \shortstack{$2.53$\\{\scriptsize [2.02, 2.89]}} & ab \\
Mol-JEPA $\cdot$ modalities & $.285$ & $.221$ & $.246$ & \shortstack{$2.89$\\{\scriptsize [2.52, 3.23]}} & b \\
Morgan + RDKit $\cdot$ LGBM & $.312$ & $.192$ & $.228$ & \shortstack{$2.99$\\{\scriptsize [2.57, 3.39]}} & b \\
CKERMT & $.214$ & $.171$ & $.187$ & \shortstack{$4.63$\\{\scriptsize [4.39, 4.78]}} & c \\
\midrule
\bottomrule
\end{tabular}%
}
\end{table*}

\subsection{Biogen ADME}
\begin{table}[H]
\setlength{\tabcolsep}{4.5pt}
\centering
    \caption{Test MAE mean $\pm$ standard deviation across scaffold splits across tasks for Biogen dataset. $\dag$: Data from Figure 13 in~\cite{DBLP:journals/corr/abs-2606-11508}.}
    \label{tab:biogen_error_scaffold}
\adjustbox{max width=\textwidth}{%
    \begin{tabular}{lccccccc}
    \toprule
    \textbf{Model} & \textbf{HLM CLint} & \textbf{RLM CLint} & \textbf{MDCK}
    & \textbf{logS} & \textbf{PPB Human} & \textbf{PPB Rat} & \textbf{Mean} \\
    \midrule
    
    \multicolumn{8}{l}{\textit{GBT models}} \\
    \quad RDKit          & $.3773 \err{.0190}$ & $.4347 \err{.0268}$ & $.3493 \err{.0116}$ & $.3725 \err{.0204}$ & $.4713 \err{.1293}$ & $.4872 \err{.0594}$ & $.4154 \err{.0298}$ \\
    \quad Morgan         & $.4115 \err{.0242}$ & $.4616 \err{.0182}$ & $.3682 \err{.0131}$ & $.3838 \err{.0230}$ & $.5085 \err{.1062}$ & $.6293 \err{.0828}$ & $.4605 \err{.0260}$ \\
    \quad Morgan + RDKit & $.3712 \err{.0140}$ & $.4136 \err{.0239}$ & $.3362 \err{.0103}$ & $.3663 \err{.0143}$ & $.4462 \err{.1109}$ & $.5125 \err{.0984}$ & $.4077 \err{.0339}$ \\
    \quad Morgan + RDKit + Avalon + ErG & $.3863 \err{.0186}$ & $.4160 \err{.0196}$ & $.3497 \err{.0154}$ & $.3882 \err{.0252}$ & $.5454 \err{.1507}$ & $.5686 \err{.0660}$ & $.4424 \err{.0247}$ \\
    \quad Mol-JEPA CLS & $.4051 \err{.0218}$ & $.4523 \err{.0187}$ & $.3689 \err{.0096}$ & $.3887 \err{.0262}$ & $.5302 \err{.1025}$ & $.5264 \err{.0835}$ & $.4453 \err{.0317}$ \\
    \rowcolor{atomjepablue}
    \quad Atom-JEPA Last Layer & $.4300 \err{.0221}$ & $.4869 \err{.0206}$ & $.3791 \err{.0127}$ & $.3813 \err{.0201}$ & $.5182 \err{.0828}$ & $.5090 \err{.0991}$ & $.4507 \err{.0213}$\\
    \rowcolor{atomjepablue}
    \quad Atom-JEPA All Layer  & $.4021 \err{.0134}$ & $.4613 \err{.0198}$ & $.3563 \err{.0039}$ & $.3691 \err{.0229}$ & $.4798 \err{.1242}$ & $.4753 \err{.0763}$ & $.4240 \err{.0266}$ \\

    \addlinespace
    \multicolumn{8}{l}{\textit{MLP models}} \\
    \quad RDKit           & $.3772 \err{.0249}$ & $.4400 \err{.0239}$ & $.3383 \err{.0192}$ & $.3798 \err{.0272}$ & $.4112 \err{.0802}$ & $.4227 \err{.0932}$ & $.3949 \err{.0292}$ \\
    \quad Morgan           & $.4323 \err{.0271}$ & $.5055 \err{.0343}$ & $.3917 \err{.0215}$ & $.4159 \err{.0221}$ & $.4793 \err{.0778}$ & $.4794 \err{.0793}$ & $.4507 \err{.0218}$ \\
    \quad Morgan + RDKit  & $.3469 \err{.0197}$ & $.3998 \err{.0202}$ & $.3344 \err{.0238}$ & $.3772 \err{.0133}$ & $.4034 \err{.0778}$ & $.4412 \err{.0778}$ & $.3838 \err{.0194}$ \\
    \quad Morgan + RDKit + Avalon + ErG  & $.3336 \err{.0161}$ & $.3892 \err{.0139}$ & $.3197 \err{.0193}$ & $.3679 \err{.0241}$ & $.4538 \err{.0758}$ & $.4651 \err{.0205}$ & $.3882 \err{.0137}$ \\
    \quad Mol-JEPA CLS & $.3710 \err{.0209}$ & $.4148 \err{.0205}$ & $.3357 \err{.0172}$ & $.3660 \err{.0176}$ & $.4000 \err{.0441}$ & $.3810 \err{.0375}$ & $.3781 \err{.0137}$ \\
    \quad Mol-JEPA modalities  & $.3423 \err{.0115}$ & $.3932 \err{.0221}$ & $.3237 \err{.0143}$ & $.3293 \err{.0174}$ & $.4048 \err{.0585}$ & $\mathbf{.3782} \err{.0409}$ & $.3619 \err{.0141}$ \\
    \rowcolor{atomjepablue}
    \quad Atom-JEPA Last Layer & $.3754 \err{.0244}$ & $.4304 \err{.0144}$ & $.3383 \err{.0086}$ & $.3594 \err{.0238}$ & $.4297 \err{.0637}$ & $.4516 \err{.0503}$ & $.3975 \err{.0181}$ \\
    \rowcolor{atomjepablue}
    \quad Atom-JEPA All Layer  & $.3757 \err{.0230}$ & $.4322 \err{.0257}$ & $.3153 \err{.0115}$ & $.3518 \err{.0197}$ & $.4282 \err{.0883}$ & $.4320 \err{.0772}$ & $.3892 \err{.0270}$\\
    
    \addlinespace
    \multicolumn{8}{l}{\textit{Finetuning}} \\
    \rowcolor{atomjepablue}
    \quad No Pretrain & $.3232 \err{.0170}$ & $.3781 \err{.0082}$ & $.3322 \err{.0114}$ & $.3274 \err{.0091}$ & $.4696 \err{.1024}$ & $.4623 \err{.0711}$ & $.3821 \err{.0281}$ \\

    \rowcolor{atomjepablue}
    \quad Atom-JEPA & $.2950 \err{.0146}$ & $.3548 \err{.0185}$ & $.2816 \err{.0107}$ & $.3246 \err{.0110}$ & $.4221 \err{.0922}$ & $.3970 \err{.0747}$ & $.3458 \err{.0260}$ \\
    \rowcolor{atomjepablue}
    \quad Atom-JEPA (10-conf. ensemble) & $\mathbf{.2931} \err{.0147}$ & $\mathbf{.3523} \err{.0186}$ & $.2796 \err{.0110}$ & $\mathbf{.3231} \err{.0112}$ & $.4199 \err{.0940}$ & $.3932 \err{.0759}$ & $\mathbf{.3435} \err{.0267}$ \\

    \addlinespace
    \multicolumn{8}{l}{\textit{Literature models}} \\

    \quad Chemprop & $.3133 \err{.0058}$ & $.3683 \err{.0141}$ & $.2954 \err{.0070}$ & $.3305 \err{.0143}$ & $\mathbf{.3968} \err{.1017}$ & $.4062 \err{.1136}$ & $.3518 \err{.0317}$ \\

    \quad CKERMT Our runs & $.2996 \err{.0095}$ & $.3646 \err{.0194}$ & $.2892 \err{.0136}$ & $.3265 \err{.0104}$ & $.4093 \err{.0820}$ & $.4451 \err{.0953}$ & $.3557 \err{.0290}$ \\
    
    \quad KERMT base\textsuperscript{\dag}
                                & $.320$ & $.402$ & $.289$ & $.344$ & -- & -- & -- \\
    \quad KERMT task-specific\textsuperscript{\dag}
                                & $.309$ & $.402$ & $.278$ & $.339$ & -- & -- & -- \\
    \quad CKERMT base\textsuperscript{\dag}
                                & $.302$ & $.389$ & $.275$ & $.329$ & -- & -- & -- \\
    \quad CKERMT task-specific\textsuperscript{\dag}
                                & $.294$ & $.389$ & $\mathbf{.268}$ & $.333$ & -- & -- & -- \\

    \bottomrule
    \end{tabular}} % Adjustbox
\end{table}

\begin{table}[H]
    \centering
    \setlength{\tabcolsep}{4.5pt}
        \caption{Test Pearson $r^2$ $\pm$ standard deviation across cluster splits across tasks for Biogen dataset. All literature model data from ~\cite{DBLP:journals/corr/abs-2510-12719}.}
    \label{tab:biogen_error_cluster}
    \adjustbox{max width=\textwidth}{%
    \begin{tabular}{lccccccc}
    \toprule
    \textbf{Model} & \textbf{HLM CLint} & \textbf{RLM CLint} & \textbf{MDCK}
    & \textbf{logS} & \textbf{PPB Human} & \textbf{PPB Rat} & \textbf{Mean} \\
    \midrule
    
    \multicolumn{8}{l}{\textit{GBT models}} \\
    \quad RDKit            & $.3846 \err{.0170}$ & $.4337 \err{.0041}$ & $.3921 \err{.0071}$ & $.2143 \err{.0069}$ & $.3666 \err{.0005}$ & $.3733 \err{.0807}$ & $.3608 \err{.0134}$ \\
    \quad Morgan           & $.3256 \err{.0244}$ & $.3722 \err{.0079}$ & $.3635 \err{.0093}$ & $.1569 \err{.0032}$ & $.3720 \err{.0551}$ & $.0372 \err{.0293}$ & $.2712 \err{.0103}$ \\
    \quad Morgan + RDKit  & $.4411 \err{.0020}$ & $.4950 \err{.0055}$ & $.4397 \err{.0073}$ & $.2346 \err{.0007}$ & $.4152 \err{.0089}$ & $.3111 \err{.0075}$ & $.3895 \err{.0008}$ \\
    \quad Morgan + RDKit + Avalon + ErG  & $.4469 \err{.0229}$ & $.4800 \err{.0083}$ & $.3881 \err{.0075}$ & $.2428 \err{.0125}$ & $.3454 \err{.0115}$ & $.2428 \err{.0730}$ & $.3577 \err{.0146}$ \\
    \rowcolor{atomjepablue}
    \quad Atom-JEPA Last Layer & $.2940 \err{.0205}$ & $.3840 \err{.0111}$ & $.2995 \err{.0118}$ & $.2293 \err{.0027}$ & $.3661 \err{.0463}$ & $.2613 \err{.0799}$ & $.3057 \err{.0057}$ \\
    \rowcolor{atomjepablue}
    \quad Atom-JEPA All Layer & $.3598 \err{.0168}$ & $.4226 \err{.0057}$ & $.4018 \err{.0167}$ & $.2429 \err{.0378}$ & $.4674 \err{.0436}$ & $.4564 \err{.0267}$ & $.3918 \err{.0008}$ \\

    \addlinespace
    \multicolumn{8}{l}{\textit{MLP models}} \\
    \quad RDKit           & $.4821 \err{.0022}$ & $.5182 \err{.0073}$ & $.4707 \err{.0073}$ & $.2141 \err{.0092}$ & $.4479 \err{.0370}$ & $.4347 \err{.0364}$ & $.4279 \err{.0141}$ \\
    \quad Morgan          & $.2489 \err{.0071}$ & $.3001 \err{.0031}$ & $.2284 \err{.0060}$ & $.2256 \err{.0144}$ & $.3418 \err{.0352}$ & $.2555 \err{.0316}$ & $.2667 \err{.0119}$ \\
    \quad Morgan + RDKit  & $.4223 \err{.0139}$ & $.4911 \err{.0116}$ & $.4131 \err{.0040}$ & $.1898 \err{.0021}$ & $.4922 \err{.0395}$ & $.4088 \err{.0663}$ & $.4029 \err{.0123}$ \\
    \quad Morgan + RDKit + Avalon + ErG  & $.4639 \err{.0092}$ & $.5312 \err{.0098}$ & $.5027 \err{.0168}$ & $.2164 \err{.0070}$ & $.4828 \err{.0407}$ & $.3682 \err{.0311}$ & $.4275 \err{.0135}$ \\
    \rowcolor{atomjepablue}
    \quad Atom-JEPA Last Layer & $.4515 \err{.0043}$ & $.4706 \err{.0036}$ & $.4180 \err{.0013}$ & $.3240 \err{.0147}$ & $.4224 \err{.0350}$ & $.3639 \err{.0450}$ & $.4084 \err{.0159}$ \\
    \rowcolor{atomjepablue}
    \quad Atom-JEPA All Layer & $.4566 \err{.0053}$ & $.4836 \err{.0078}$ & $.5208 \err{.0054}$ & $.3518 \err{.0005}$ & $.5498 \err{.0067}$ & $.5042 \err{.0059}$ & $.4778 \err{.0011}$\\
    
    \addlinespace
    \multicolumn{8}{l}{\textit{Finetuning}} \\
    \rowcolor{atomjepablue}
    \quad Atom-JEPA   & $.6384 \err{.0048}$ & $.6417 \err{.0053}$ & $.6021 \err{.0104}$ & $.3728 \err{.0154}$ & $.5808 \err{.0187}$ & $.4332 \err{.0532}$ & $.5448 \err{.0146}$\\
    \rowcolor{atomjepablue}
    \quad Atom-JEPA (10-conf. ensemble) & $.6444 \err{.0048}$ & $.6465 \err{.0052}$ & $.6070 \err{.0105}$ & $.3777 \err{.0153}$ & $.5881 \err{.0190}$ & $.4380 \err{.0534}$ & $.5503 \err{.0146}$\\

    \addlinespace \multicolumn{8}{l}{\textit{Literature models}} \\ 
    \quad MolFormer ST & $.510\err{.004}$ & $.570\err{.015}$ & $.415\err{.007}$ & $.181\err{.010}$ & $.269\err{.035}$ & $.119\err{.023}$ & $.344$ \\
    \quad Chemprop ST & $.566\err{.022}$ & $.596\err{.014}$ & $.475\err{.009}$ & $.296\err{.027}$ & $.190\err{.012}$ & $.121\err{.085}$ & $.374$ \\
    \quad KPGT ST$_{\mathrm{complex}}$ & $.505\err{.004}$ & $.548\err{.013}$ & $.520\err{.007}$ & $.337\err{.003}$ & $.291\err{.031}$ & $.269\err{.043}$ & $.412$ \\
    \quad KPGT ST & $.580\err{.009}$ & $.597\err{.013}$ & $.538\err{.012}$ & $.369\err{.012}$ & $.358\err{.062}$ & $.284\err{.064}$ & $.454$ \\
    \quad KERMT ST & $.600\err{.019}$ & $.577\err{.017}$ & $.525\err{.009}$ & $.343\err{.009}$ & $.381\err{.018}$ & $.314\err{.064}$ & $.457$ \\
    \quad Chemprop MT & $.558\err{.019}$ & $.602\err{.011}$ & $.519\err{.012}$ & $.380\err{.012}$ & $.572\err{.059}$ & $.501\err{.046}$ & $.522$ \\
    \quad KPGT MT & $.562\err{.007}$ & $.580\err{.020}$ & $.529\err{.007}$ & $.318\err{.013}$ & $.362\err{.019}$ & $.315\err{.020}$ & $.444$ \\
    \quad KERMT MT & $.634\err{.010}$ & $.602\err{.011}$ & $.535\err{.009}$ & $.365\err{.008}$ & $.481\err{.040}$ & $.427\err{.038}$ & $.507$ \\
    \bottomrule
    \end{tabular}}%Adjusttable
\end{table}

\subsection{ExpansionRX and ChEMBL-MT}
\begin{sidewaystable*}
    \centering
\tiny
    \caption{Test MAE mean $\pm$ standard deviation across cluster splits across tasks for ExpansionRX dataset. $\dag$: Data from~\cite{DBLP:journals/corr/abs-2606-11508}}.
    \label{tab:expansionrx_error}
\begin{tabular}{lcccccccccc}
\toprule
\textbf{Model}
& $\boldsymbol{\log D}$
& \textbf{KSOL}
& \shortstack{\textbf{HLM}\\\textbf{CLint}}
& \shortstack{\textbf{MLM}\\\textbf{CLint}}
& \shortstack{\textbf{Caco-2 Papp}\\\textbf{A$\boldsymbol{\rightarrow}$B}}
& \shortstack{\textbf{Caco-2}\\\textbf{Efflux}}
& \textbf{MPPB}
& \textbf{MBPB}
& \textbf{MGMB}
& \shortstack{\textbf{Macro}\\\textbf{test}} \\
\midrule

\multicolumn{11}{l}{\textit{GBDT models}} \\
\quad Morgan \textsuperscript{\dag}
                            & $.517 \err{.011}$ & $.490 \err{.028}$ & $.382 \err{.007}$ & $.406 \err{.011}$ & $.511 \err{.013}$ & $.518 \err{.032}$ & $.313 \err{.020}$ & $.317 \err{.007}$ & $.396 \err{.012}$ & $.428 \err{.004}$ \\
\quad RDKit           & $.567 \err{.009}$ & $.535 \err{.007}$ & $.363 \err{.004}$ & $.401 \err{.008}$ & $.468 \err{.005}$ & $.494 \err{.010}$ & $.346 \err{.011}$ & $.322 \err{.005}$ & $.353 \err{.014}$ & $.428 \err{.002}$ \\
\quad Morgan + RDKit  & $.523 \err{.009}$ & $.540 \err{.010}$ & $.355 \err{.009}$ & $.409 \err{.008}$ & $.458 \err{.009}$ & $.514 \err{.012}$ & $.332 \err{.006}$ & $.277 \err{.008}$ & $.357 \err{.013}$ & $.418 \err{.003}$ \\
\quad Morgan + RDKit + Avalon + ErG & $.511 \err{.019}$ & $.477 \err{.010}$ & $.371 \err{.009}$ & $.445 \err{.017}$ & $.464 \err{.022}$ & $.482 \err{.032}$ & $.333 \err{.012}$ & $.311 \err{.021}$ & $.328 \err{.022}$ & $.414 \err{.005}$ \\
\quad Mol-JEPA cls & $.586 \err{.005}$ & $.521 \err{.007}$ & $.381 \err{.006}$ & $.435 \err{.007}$ & $.530 \err{.005}$ & $.524 \err{.004}$ & $.337 \err{.015}$ & $.376 \err{.018}$ & $.331 \err{.007}$ & $.447 \err{.005}$ \\
\rowcolor{atomjepablue}
\quad Atom-JEPA Last Layer & $.613 \err{.017}$ & $.495 \err{.004}$ & $.389 \err{.004}$ & $.426 \err{.003}$ & $.586 \err{.020}$ & $.556 \err{.008}$ & $.328 \err{.018}$ & $.327 \err{.014}$ & $.374 \err{.010}$ & $.455 \err{.004}$ \\
\rowcolor{atomjepablue}
\quad Atom-JEPA All Layers & $.577 \err{.015}$ & $.509 \err{.004}$ & $.357 \err{.008}$ & $.431 \err{.008}$ & $.533 \err{.017}$ & $.522 \err{.009}$ & $.354 \err{.007}$ & $.331 \err{.009}$ & $.309 \err{.009}$ & $.436 \err{.003}$ \\

\addlinespace
\multicolumn{11}{l}{\textit{MLP models}} \\
\quad Morgan & $.635 \err{.014}$ & $.464 \err{.032}$ & $.417 \err{.013}$ & $.448 \err{.006}$ & $.630 \err{.021}$ & $.579 \err{.019}$ & $.261 \err{.003}$ & $.306 \err{.014}$ & $.292 \err{.011}$ & $.448 \err{.006}$ \\
\quad RDKit & $.647 \err{.010}$ & $.598 \err{.008}$ & $.406 \err{.006}$ & $.440 \err{.006}$ & $.516 \err{.004}$ & $.548 \err{.004}$ & $.437 \err{.016}$ & $.391 \err{.011}$ & $.414 \err{.013}$ & $.489 \err{.006}$ \\
\quad Morgan + RDKit & $.494 \err{.004}$ & $.418 \err{.014}$ & $.373 \err{.011}$ & $.454 \err{.020}$ & $.558 \err{.012}$ & $.522 \err{.010}$ & $.259 \err{.009}$ & $.263 \err{.009}$ & $.300 \err{.006}$ & $.405 \err{.006}$ \\
\quad Morgan + RDKit + Avalon + ErG & $.476 \err{.010}$ & $.389 \err{.004}$ & $.396 \err{.004}$ & $.419 \err{.019}$ & $.512 \err{.011}$ & $.454 \err{.007}$ & $.229 \err{.008}$ & $.244 \err{.004}$ & $.282 \err{.003}$ & $.378 \err{.004}$ \\
\quad Mol-JEPA CLS & $.573 \err{.013}$ & $.509 \err{.020}$ & $.374 \err{.004}$ & $.456 \err{.007}$ & $.457 \err{.005}$ & $.483 \err{.010}$ & $.338 \err{.017}$ & $.372 \err{.021}$ & $.394 \err{.014}$ & $.440 \err{.009}$ \\
\quad Mol-JEPA Modalities & $.545 \err{.014}$ & $.486 \err{.010}$ & $.396 \err{.011}$ & $.434 \err{.018}$ & $.477 \err{.010}$ & $.457 \err{.012}$ & $.356 \err{.007}$ & $.369 \err{.007}$ & $.337 \err{.003}$ & $.428 \err{.003}$ \\

\rowcolor{atomjepablue}
\quad Atom-JEPA All Layers & $.624 \err{.015}$ & $.522 \err{.012}$ & $.361 \err{.009}$ & $.445 \err{.007}$ & $.520 \err{.010}$ & $.551 \err{.012}$ & $.337 \err{.011}$ & $.265 \err{.005}$ & $.283 \err{.004}$ & $.434 \err{.006}$ \\

\addlinespace
\multicolumn{11}{l}{\textit{Finetuning}} \\
\rowcolor{atomjepablue}
\quad  No Pretrain & $.390 \err{.007}$ & $.376 \err{.005}$ & $.347 \err{.007}$ & $.411 \err{.005}$ & $.529 \err{.005}$ & $.486 \err{.007}$ & $.258 \err{.007}$ & $.227 \err{.007}$ & $.282 \err{.004}$ & $.367 \err{.004}$ \\
\rowcolor{atomjepablue}
\quad Atom-JEPA                  & $.363 \err{.006}$ & $.346 \err{.003}$ & $.316 \err{.006}$ & $.403 \err{.006}$ & $.475 \err{.006}$ & $.439 \err{.006}$ & $.218 \err{.006}$ & $.188 \err{.003}$ & $.259 \err{.004}$ & $.334 \err{.002}$ \\
\rowcolor{atomjepablue}
\quad Atom-JEPA (10-conf. ensemble) & $.363 \err{.006}$ & $.344 \err{.003}$ & $.316 \err{.006}$ & $.403 \err{.006}$ & $.475 \err{.006}$ & $.439 \err{.006}$ & $.218 \err{.005}$ & $.188 \err{.003}$ & $.259 \err{.004}$ & $.334 \err{.002}$ \\
\addlinespace

\multicolumn{8}{l}{\textit{Literature models}} \\
\quad Chemprop & $.449 \err{.009}$ & $.404 \err{.006}$ & $.552 \err{.105}$ & $.487 \err{.025}$ & $.528 \err{.032}$ & $.456 \err{.024}$ & $.251 \err{.016}$ & $.339 \err{.019}$ & $.371\err{.031}$ & $.426 \err{.012}$ \\
\quad CKERMT Our runs & $.407 \err{.009}$ & $.406 \err{.008}$ & $.350 \err{.010}$ & $.385 \err{.009}$ & $.472 \err{.009}$ & $.466 \err{.007}$ & $.250 \err{.003}$ & $.273 \err{.011}$ & $.333 \err{.023}$ & $.372 \err{.003}$ \\

\quad KERMT base\textsuperscript{\ddag}
    & $.393$ & $.417$ & $.429$ & $.458$
    & $.472$ & $.489$ & $.247$ & $.225$
    & $.287$ & $.380$ \\
\quad KERMT task-specific\textsuperscript{\ddag}
    & $.396$ & $.396$ & $.400$ & $.423$
    & $.469$ & $.479$ & $.263$ & $.251$
    & $.299$ & $.375$ \\
\quad CMIM only task-specific\textsuperscript{\ddag}
    & $.461$ & $.477$ & $.418$ & $.420$
    & $.468$ & $.479$ & $.284$ & $.247$
    & $.293$ & $.394$ \\
\quad CKERMT base
    & -- & -- & -- & -- & -- & -- & -- & -- & -- & -- \\
\quad CKERMT task-specific\textsuperscript{\ddag}
    & $.390$ & $.387$ & $.405$ & $.396$
    & $.470$ & $.471$ & $.252$ & $.232$
    & $.286$ & $.365$ \\
\quad CKERMT upscale task-specific\textsuperscript{\dag}
    & $.382$ & $.385$ & $.401$ & $.394$
    & $.463$ & $.455$ & $.259$ & $.233$
    & $.288$ & $.362$ \\
\quad CKERMT Biogen task-specific\textsuperscript{\dag}
    & $.388$ & $.395$ & $.393$ & $.406$
    & $.457$ & $.460$ & $.257$ & $.234$
    & $.289$ & $.364$ \\
\quad CKERMT Biogen sampled-30k task-specific\textsuperscript{\dag}
    & $.393$ & $.373$ & $.413$ & $.410$
    & $.453$ & $.447$ & $.228$ & $.220$
    & $.290$ & $.359$ \\
\quad CKERMT Biogen+OA+CM task-specific\textsuperscript{\dag}
    & $.377$ & $.375$ & $.397$ & $.395$
    & $.460$ & $.459$ & $.255$ & $.226$
    & $.300$ & $.360$ \\
\quad CKERMT upscale+Biogen task-specific\textsuperscript{\dag}
    & $.388$ & $.375$ & $.411$ & $.403$
    & $.452$ & $.458$ & $.249$ & $.225$
    & $.288$ & $.361$ \\

\bottomrule
\end{tabular}
\end{sidewaystable*}
\clearpage
\begin{sidewaystable*}
\centering
\tiny
\caption{Test MAE on the ChEMBL-MT benchmark. Values marked with
$\textsuperscript{\ddag}$ are taken from Figure~15 in~\cite{DBLP:journals/corr/abs-2606-11508}. Mic-H, Mic-M, and Mic-R denote microsomal clearance in human, mouse, and rat, respectively. Tot-D, Tot-H, Tot-M, and Tot-R denote total clearance in dog, human, monkey, and rat. fu-D, fu-H, fu-M, and fu-R denote the fraction unbound in plasma for the corresponding species. VD-D, VD-H, VD-M, and VD-R denote steady-state volume of distribution. Kin-S and Therm-S denote kinetic and thermodynamic $\log S_{\mathrm{aq}}$. Mean is the Overall value reported in Figure~15. The figure reports means to three decimal places without numerical standard
deviations. Lower is better.}
\label{tab:chemblmt_error}
\setlength{\tabcolsep}{2pt}
\renewcommand{\arraystretch}{1.15}
\resizebox{\textheight}{!}{%
\begin{tabular}{l*{26}{c}}
\toprule
\textbf{Model}
& \textbf{Mic-H}
& \textbf{Mic-M}
& \textbf{Mic-R}
& \textbf{Tot-D}
& \textbf{Tot-H}
& \textbf{Tot-M}
& \textbf{Tot-R}
& \textbf{2C8}
& \textbf{2C9}
& \textbf{2D6}
& \textbf{3A4}
& \textbf{fu-D}
& \textbf{fu-H}
& $\boldsymbol{\log D}$
& \textbf{fu-M}
& \textbf{Papp}
& \textbf{P-gp}
& \textbf{fu-R}
& \textbf{VD-D}
& \textbf{VD-H}
& \textbf{VD-M}
& \textbf{VD-R}
& \textbf{hERG}
& \textbf{Kin-S}
& \textbf{Therm-S}
& \textbf{Mean} \\
\midrule

\multicolumn{27}{l}{\textit{LBGM}} \\
\quad Morgan
    & $.421$ & $.452$ & $.502$ & $.463$ & $.512$ & $.347$ & $.470$
    & $.506$ & $.570$ & $.617$ & $.649$
    & $.504$ & $.458$ & $.766$ & $.537$
    & $.476$ & $.456$ & $.504$
    & $.430$ & $.466$ & $.362$ & $.488$
    & $.514$ & $.379$ & $1.135$ & $.519$ \\
\quad RDKit
    & $.400$ & $.406$ & $.506$ & $.421$ & $.472$ & $.313$ & $.459$
    & $.513$ & $.554$ & $.589$ & $.660$
    & $.434$ & $.392$ & $.630$ & $.539$
    & $.467$ & $.436$ & $.512$
    & $.389$ & $.403$ & $.356$ & $.444$
    & $.530$ & $.369$ & $.813$ & $.480$ \\
\quad RDKit + Morgan
    & $.399$ & $.385$ & $.513$ & $.411$ & $.464$ & $.335$ & $.456$
    & $.496$ & $.556$ & $.576$ & $.633$
    & $.443$ & $.390$ & $.614$ & $.549$
    & $.457$ & $.433$ & $.504$
    & $.377$ & $.409$ & $.425$ & $.443$
    & $.509$ & $.352$ & $.809$ & $.478$ \\
\quad Morgan + RDKit + Avalon + ErG
    & $.401$ & $.442$ & $.498$ & $.453$ & $.508$ & $.317$ & $.454$
    & $.523$ & $.551$ & $.608$ & $.634$
    & $.433$ & $.402$ & $.616$ & $.507$
    & $.480$ & $.431$ & $.516$
    & $.421$ & $.424$ & $.416$ & $.468$
    & $.506$ & $.361$ & $.840$ & $.488$ \\
\quad Mol-JEPA CLS
    & $.431$ & $.415$ & $.542$ & $.417$ & $.506$ & $.304$ & $.449$
    & $.515$ & $.567$ & $.605$ & $.665$
    & $.483$ & $.449$ & $.772$ & $.573$
    & $.534$ & $.438$ & $.468$
    & $.442$ & $.445$ & $.392$ & $.477$
    & $.545$ & $.389$ & $1.149$ & $.519$ \\
\rowcolor{atomjepablue}
\quad Atom-JEPA Last Layers
    & $.455$ & $.436$ & $.531$ & $.413$ & $.506$ & $.333$ & $.445$
    & $.509$ & $.602$ & $.635$ & $.729$
    & $.479$ & $.436$ & $.868$ & $.563$
    & $.536$ & $.467$ & $.518$
    & $.419$ & $.463$ & $.459$ & $.465$
    & $.581$ & $.407$ & $.932$ & $.527$ \\
    \rowcolor{atomjepablue}
\quad Atom-JEPA All Layers
    & $.425$ & $.431$ & $.502$ & $.413$ & $.496$ & $.339$ & $.450$
    & $.519$ & $.574$ & $.606$ & $.673$
    & $.454$ & $.399$ & $.751$ & $.600$
    & $.508$ & $.445$ & $.491$
    & $.428$ & $.436$ & $.420$ & $.457$
    & $.541$ & $.382$ & $.895$ & $.505$ \\

\multicolumn{27}{l}{\textit{MLP}} \\
  \quad Morgan
    & $.443$ & $.503$ & $.467$ & $.451$ & $.548$ & $.385$ & $.476$
    & $.480$ & $.597$ & $.672$ & $.669$
    & $.460$ & $.478$ & $.842$ & $.475$
    & $.545$ & $.495$ & $.468$
    & $.465$ & $.459$ & $.488$ & $.471$
    & $.598$ & $.442$ & $1.365$ & $.550$ \\
    
  \quad RDKit
    & $.413$ & $.556$ & $.566$ & $.437$ & $.481$ & $.320$ & $.421$
    & $.585$ & $.591$ & $.651$ & $.688$
    & $.460$ & $.408$ & $.666$ & $.391$
    & $.502$ & $.430$ & $.477$
    & $.365$ & $.430$ & $.336$ & $.450$
    & $.559$ & $.398$ & $.893$ & $.499$ \\
    
\quad RDKit + Morgan
    & $.513$ & $.485$ & $.560$ & $.467$ & $.553$ & $.346$ & $.479$
    & $.528$ & $.599$ & $.677$ & $.730$
    & $.598$ & $.681$ & $1.040$ & $.581$
    & $.613$ & $.507$ & $.558$
    & $.542$ & $.514$ & $.526$ & $.556$
    & $.672$ & $.554$ & $1.818$ & $.628$ \\
    
\quad Morgan + RDKit + Avalon + ErG
    & $.510$ & $.480$ & $.559$ & $.459$ & $.548$ & $.347$ & $.476$
    & $.521$ & $.599$ & $.675$ & $.729$
    & $.554$ & $.623$ & $.981$ & $.598$
    & $.606$ & $.508$ & $.512$
    & $.517$ & $.491$ & $.506$ & $.521$
    & $.672$ & $.485$ & $1.391$ & $.595$ \\
    
\quad Mol-JEPA cls
    & $.428$ & $.460$ & $.553$ & $.388$ & $.513$ & $.394$ & $.446$
    & $.564$ & $.560$ & $.619$ & $.652$
    & $.410$ & $.382$ & $.682$ & $.363$
    & $.543$ & $.451$ & $.389$
    & $.383$ & $.427$ & $.414$ & $.461$
    & $.550$ & $.378$ & $1.019$ & $.497$ \\

\quad Mol-JEPA Modalities
    & $.405$ & $.444$ & $.545$ & $.395$ & $.510$ & $.358$ & $.428$
    & $.536$ & $.533$ & $.584$ & $.647$
    & $.477$ & $.388$ & $.619$ & $.445$
    & $.505$ & $.444$ & $.434$
    & $.379$ & $.411$ & $.381$ & $.460$
    & $.521$ & $.369$ & $.979$ & $.488$ \\
    
\rowcolor{atomjepablue}
\quad Atom-JEPA Last Layers
    & $.443$ & $.482$ & $.559$ & $.390$ & $.478$ & $.337$ & $.435$
    & $.492$ & $.597$ & $.614$ & $.716$
    & $.437$ & $.415$ & $.737$ & $.523$
    & $.544$ & $.457$ & $.495$
    & $.435$ & $.476$ & $.416$ & $.454$
    & $.572$ & $.395$ & $.861$ & $.510$ \\
\rowcolor{atomjepablue}
\quad Atom-JEPA All Layers
    & $.451$ & $.459$ & $.577$ & $.402$ & $.505$ & $.359$ & $.447$
    & $.502$ & $.597$ & $.617$ & $.711$
    & $.437$ & $.412$ & $.746$ & $.471$
    & $.556$ & $.446$ & $.462$
    & $.490$ & $.482$ & $.411$ & $.468$
    & $.578$ & $.405$ & $.948$ & $.518$ \\

\addlinespace
\multicolumn{8}{l}{\textit{Finetuning}} \\
\rowcolor{atomjepablue}
\quad No Pretrain & $.376$ & $.417$ & $.473$ & $.414$ & $.488$ & $.301$ & $.443$ & $.472$ & $.565$ & $.575$ & $.585$ & $.442$ & $.382$ & $.442$ & $.384$ & $.457$ & $.443$ & $.461$ & $.561$ & $.281$ & $.755$ & $.406$ & $.407$ & $.400$ & $.427$ & $.454$ \\
\rowcolor{atomjepablue}
\quad Atom-JEPA & $.365$ & $.447$ & $.472$ & $.402$ & $.472$ & $.319$ & $.436$ & $.451$ & $.534$ & $.607$ & $.592$ & $.420$ & $.378$ & $.409$ & $.390$ & $.471$ & $.459$ & $.484$ & $.560$ & $.279$ & $.781$ & $.422$ & $.416$ & $.423$ & $.427$ & $.457$ \\
\rowcolor{atomjepablue}
\quad Atom-JEPA (10-conf. ensemble) & $.360$ & $.443$ & $.468$ & $.401$ & $.470$ & $.316$ & $.433$ & $.447$ & $.527$ & $.599$ & $.585$ & $.419$ & $.374$ & $.408$ & $.388$ & $.466$ & $.453$ & $.477$ & $.552$ & $.274$ & $.774$ & $.420$ & $.414$ & $.420$ & $.426$ & $.453$ \\

\addlinespace
\multicolumn{8}{l}{\textit{Literature models}} \\

\quad Chemprop
    & $.368$ & $.490$ & $.476$ & $.427$ & $.483$ & $.377$ & $.465$
    & $.494$ & $.534$ & $.621$ & $.626$
    & $.429$ & $.368$ & $.536$ & $.438$
    & $.409$ & $.449$ & $.431$
    & $.498$ & $.401$ & $.452$ & $.472$
    & $.484$ & $.271$ & $.920$ & $.477$ \\

\quad CKERMT Our runs & $.361$ & $.464$ & $.459$ & $.420$ & $.484$ & $.317$ & $.463$ & $.492$ & $.509$ & $.586$ & $.582$ & $.395$ & $.350$ & $.401$ & $.443$ & $.408$ & $.414$ & $.470$ & $.516$ & $.266$ & $.869$ & $.422$ & $.398$ & $.354$ & $.463$ & $.452$ \\

\quad KERMT base\textsuperscript{\dag}
    & $.357$ & $.441$ & $.439$ & $.454$ & $.482$ & $.414$ & $.482$
    & $.486$ & $.509$ & $.576$ & $.604$
    & $.409$ & $.376$ & $.514$ & $.442$
    & $.426$ & $.430$ & $.449$
    & $.417$ & $.408$ & $.369$ & $.454$
    & $.480$ & $.287$ & $.959$ & $.466$ \\

\quad KERMT task-specific\textsuperscript{\dag}
    & $.360$ & $.435$ & $.455$ & $.460$ & $.473$ & $.353$ & $.488$
    & $.495$ & $.512$ & $.573$ & $.598$
    & $.413$ & $.361$ & $.525$ & $.436$
    & $.428$ & $.424$ & $.418$
    & $.405$ & $.387$ & $.371$ & $.460$
    & $.476$ & $.276$ & $.910$ & $.460$ \\

\quad CMIM only task-specific\textsuperscript{\dag}
    & $.369$ & $.433$ & $.467$ & $.439$ & $.456$ & $.324$ & $.423$
    & $.514$ & $.512$ & $.593$ & $.606$
    & $.388$ & $.376$ & $.576$ & $.420$
    & $.448$ & $.432$ & $.432$
    & $.402$ & $.385$ & $.354$ & $.447$
    & $.491$ & $.293$ & $.856$ & $.458$ \\

\quad CKERMT task-specific\textsuperscript{\dag}
    & $.362$ & $.442$ & $.432$ & $.423$ & $.471$ & $.326$ & $.453$
    & $.499$ & $.506$ & $.592$ & $.586$
    & $.414$ & $.348$ & $.515$ & $.394$
    & $.424$ & $.427$ & $.415$
    & $.409$ & $.383$ & $.360$ & $.455$
    & $.475$ & $.267$ & $.878$ & $.450$ \\

\quad CKERMT upscale task-specific\textsuperscript{\dag}
    & $.354$ & $.433$ & $.439$ & $.425$ & $.479$ & $.338$ & $.442$
    & $.490$ & $.495$ & $.568$ & $.582$
    & $.402$ & $.348$ & $.507$ & $.401$
    & $.412$ & $.420$ & $.410$
    & $.406$ & $.393$ & $.370$ & $.448$
    & $.469$ & $.265$ & $.834$ & $.445$ \\

\quad CKERMT Biogen task-specific\textsuperscript{\dag}
    & $.360$ & $.446$ & $.436$ & $.424$ & $.466$ & $.330$ & $.445$
    & $.499$ & $.508$ & $.578$ & $.580$
    & $.414$ & $.348$ & $.516$ & $.388$
    & $.413$ & $.425$ & $.417$
    & $.404$ & $.388$ & $.366$ & $.442$
    & $.474$ & $.269$ & $.885$ & $.449$ \\

\quad \shortstack[l]{CKERMT Biogen sampled-30k\\task-specific}
    \textsuperscript{\dag}
    & $.361$ & $.471$ & $.436$ & $.421$ & $.474$ & $.332$ & $.447$
    & $.494$ & $.500$ & $.576$ & $.571$
    & $.427$ & $.343$ & $.510$ & $.399$
    & $.407$ & $.429$ & $.397$
    & $.404$ & $.387$ & $.363$ & $.444$
    & $.459$ & $.263$ & $.816$ & $.445$ \\

\quad \shortstack[l]{CKERMT Biogen+OA+CM\\task-specific}
    \textsuperscript{\dag}
    & $.355$ & $.443$ & $.432$ & $.420$ & $.473$ & $.345$ & $.446$
    & $.480$ & $.499$ & $.572$ & $.569$
    & $.409$ & $.349$ & $.506$ & $.370$
    & $.410$ & $.420$ & $.402$
    & $.400$ & $.386$ & $.356$ & $.452$
    & $.469$ & $.262$ & $.814$ & $.442$ \\

\quad \shortstack[l]{CKERMT upscale+Biogen\\task-specific}
    \textsuperscript{\dag}
    & $.365$ & $.443$ & $.436$ & $.413$ & $.467$ & $.345$ & $.456$
    & $.490$ & $.496$ & $.582$ & $.588$
    & $.417$ & $.353$ & $.506$ & $.394$
    & $.422$ & $.426$ & $.412$
    & $.392$ & $.383$ & $.347$ & $.448$
    & $.467$ & $.265$ & $.841$ & $.446$ \\

\bottomrule
\end{tabular}%
}
\end{sidewaystable*}
\clearpage

\subsection{TDC ADMET}
\begin{table}[H]
\centering
\tiny
\setlength{\tabcolsep}{2pt}
\renewcommand{\arraystretch}{1.15}
\caption{
Results on the TDC-ADMET benchmark: absorption.
Columns correspond, from left to right, to
\texttt{caco2\_wang},
\texttt{hia\_hou},
\texttt{pgp\_broccatelli},
\texttt{bioavailability\_ma},
\texttt{lipophilicity\_astrazeneca}, and
\texttt{solubility\_aqsoldb}.
Numbers from \cite{koleiev2026end}.
}
\begin{tabular}{l*{6}{c}}
\toprule
\textbf{Model} & \textbf{Caco2} & \textbf{HIA} & \textbf{P-gp} & \textbf{Bioav.} & \textbf{Lipo.} & \textbf{Solub.} \\
\textbf{Unit} & \textbf{MAE$\downarrow$} & \textbf{AUROC$\uparrow$} & \textbf{AUROC$\uparrow$} & \textbf{AUROC$\uparrow$} & \textbf{MAE$\downarrow$} & \textbf{MAE$\downarrow$} \\
\midrule
\multicolumn{7}{l}{\textit{GBDT models}} \\
\quad Morgan
& $.3530 \err{.0370}$
& $.8170 \err{.0300}$
& $.8650 \err{.0230}$
& $.5690 \err{.0730}$
& $.6060 \err{.0170}$
& $.9700 \err{.0060}$ \\
\quad RDKit
& $.3020 \err{.0110}$
& $.8900 \err{.0500}$
& $.8690 \err{.0190}$
& $.5800 \err{.0430}$
& $.5690 \err{.0110}$
& $.7630 \err{.0110}$ \\
\quad Morgan + RDKit
& $.2950 \err{.0060}$
& $.8940 \err{.0400}$
& $.8590 \err{.0080}$
& $.5960 \err{.0550}$
& $.5420 \err{.0050}$
& $.7760 \err{.0280}$ \\
\quad Morgan + RDKit + Avalon + ErG
& $.3290 \err{.0150}$
& $.9410 \err{.0170}$
& $.8690 \err{.0160}$
& $.5790 \err{.0770}$
& $.5400 \err{.0110}$
& $.8030 \err{.0050}$ \\
\rowcolor{atomjepablue}
\quad Atom-JEPA Last Layer
& $.3390 \err{.0040}$
& $.9380 \err{.0330}$
& $.8760 \err{.0110}$
& $.6100 \err{.0560}$
& $.7640 \err{.0190}$
& $.9270 \err{.0150}$ \\
\rowcolor{atomjepablue}
\quad Atom-JEPA All Layers
& $.3090 \err{.0150}$
& $.9410 \err{.0280}$
& $.8670 \err{.0210}$
& $.5850 \err{.0550}$
& $.6490 \err{.0050}$
& $.8620 \err{.0130}$ \\
\midrule
\multicolumn{7}{l}{\textit{MLP models}} \\
\quad Morgan
& $.4410 \err{.0050}$
& $.8940 \err{.0050}$
& $.8900 \err{.0240}$
& $.5990 \err{.0540}$
& $.6720 \err{.0080}$
& $1.1700 \err{.0080}$ \\
\quad RDKit
& $.3590 \err{.0100}$
& $.9790 \err{.0050}$
& $.8990 \err{.0050}$
& $.6890 \err{.0080}$
& $.5640 \err{.0080}$
& $.7930 \err{.0380}$ \\
\quad RDKit + Morgan
& $.3160 \err{.0100}$
& $.9400 \err{.0180}$
& $.9120 \err{.0130}$
& $.6730 \err{.0240}$
& $.5310 \err{.0020}$
& $.8320 \err{.0060}$ \\
\quad RDKit + Morgan + Avalon + ErG
& $.3620 \err{.1170}$
& $.9490 \err{.0040}$
& $.9280 \err{.0030}$
& $.6060 \err{.0540}$
& $.6800 \err{.2390}$
& $.9390 \err{.2730}$ \\
\rowcolor{atomjepablue}
\quad Atom-JEPA Last Layer
& $.2970 \err{.0050}$
& $.9440 \err{.0140}$
& $.8850 \err{.0090}$
& $.6370 \err{.0330}$
& $.6310 \err{.0100}$
& $.8000 \err{.0130}$ \\
\rowcolor{atomjepablue}
\quad Atom-JEPA All Layers
& $.2970 \err{.0010}$
& $.9600 \err{.0190}$
& $.8810 \err{.0070}$
& $.6040 \err{.0150}$
& $.5430 \err{.0230}$
& $.7720 \err{.0190}$ \\
\midrule
\multicolumn{7}{l}{\textit{Other models}} \\
\quad MapLight
& $.276 \err{.005}$
& $.983 \err{.001}$
& $.932 \err{.001}$
& $.741 \err{.007}$
& $.542 \err{.003}$
& $.787 \err{.006}$ \\
\quad MapLight+GNN
& $.290 \err{.008}$
& $.989 \err{.001}$
& $.938 \err{.0}$
& $.740 \err{.01}$
& $.525 \err{.003}$
& $.795 \err{.004}$ \\
\quad CaliciBoost
& $.256 \err{.006}$
& --
& --
& --
& --
& -- \\
\rowcolor{atomjepablue}
\quad Atom-JEPA
& $.2910 \err{.0100}$
& $.9800 \err{.0100}$
& $.9180 \err{.0100}$
& $.6900 \err{.0370}$
& $.3970 \err{.0030}$
& $.7800 \err{.0180}$ \\
\bottomrule
\end{tabular}%
\label{tab:tdc-admet-absorption}
\end{table}

\begin{table}[H]
\centering
\tiny
\caption{
Results on the TDC-ADMET benchmark: distribution.
Columns correspond, from left to right, to
\texttt{bbb\_martins},
\texttt{ppbr\_az}, and
\texttt{vdss\_lombardo}.
Numbers from \cite{koleiev2026end}. RDKit + Morgan and RDKit + Morgan + Avalon + ErG MLP diverged during training on VDSS Lombardo
}
\label{tab:tdc-admet-distribution}
\setlength{\tabcolsep}{2pt}
\renewcommand{\arraystretch}{1.15}
\begin{tabular}{l*{3}{c}}
\toprule
\textbf{Model} & \textbf{BBB} & \textbf{PPBR} & \textbf{VDss} \\
\textbf{Unit} & \textbf{AUROC$\uparrow$} & \textbf{MAE$\downarrow$} & \textbf{Spear$\uparrow$} \\
\midrule
\multicolumn{4}{l}{\textit{GBDT models}} \\
\quad Morgan
& $.8300 \err{.0110}$
& $8.6780 \err{.1380}$
& $.6070 \err{.0230}$ \\
\quad RDKit
& $.8770 \err{.0150}$
& $7.5460 \err{.1110}$
& $.6220 \err{.0090}$ \\
\quad Morgan + RDKit
& $.8600 \err{.0140}$
& $7.5100 \err{.0980}$
& $.6250 \err{.0260}$ \\
\quad Morgan + RDKit + Avalon + ErG
& $.8600 \err{.0260}$
& $8.1450 \err{.3010}$
& $.5750 \err{.0230}$ \\
\rowcolor{atomjepablue}
\quad Atom-JEPA Last Layer
& $.8540 \err{.0080}$
& $8.8830 \err{.1440}$
& $.5800 \err{.0260}$ \\
\rowcolor{atomjepablue}
\quad Atom-JEPA All Layers
& $.8500 \err{.0210}$
& $8.1380 \err{.0690}$
& $.5840 \err{.0400}$ \\
\midrule
\multicolumn{4}{l}{\textit{MLP models}} \\
\quad Morgan
& $.8380 \err{.0110}$
& $8.5710 \err{.0550}$
& $.5070 \err{.0460}$ \\
\quad RDKit
& $.8910 \err{.0140}$
& $7.6240 \err{.0730}$
& $.5550 \err{.0110}$ \\
\quad RDKit + Morgan
& $.9040 \err{.0060}$
& $8.1030 \err{.0880}$
& $nan \err{nan}$ \\
\quad RDKit + Morgan + Avalon + ErG
& $.9020 \err{.0090}$
& $9.9660 \err{1.4280}$
& $nan \err{nan}$ \\
\rowcolor{atomjepablue}
\quad Atom-JEPA Last Layer
& $.8690 \err{.0150}$
& $8.1870 \err{.1250}$
& $.5370 \err{.0430}$ \\
\rowcolor{atomjepablue}
\quad Atom-JEPA All Layers
& $.8920 \err{.0230}$
& $7.9370 \err{.0390}$
& $.5770 \err{.0460}$ \\
\midrule
\multicolumn{4}{l}{\textit{Other models}} \\
\quad MapLight
& $.920 \err{.002}$
& $7.622 \err{.065}$
& $.716 \err{.005}$ \\
\quad MapLight+GNN
& $.916 \err{.002}$
& $7.573 \err{.163}$
& $.716 \err{.006}$ \\
\rowcolor{atomjepablue}
\quad Atom-JEPA
& $.8960 \err{.0150}$
& $7.6690 \err{.2530}$
& $.6800 \err{.0360}$ \\
\bottomrule
\end{tabular}%
\end{table}

\begin{table}[H]
\centering
\tiny
\caption{
Results on the TDC-ADMET benchmark: metabolism.
Columns correspond, from left to right, to
\texttt{cyp2c9\_veith},
\texttt{cyp2d6\_veith},
\texttt{cyp3a4\_veith},
\texttt{cyp2c9\_substrate\_carbonmangels},
\texttt{cyp2d6\_substrate\_carbonmangels}, and
\texttt{cyp3a4\_substrate\_carbonmangels}.
Numbers from \cite{koleiev2026end}.
}
\label{tab:tdc-admet-metabolism}
\setlength{\tabcolsep}{2pt}
\renewcommand{\arraystretch}{1.15}

\begin{tabular}{l*{6}{c}}
\toprule
\textbf{Model} & \textbf{2C9-I} & \textbf{2D6-I} & \textbf{3A4-I} & \textbf{2C9-S} & \textbf{2D6-S} & \textbf{3A4-S} \\
\textbf{Unit} & \textbf{AUPRC$\uparrow$} & \textbf{AUPRC$\uparrow$} & \textbf{AUPRC$\uparrow$} & \textbf{AUPRC$\uparrow$} & \textbf{AUPRC$\uparrow$} & \textbf{AUROC$\uparrow$} \\
\midrule
\multicolumn{7}{l}{\textit{GBDT models}} \\
\quad Morgan
& $.6730 \err{.0590}$
& $.5300 \err{.0100}$
& $.7800 \err{.0050}$
& $.2810 \err{.0000}$
& $.4660 \err{.0960}$
& $.5990 \err{.0190}$ \\
\quad RDKit
& $.7110 \err{.0090}$
& $.5880 \err{.0150}$
& $.7840 \err{.0100}$
& $.2970 \err{.0250}$
& $.5420 \err{.1280}$
& $.6170 \err{.0280}$ \\
\quad Morgan + RDKit
& $.7120 \err{.0100}$
& $.6000 \err{.0110}$
& $.8090 \err{.0060}$
& $.2810 \err{.0090}$
& $.5560 \err{.0500}$
& $.6310 \err{.0090}$ \\
\quad Morgan + RDKit + Avalon + ErG
& $.6930 \err{.0120}$
& $.5630 \err{.0290}$
& $.8100 \err{.0110}$
& $.2930 \err{.0150}$
& $.5140 \err{.1230}$
& $.6210 \err{.0300}$ \\
\rowcolor{atomjepablue}
\quad Atom-JEPA Last Layer
& $.6450 \err{.0750}$
& $.3360 \err{.0790}$
& $.7740 \err{.0100}$
& $.3090 \err{.0370}$
& $.5330 \err{.0780}$
& $.5880 \err{.0390}$ \\
\rowcolor{atomjepablue}
\quad Atom-JEPA All Layers
& $.7000 \err{.0180}$
& $.5100 \err{.0050}$
& $.7950 \err{.0040}$
& $.2940 \err{.0170}$
& $.5170 \err{.1090}$
& $.6130 \err{.0340}$ \\
\midrule
\multicolumn{7}{l}{\textit{MLP models}} \\
\quad Morgan
& $.7110 \err{.0050}$
& $.6040 \err{.0050}$
& $.8320 \err{.0050}$
& $.4320 \err{.0330}$
& $.6420 \err{.0370}$
& $.6240 \err{.0130}$ \\
\quad RDKit
& $.7390 \err{.0130}$
& $.6260 \err{.0030}$
& $.8350 \err{.0050}$
& $.3900 \err{.0140}$
& $.6680 \err{.0290}$
& $.5990 \err{.0160}$ \\
\quad RDKit + Morgan
& $.7510 \err{.0040}$
& $.6520 \err{.0050}$
& $.8700 \err{.0020}$
& $.4310 \err{.0280}$
& $.7200 \err{.0210}$
& $.6290 \err{.0470}$ \\
\quad RDKit + Morgan + Avalon + ErG
& $.7460 \err{.0200}$
& $.5660 \err{.2060}$
& $.8450 \err{.0640}$
& $.4220 \err{.0560}$
& $.6280 \err{.1530}$
& $.5840 \err{.0500}$ \\
\rowcolor{atomjepablue}
\quad Atom-JEPA Last Layer
& $.7560 \err{.0060}$
& $.6480 \err{.0060}$
& $.8400 \err{.0050}$
& $.3680 \err{.0190}$
& $.5870 \err{.0530}$
& $.5950 \err{.0230}$ \\
\rowcolor{atomjepablue}
\quad Atom-JEPA All Layers
& $.7620 \err{.0040}$
& $.6470 \err{.0050}$
& $.8480 \err{.0020}$
& $.4150 \err{.0710}$
& $.6420 \err{.0300}$
& $.5900 \err{.0120}$ \\
\midrule
\multicolumn{7}{l}{\textit{Other models}} \\
\quad MapLight
& $.786 \err{.004}$
& $.720 \err{.002}$
& $.881 \err{.001}$
& $.411 \err{.005}$
& $.718 \err{.008}$
& $.649 \err{.006}$ \\
\quad MapLight+GNN
& $.859 \err{.001}$
& $.790 \err{.001}$
& $.916 \err{.00}$
& $.446 \err{.015}$
& $.716 \err{.007}$
& $.642 \err{.004}$ \\
\rowcolor{atomjepablue}
\quad Atom-JEPA
& $.7720 \err{.0060}$
& $.6610 \err{.0180}$
& $.8680 \err{.0050}$
& $.4560 \err{.0270}$
& $.6840 \err{.0550}$
& $.5790 \err{.0230}$ \\
\bottomrule
\end{tabular}%
\end{table}

\begin{table}[H]
\centering
\tiny
\caption{
Results on the TDC-ADMET benchmark: excretion.
Columns correspond, from left to right, to
\texttt{half\_life\_obach},
\texttt{clearance\_microsome\_az}, and
\texttt{clearance\_hepatocyte\_az}.
Numbers from \cite{koleiev2026end}.
}
\label{tab:tdc-admet-excretion}
\setlength{\tabcolsep}{2pt}
\renewcommand{\arraystretch}{1.15}
\begin{tabular}{l*{3}{c}}
\toprule
\textbf{Model} & \textbf{Half-life} & \textbf{Cl-Mic} & \textbf{Cl-Hep} \\
\textbf{Unit} & \textbf{Spear$\uparrow$} & \textbf{Spear$\uparrow$} & \textbf{Spear$\uparrow$} \\
\midrule
\multicolumn{4}{l}{\textit{GBDT models}} \\
\quad Morgan
& $.4130 \err{.0430}$
& $.4670 \err{.0290}$
& $.3690 \err{.0250}$ \\
\quad RDKit
& $.3520 \err{.0350}$
& $.5710 \err{.0310}$
& $.4030 \err{.0480}$ \\
\quad Morgan + RDKit
& $.4150 \err{.0370}$
& $.5510 \err{.0190}$
& $.4270 \err{.0330}$ \\
\quad Morgan + RDKit + Avalon + ErG
& $.3810 \err{.0440}$
& $.4620 \err{.0680}$
& $.3750 \err{.0190}$ \\
\rowcolor{atomjepablue}
\quad Atom-JEPA Last Layer
& $.4120 \err{.1510}$
& $.4560 \err{.0470}$
& $.2500 \err{.0390}$ \\
\rowcolor{atomjepablue}
\quad Atom-JEPA All Layers
& $.4310 \err{.0740}$
& $.4980 \err{.0280}$
& $.2480 \err{.0760}$ \\
\midrule
\multicolumn{4}{l}{\textit{MLP models}} \\
\quad Morgan
& $.3440 \err{.0360}$
& $.4990 \err{.0160}$
& $.2910 \err{.0520}$ \\
\quad RDKit
& $.3350 \err{.0700}$
& $.5890 \err{.0330}$
& $.3670 \err{.0180}$ \\
\quad RDKit + Morgan
& $.2550 \err{.1480}$
& $.5570 \err{.0150}$
& $.3840 \err{.0090}$ \\
\quad RDKit + Morgan + Avalon + ErG
& $.2400 \err{.1120}$
& $.4090 \err{.2030}$
& $.2290 \err{.2250}$ \\
\rowcolor{atomjepablue}
\quad Atom-JEPA Last Layer
& $.4560 \err{.0480}$
& $.5650 \err{.0310}$
& $.3460 \err{.0250}$ \\
\rowcolor{atomjepablue}
\quad Atom-JEPA All Layers
& $.3830 \err{.1100}$
& $.5450 \err{.0270}$
& $.3020 \err{.0320}$ \\
\midrule
\multicolumn{4}{l}{\textit{Other models}} \\
\quad MapLight
& $.559 \err{.011}$
& $.625 \err{.009}$
& $.469 \err{.016}$ \\
\quad MapLight+GNN
& $.563 \err{.009}$
& $.635 \err{.007}$
& $.500 \err{.012}$ \\
\quad CaliciBoost
& --
& --
& -- \\
\rowcolor{atomjepablue}
\quad Atom-JEPA
& $.4910 \err{.0900}$
& $.6020 \err{.0210}$
& $.4990 \err{.0230}$ \\
\bottomrule
\end{tabular}%
\end{table}

\begin{table}[H]
\centering
\tiny
\caption{
Results on the TDC-ADMET benchmark: toxicity.
Columns correspond, from left to right, to
\texttt{herg},
\texttt{ames},
\texttt{dili}, and
\texttt{ld50\_zhu}.
Numbers from \cite{koleiev2026end}.
}
\label{tab:tdc-admet-toxicity}
\setlength{\tabcolsep}{2pt}
\renewcommand{\arraystretch}{1.15}

\begin{tabular}{l*{4}{c}}
\toprule
\textbf{Model} & \textbf{hERG} & \textbf{AMES} & \textbf{DILI} & \textbf{LD50} \\
\textbf{Unit} & \textbf{AUROC$\uparrow$} & \textbf{AUROC$\uparrow$} & \textbf{AUROC$\uparrow$} & \textbf{MAE$\downarrow$} \\
\midrule
\multicolumn{5}{l}{\textit{GBDT models}} \\
\quad Morgan
& $.8100 \err{.0180}$
& $.7690 \err{.0170}$
& $.8640 \err{.0170}$
& $.6750 \err{.0170}$ \\
\quad RDKit
& $.7910 \err{.0220}$
& $.8050 \err{.0050}$
& $.8920 \err{.0230}$
& $.6370 \err{.0120}$ \\
\quad Morgan + RDKit
& $.8120 \err{.0170}$
& $.8160 \err{.0100}$
& $.8870 \err{.0180}$
& $.6500 \err{.0090}$ \\
\quad Morgan + RDKit + Avalon + ErG
& $.7680 \err{.0530}$
& $.8050 \err{.0130}$
& $.8300 \err{.0470}$
& $.6400 \err{.0130}$ \\
\rowcolor{atomjepablue}
\quad Atom-JEPA Last Layer
& $.7600 \err{.0210}$
& $.7660 \err{.0110}$
& $.8830 \err{.0480}$
& $.6970 \err{.0090}$ \\
\rowcolor{atomjepablue}
\quad Atom-JEPA All Layers
& $.8080 \err{.0160}$
& $.8050 \err{.0050}$
& $.9130 \err{.0390}$
& $.6540 \err{.0100}$ \\
\midrule
\multicolumn{5}{l}{\textit{MLP models}} \\
\quad Morgan
& $.7770 \err{.0150}$
& $.7920 \err{.0140}$
& $.8520 \err{.0180}$
& $.6670 \err{.0120}$ \\
\quad RDKit
& $.8360 \err{.0040}$
& $.8160 \err{.0070}$
& $.8480 \err{.0310}$
& $.6750 \err{.0130}$ \\
\quad RDKit + Morgan
& $.8170 \err{.0300}$
& $.8380 \err{.0190}$
& $.8590 \err{.0100}$
& $.6030 \err{.0150}$ \\
\quad RDKit + Morgan + Avalon + ErG
& $.6460 \err{.2500}$
& $.8330 \err{.0080}$
& $.8290 \err{.0210}$
& $.5530 \err{.0210}$ \\
\rowcolor{atomjepablue}
\quad Atom-JEPA Last Layer
& $.7930 \err{.0080}$
& $.8190 \err{.0070}$
& $.9270 \err{.0090}$
& $.6520 \err{.0110}$ \\
\rowcolor{atomjepablue}
\quad Atom-JEPA All Layers
& $.8250 \err{.0250}$
& $.8320 \err{.0020}$
& $.9070 \err{.0020}$
& $.6230 \err{.0210}$ \\
\midrule
\multicolumn{5}{l}{\textit{Other models}} \\
\quad MapLight
& $.873 \err{.004}$
& $.865 \err{.001}$
& $.883 \err{.003}$
& $.618 \err{.005}$ \\
\quad MapLight+GNN
& $.880 \err{.003}$
& $.872 \err{.001}$
& $.918 \err{.005}$
& $.563 \err{.009}$ \\
\rowcolor{atomjepablue}
\quad Atom-JEPA
& $.8450 \err{.0190}$
& $.8400 \err{.0120}$
& $.8980 \err{.0180}$
& $.5590 \err{.0100}$ \\
\bottomrule
\end{tabular}%
\end{table}
\clearpage

\section{Effect of Pretraining Objectives and Data Scale Full Results}
\label{app:qm9_probe_results}
Table~\ref{tab:qm9_probe_results} shows the full results for the MLP probe evaluation of each pretrained encoder as described in Section~\ref{sec:ablation_studies}. The pretraining configurations differ to some degree between QM9 and Uni-Mol to accommodate differences in dataset size and molecular composition, as detailed in Table~\ref{tab:pretraining_hparams}. These differences include architectural, graph-partitioning, and optimization hyperparameters. Within QM9, the three objective variants share the exact same pretraining hyperparameters. Within Uni-Mol, the atom-only encoder was pretrained using multiple nodes to speed up training time, with the global batch size increased from 128 to 384 and the maximum learning rate increased from $1.4\times10^{-4}$ to $2.42\times10^{-4}$, approximately following square-root scaling with batch size. This adjustment accommodates the distributed training configuration. 

All encoders are evaluated using the same all-layer MLP probing protocol on frozen features. The results therefore support the observed performance trends across the evaluated configurations, including the stronger relative performance of the joint objective with Uni-Mol pretraining and the broad improvements from pretraining on the larger dataset. 

\begin{table*}[htbp]
\caption{All-layer MLP probes on frozen encoder features pretrained on the QM9 dataset and Unimol dataset respectively across combinations of the pretraining objectives. The evaluation metric is MAE
$(\downarrow)$. \textbf{Bold:} best result within each pretraining group;
\underline{\textbf{bold and underline}}: best overall.
``atom'' and ``sub'' denote atom-level and substructure-level objectives,
respectively.}
\label{tab:qm9_probe_results}
\centering
\normalsize
\scalebox{\tablescale}{%
\begin{tabular}{lcccccccccccc}
\toprule
Task
& $\alpha$
& $\Delta\varepsilon$
& $\varepsilon_{\text{HOMO}}$
& $\varepsilon_{\text{LUMO}}$
& $\mu$
& $C_\nu$
& $G$
& $H$
& $R^2$
& $U$
& $U_0$
& ZPVE \\
Units
& $ma_0^3$
& meV
& meV
& meV
& mD
& $\frac{\text{mcal}}{\text{mol K}}$
& meV
& meV
& $ma_0^2$
& meV
& meV
& meV \\
\midrule

\multicolumn{13}{l}{\textit{Pretrained on QM9 (130k)}} \\
\quad atom+sub
& 115 & 130.6 & 86.1 & 88.0 & 56.0 & 40.8
& 22.3 & 22.9 & \underline{\textbf{396}}
& 22.6 & 22.1 & 1.8 \\

\quad atom-only
& \textbf{97} & \textbf{114.4} & \textbf{69.6} & 81.4
& \textbf{49.4} & \textbf{32.4}
& \textbf{18.4} & \textbf{18.4} & 398
& \textbf{18.4} & \textbf{18.6} & \textbf{1.5} \\

\quad sub-only
& 104 & 131.6 & 90.0 & \textbf{80.2} & 82.3 & 42.7
& 23.0 & 22.7 & 420 & 22.5 & 22.4 & 1.9 \\

\addlinespace
\multicolumn{13}{l}{\textit{Pretrained on Uni-Mol (19M)}} \\
\quad atom+sub
& \underline{\textbf{91}}
& \underline{\textbf{94.9}}
& \underline{\textbf{60.5}}
& \underline{\textbf{64.0}}
& 32.2
& \underline{\textbf{30.9}}
& \underline{\textbf{13.2}}
& \underline{\textbf{12.4}}
& \textbf{419}
& \underline{\textbf{12.4}}
& \underline{\textbf{12.4}}
& \underline{\textbf{1.4}} \\

\quad atom-only
& 96 & 95.7 & 63.3 & 64.9
& \underline{\textbf{29.2}}
& 31.4 & 13.4 & 13.0 & 464 & 13.3 & 13.6
& \underline{\textbf{1.4}} \\

\bottomrule
\end{tabular}%
}
\end{table*}

\section{Pretraining Implementation Details}
\label{app:pretrain_details}
\subsection{Hyperparameters}
Table~\ref{tab:pretraining_hparams} contains the hyperparameters used for
the pretraining on Uni-Mol and Alexandria as well as the hyperparameters used for pretraining the three encoders on QM9, which we use for the ablation studies on prediction objectives and data scale. We use
EquiformerV3~\citep{eqv3} for the context encoder, target encoder, and predictor throughout,
following the notation of Equiformer V3 for the architectural
hyperparameters. The context and target encoders use identical
architectures. The predictor uses the same configuration as the encoders
except for the hyperparameters explicitly listed in the predictor section
of the table. In particular, the predictor GNN body is more shallow with 3 transformer blocks instead of 8 and a feature dimension of 128 rather than 256. 

Pretraining was performed on 8 NVIDIA H100 80GB HBM3 GPUs per run. The Uni-Mol run took approximately 309 hours (~2,475 GPU-hours) for 1,364,000 training steps, and the Alexandria run took approximately 268 hours (~2,147 GPU-hours) for 1,321,000 training steps.

\begin{table*}[t]
    \centering
    \small
    \setlength{\tabcolsep}{5pt}
    \renewcommand{\arraystretch}{1.08}

    \caption{
        Hyperparameters used for pretraining on Uni-Mol, Alexandria, and QM9.
        We use EquiformerV3 for the context encoder, target encoder, and
        predictor. The context and target encoders use identical
        architectures. Predictor hyperparameters not explicitly listed are
        identical to those of the corresponding encoder.
    }
    \label{tab:pretraining_hparams}

    \resizebox{\textwidth}{!}{%

    \begin{tabular}{lccc}
        \toprule
        \textbf{Hyperparameter}
        & \textbf{Uni-Mol}
        & \textbf{Alexandria}
        & \textbf{QM9} \\
        \midrule

        \multicolumn{4}{l}{\textbf{Encoder}} \\
        \addlinespace[2pt]

        Maximum degree $L_{\max}$
        & 2 & 2 & 6 \\

        Maximum order $M_{\max}$
        & 2 & 2 & 2 \\

        Number of Transformer blocks
        & 8 & 8 & 8 \\

        Embedding dimension $d_{\mathrm{embed}}$
        & $(2,256)$ & $(2,256)$ & $(6,256)$ \\

        Attention hidden dimension $d_{\mathrm{attn\text{-}hidden}}$
        & $(2,64)$ & $(2,64)$ & $(6,64)$ \\

        Number of attention heads $h$
        & 8 & 8 & 8 \\

        Attention alpha dimension $d_{\mathrm{attn\text{-}\alpha}}$
        & $(0,64)$ & $(0,64)$ & $(0,64)$ \\

        Attention value dimension $d_{\mathrm{attn\text{-}value}}$
        & $(2,16)$ & $(2,16)$ & $(6,16)$ \\

        FFN hidden dimension $d_{\mathrm{ffn}}$
        & $(2,512)$ & $(2,512)$ & $(6,512)$ \\

        Attention grid resolution $(R_{\phi},R_{\theta})$
        & $(8,14)$ & $(8,14)$ & $(8,20)$ \\

        FFN grid resolution $(R_{\phi},R_{\theta})$
        & $(14,14)$ & $(14,14)$ & $(20,20)$ \\

        Radial basis function
        & Gaussian & Gaussian & Gaussian \\

        Number of radial bases
        & 128 & 128 & 128 \\

        Radial hidden scalar dimension $d_{\mathrm{edge}}$
        & $(0,128)$ & $(0,128)$ & $(0,128)$ \\

        \midrule
        \multicolumn{4}{l}{\textbf{Predictor}} \\
        \addlinespace[2pt]

        Number of Transformer blocks
        & 3 & 3 & 3 \\

        Embedding dimension $d_{\mathrm{embed}}$
        & $(2,128)$ & $(2,128)$ & $(6,128)$ \\

        Encoder-to-predictor projection
        & $(2,256)\!\rightarrow\!(2,128)$
        & $(2,256)\!\rightarrow\!(2,128)$
        & $(6,256)\!\rightarrow\!(6,128)$ \\

        Pooling readout hidden dimension
        & 128 & 128 & 128 \\

        Pooling readout output dimension
        & 256 & 256 & 256 \\

        Number of positional radial bases
        & 128 & 128 & 128 \\

        Positional radial range (\AA)
        & 12 & 12 & 12 \\

        Positional embedding dimension
        & $(2,128)$ & $(2,128)$ & $(6,128)$ \\

        Atom-head input dimension
        & 512 & 512 & 1024 \\

        Atom-head hidden dimension
        & 128 & 128 & 128 \\

        Atom-head output dimension
        & 256 & 256 & 256 \\

        \midrule
        \multicolumn{4}{l}{\textbf{Graph Construction and Partitioning}} \\
        \addlinespace[2pt]

        Radius-graph cutoff $r_c$ (\AA)
        & 6 & 6 & 6 \\

        $k$-EgoNet cutoff $r_k$ (\AA)
        & 1.8 & 3.5 & 1.8 \\

        $k$-EgoNet hops
        & $\{1,3,5,10\}$
        & $\{1,2,3,5\}$
        & $\{1,2\}$ \\

        \midrule
        \multicolumn{4}{l}{\textbf{Optimization}} \\
        \addlinespace[2pt]

        Optimizer
        & AdamW & AdamW & AdamW \\

        Learning-rate schedule
        & Cosine + linear warmup
        & Cosine + linear warmup
        & Cosine + linear warmup \\

        Warmup fraction
        & .15 & .15 & .15 \\

        Maximum learning rate
        & \makecell{$1.4\times10^{-4}$ (atom+pool)\\
                    $2.42\times10^{-4}$ (atom-only)}
        & $1.4\times10^{-4}$
        & $1.0\times10^{-4}$ \\

        Batch size
        & \makecell{128 (atom+pool)\\384 (atom-only)}
        & 128
        & 64 \\

        Number of epochs
        & 10 & 100 & 200 \\

        Weight decay
        & .04 & .04 & .04 \\

        Dropout rate
        & .0 & .0 & .0 \\

        Stochastic depth
        & .05 & .05 & .05 \\

        Gradient clipping norm
        & 100 & 100 & 100 \\

        Target EMA decay $\tau$
        & $.995 \rightarrow 1.0$ (linear)
        & $.995 \rightarrow 1.0$ (linear)
        & $.995 \rightarrow 1.0$ (linear) \\

        \bottomrule
    \end{tabular}%
    }
\end{table*}

\subsection{Representation Collapse Metrics}
To monitor representation collapse during pretraining, we track RankMe~\citep{rankme} and $\alpha$-reQ~\citep{alphareq} computed on the pooled representations of the target encoder. Both metrics operate on an embedding matrix $D\in \mathbb{R}^{B\times d_{embed}}$ formed by stacking the pooled representations across the batch, including both evaluation directions of the symmetric loss (i.e., each partition acting once as $G_y$ against the other as context, and vice versa), gathered across all 8 GPUs used for pretraining. With a per-GPU batch size of 16, this gives $B=256$, matching the target encoder's feature dimension $d_{embed}=256$. RankMe is an estimate of the effective rank of $D$ via the Shannon entropy of its singular value distribution, bounded by $\min{(B,d_{embed})}=256$ in our setting. A RankMe value collapsing towards 1 indicates that the target encoder is producing near-constant or low-dimensional outputs, which would mean representation collapse. $\alpha$-ReQ instead fits a power-law exponent $\alpha$ to the decay of $D$'s covariance eigen spectrum, such that a large $\alpha$ is an indication of a small number of directions dominating the representation, and small $\alpha$ indicates an overly flat, noise-dominated spectrum. 

\begin{figure}[t]
    \centering
    \includegraphics[width=\linewidth]{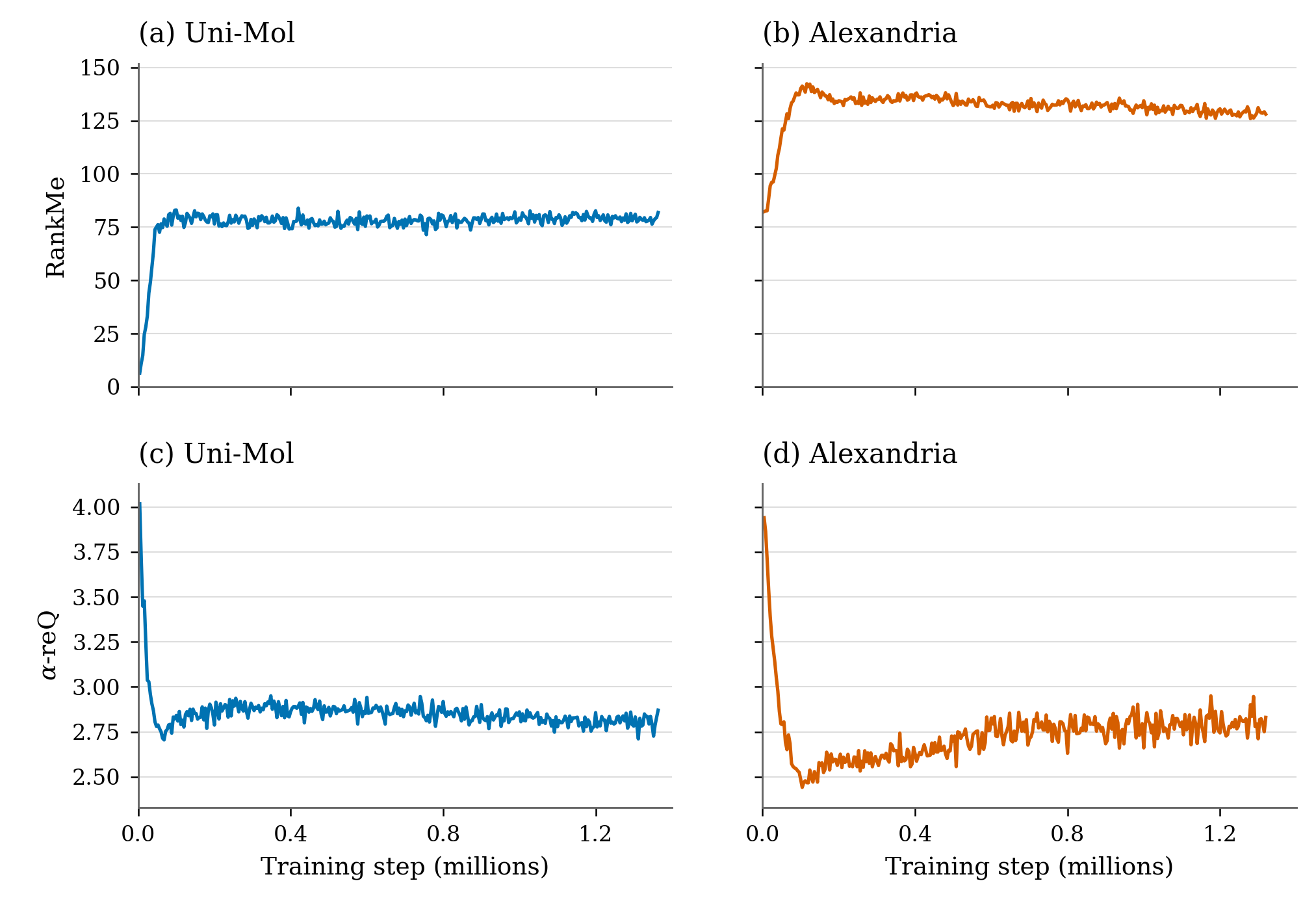}
    \caption{
        Evolution of RankMe (top) and $\alpha$-reQ (bottom)
        during pretraining for the Uni-Mol (left) and
        Alexandria (right) encoders.
    }
    \label{fig:pretraining-representation-metrics}
\end{figure}

Figure~\ref{fig:pretraining-representation-metrics} shows both metrics over the course of pretraining on the Uni-Mol and Alexandria datasets, for the joint-objective encoder. In both cases, we observe that the RankMe metric rises sharply from the initial value and then plateaus at around ($\sim$80 for Uni-Mol, $\sim$130--140 for Alexandria). On Uni-Mol $\alpha$-ReQ falls from its initial value and then stabilizes in a moderate range. For Alexandria it also falls from its initial value, and then slowly increases throughout training, remaining still within a moderate range. Overall, both metrics remain stable throughout training, indicating that the context and target encoders do not undergo representation collapse over the course of pretraining.

\begin{figure}[t]
    \centering
    \includegraphics[width=\linewidth]{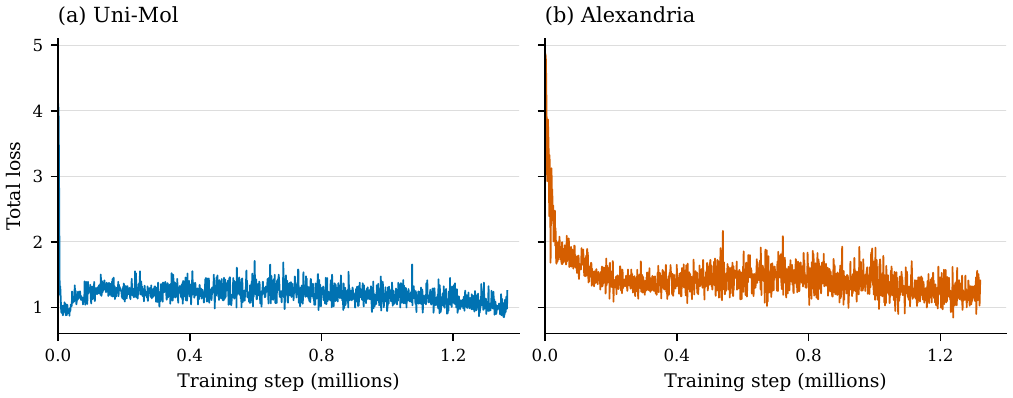}
    \caption{
        Total training loss during pretraining for the
        Uni-Mol (left) and Alexandria (right) encoders.
    }
    \label{fig:pretraining-total-loss}
\end{figure}

Figure~\ref{fig:pretraining-total-loss} shows the total loss for both runs during training. In both cases we observe a pattern of a sharp initial decline as the predictor and encoders leave initialization, followed by a period of gradual increase as the EMA target encoder drifts away form the context encoder, and finally a convergence phase towards the end of training as the target and context encoders are brought back into closer alignment, allowing the loss to decrease again.

\section{Finetuning Implementation Details} \label{app:finetune_details}

\subsection{ADMET Finetuning}
\label{appendix:admet_finetune}
\paragraph{Atom-JEPA} Due to the label imbalances and objective differences between targets present in the multi-task datasets, we follow CKERMT~\citep{DBLP:journals/corr/abs-2606-11508} and employ two tricks; an individual MLP head for each of the individual tasks, and homoscedastic uncertainty task weighting of the loss~\citep{DBLP:conf/cvpr/KendallGC18}. 
We follow the learning from~\cite{tompa2026fine}, relying on freezing the pretrained model for better generalisation, and otherwise only tune the learning rate. We deviate from their recommendations to avoid weight decay on the pre-trained model by instead using L2-SP~\citep{xuhong2018explicit} to avoid the model from drifting too far from the pretrained model.
\paragraph{LGBT and MLP Baselines} For baselines, we implement a gradient boosted tree (LGBT) and a 3-layer MLP model using different combination of features: Morgan fingerprints, RDKit features, ErG, and Avalon features. In addition, we also experiment with adding frozen Atom-JEPA features.
\paragraph{CKERMT} As \cite{DBLP:journals/corr/abs-2606-11508} does not provide test MAE of individual folds, we reran their experiments to get the uncertainty for statistical tests. We use their model with the largest self-supervised dataset called "CKERMT Biogen+OA+CM task-specific". Our results agree closely with the published mean MAEs across datasets.
\paragraph{Mol-JEPA} We also use Mol-JEPA with the same MLP and LGBT setup, either using their pooled molecule level embedding (CLS) or their full 12 modality output (modalities).
\paragraph{Chemprop} We use the Chemprop v2 version, with Biogen tuned hyperparameters from~\cite{DBLP:journals/corr/abs-2510-12719}. We also tried default parameters but found them to perform worse.

For all models except Atom-JEPA and Chemprop, we run a hyper-parameter tuning setup with Optuna using a budget of 500 iterations.

\subsubsection{Hyperparameter Tuning}
We take the training split of one of the first split in \cite{DBLP:journals/corr/abs-2510-12719} on Biogen split by clusters, which we part into a training and validation split. We select the validation macro MAE as our optimisation target. 
over all targets using Optuna for hyperparameter tuning. For all models except Atom-JEPA and CKERMT, we use a budget of 500 iterations with the TPE optimizer and hyperband pruning. All with a warm start of 15 iterations. For each feature extract(s) + feature processor, we run the full hyperparameter pipeline. If multiple feature extractors are used in the same model, all their hyperparameters are tuned at the same time.
For the LGBM models we fixed the number of estimators high at 2000, as the complexity of the model is already protected by early stopping. The ranges for the LGBM was based on the values from \cite{10.1021/acs.jcim.3c00160} and Optuna's LightGBMTuner, but by design left slightly larger. We use CDF-normalized RDKit features.
\begin{table}[t]
    \centering
    \small
    \setlength{\tabcolsep}{5pt}

    \caption{
        Intervals and priors for the hyperparameters of the specific feature processor tuned over using Optuna.
    }
    \label{tab:processor_hyperparameters}

    \begin{tabular}{lcc}
        \toprule
        \textbf{Processor Hyperparameter} & \textbf{Interval} & \textbf{Prior}\\
        \midrule

        \multicolumn{3}{l}{\textbf{MLP}} \\
        \addlinespace[2pt]
        Learning rate & $(10^{-4}, 10^{-2})$ & Log-uniform \\
        Epochs & $(5, 500)$ & Log-uniform\\
        Weight decay & $(10^{-4}, 10^{-1})$ & Log-uniform \\
        Head dropout & $(0, 0.5)$ & Uniform\\
        Head hidden units & $(32, 256)$ & Log-uniform\\
        Trunk hidden units & $\{128, 256, 512, 1024\}$ & Categorical\\
        Trunk layers & $\{2, 3, 4, 5\}$ & Categorical\\
        Trunk dropout & $(0, 0.5)$ & Uniform \\
        Trunk activations & $\{\text{ReLU, GELU. SiLU}\}$ & Categorical\\
        
        \midrule
        \multicolumn{3}{l}{\textbf{LGBM}} \\
        \addlinespace[2pt]
        Learning rate & $(5\cdot 10^{-3}, 3\cdot 10^{-1})$ & Log-uniform \\
        Number of leaves & $(8,128)$ & Log-uniform \\
        Minimum child sample & $(5,100)$ & Log-uniform \\
        Subsample fraction & $(0.5, 1.)$ & Uniform \\
        Colsample bytree & $(0.4,1)$ & Uniform \\
        Reguralizer $\alpha$ & $(10^{-8}, 10^{1})$ & Log-uniform \\
        Reguralizer $\lambda$ & $(10^{-8}, 10^{1})$ & Log-uniform \\

        \bottomrule
    \end{tabular}%
\end{table}

\begin{table}[t]
    \centering
    \small
    \setlength{\tabcolsep}{5pt}

    \caption{
        Intervals and priors for the hyperparameters of the specific feature extractor tuned over using Optuna.
    }
    \label{tab:extractor_hyperparameters}

    \begin{tabular}{lcc}
        \toprule
        \textbf{Feature Hyperparameter} & \textbf{Interval} & \textbf{Prior}\\
        \midrule

        \multicolumn{3}{l}{\textbf{Morgan}} \\
        \addlinespace[2pt]
        Radius & $\{2, 3\}$ & Categorical \\
        Number of bits & $\{1024, 2048\}$ & Categorical\\
        Use Counts & $(\text{False, True)}$ & Boolean \\
        
        \midrule
        \multicolumn{3}{l}{\textbf{RDKit}} \\
        \addlinespace[2pt]
        -- & -- & -- \\

        \multicolumn{3}{l}{\textbf{ErG}} \\
        \addlinespace[2pt]
        Fuzz Increase & $(0.1, 0.5)$ & Uniform \\

        \multicolumn{3}{l}{\textbf{Avelon}} \\
        \addlinespace[2pt]
        Use Counts & $(\text{False, True)}$ & Boolean \\

        \multicolumn{3}{l}{\textbf{Atom-JEPA}} \\
        \addlinespace[2pt]
        Node Aggregation & $(\text{Mean, Sum)}$ & Categorical \\
        Higher Order Features & $(\text{False, True)}$ & Boolean \\
        Feature Scaling & $(\text{None, Standard, Quantile)}$ & Categorical \\

        \multicolumn{3}{l}{\textbf{Mol-JEPA}} \\
        \addlinespace[2pt]
        Feature Scaling & $(\text{None, Standard, Quantile)}$ & Categorical \\

        \bottomrule
    \end{tabular}%
\end{table}

We allow a higher amount of optimisation iterations for the baseline models based on LightBGM and MLP heads compared to Atom-JEPA due to the difference in computational overhead.
Instead we reuse the hyperparameters found by CKERMT for the architectures. Instead, we perform tuning in two sweeps: first the L2SP weight decay of the encoder $\{0, 10^{-5}, 10^{-4}, 10^{-3}, 10^{-2}, 10^{-1}, 10^{-0}\}$ versus $L_2$ weight decay of the other parameters $\{0, 10^{-5}, 10^{-4}, 10^{-3}, 10^{-2}, 10^{-1}, 10^{-0}\}$, for a total of 49 iterations. Then the learning rate ($\{1\cdot 10^{-5}, 2.5\cdot10^{-5}, 5 \cdot 10^{-5}, 1\cdot10^{-4}, 2.5 \cdot 10^{-4}, 5\cdot 10^{-4}, 1\cdot 10^{-3}\}$), the number of epochs $\{100, 50, 25, 10\}$ and number of epochs the Atom-JEPA encoder is frozen for $\{20, 10, 5, 0\}$ for a total of 98 iterations excluding impossible combinations. We accommodate the much larger data size of ChEMBL-MT by setting the encoder freezing and encoder L2SP to 0.

For the baselines and Atom-JEPA we do not use early stopping on a validation set, instead we train on the full training set as we have already selected the number of epochs.. For the literature models Chemprop, CKERMT, and Mol-JEPA we follow their respective described setups. Using a single set of hyperparameters for all datasets reduces model performance, but significantly reduces the computational burden.

\subsubsection{Conformer Generation Details} \label{appendix:conformer_generation}
For datasets where the 3D structure of the compound is not given, we generate conformers following the same setup as Unimol.
For completeness we detail the setup here. To convert a SMILE string to a 3D conformer, we import the SMILE string in RDKit. If the importing fails with sanitation we retry without sanitation and afterwards sanitise it.

To get the 3D coordinates we use ETKDGv3 to generate up to 10 conformers. We first use attempt to generate with enforced chirality, but in case of failure we start from random coordinates. Since Atom-JEPA uses the SE(3) equivariant GNN EquiformerV3, the model can in theory learn to distinguish between enantiomers. The conformers are only used during training as a form of training augmentation.

For several of the ADMET-TDC datasets, the SMILE string contains multiple fragmented components, such as salt forms of CN(C)C(...)Cc1ccc(Cl)c(Cl)c1.[Cl-].[H+]. This is a problem during conformer relaxation, as the disconnected components can arbitrarily be moved around, thus geometrically ill-defined. 

\begin{table}[t]
\centering
\small
\caption{Molecules in the TDC ADMET benchmark group that RDKit parses as more than one connected component (salts, hydrates, mixtures), over the de-duplicated train\_val + test SMILES of each dataset. ``Extra atoms'' is the mean, over fragmented molecules only, of the share of atoms outside the largest fragment.}
\label{tab:admet-fragmentation}
\begin{tabular}{lrrrr}
\toprule
\textbf{Dataset} & $N$ & \textbf{Fragmented} & \textbf{Fragmented} (\%) & \textbf{Extra atoms} (\%) \\
\midrule
Caco2 Wang & 906 & 9 & 1.0 & 4.6 \\
HIA Hou & 578 & 0 & 0.0 & 0.0 \\
Pgb Broccatelli & 1212 & 0 & 0.0 & 0.0 \\
Bioavailability Ma & 640 & 1 & 0.2 & 52.5 \\
Lipophilicity AstraZeneca & 4200 & 1 & 0.0 & 4.9 \\
Solubility AqSolDB & 9982 & 1098 & 11.0 & 30.2 \\
BBB Martins & 1975 & 105 & 5.3 & 11.7 \\
PPBR AZ & 1797 & 0 & 0.0 & 0.0 \\
VDss Lombardo & 1111 & 0 & 0.0 & 0.0 \\
CYP2D6 Veith & 13130 & 549 & 4.2 & 16.6 \\
CYP3A4 Veith & 12328 & 549 & 4.5 & 16.8 \\
CYP2C9 Veith & 12092 & 538 & 4.4 & 17.0 \\
CYP2D6 Substrate Carbon-Mangels & 664 & 0 & 0.0 & 0.0 \\
CYP3A4 Substrate Carbon-Mangels & 667 & 0 & 0.0 & 0.0 \\
CYP2C9 Substrate Carbon-Mangels & 666 & 0 & 0.0 & 0.0 \\
Half-life Obach & 665 & 10 & 1.5 & 13.2 \\
Clearance Microsome AZ & 1102 & 0 & 0.0 & 0.0 \\
Clearance Mepatocyte AZ & 1020 & 0 & 0.0 & 0.0 \\
hERG & 648 & 12 & 1.9 & 20.8 \\
AMES & 7255 & 0 & 0.0 & 0.0 \\
DILI & 475 & 18 & 3.8 & 30.0 \\
LD50 Zhu & 7342 & 0 & 0.0 & 0.0 \\
\bottomrule
\end{tabular}
\end{table}

\subsubsection{Conformer Scaling} \label{appendix:conformer_influence}
\begin{figure}
    \centering
    \includegraphics[width=0.5\linewidth]{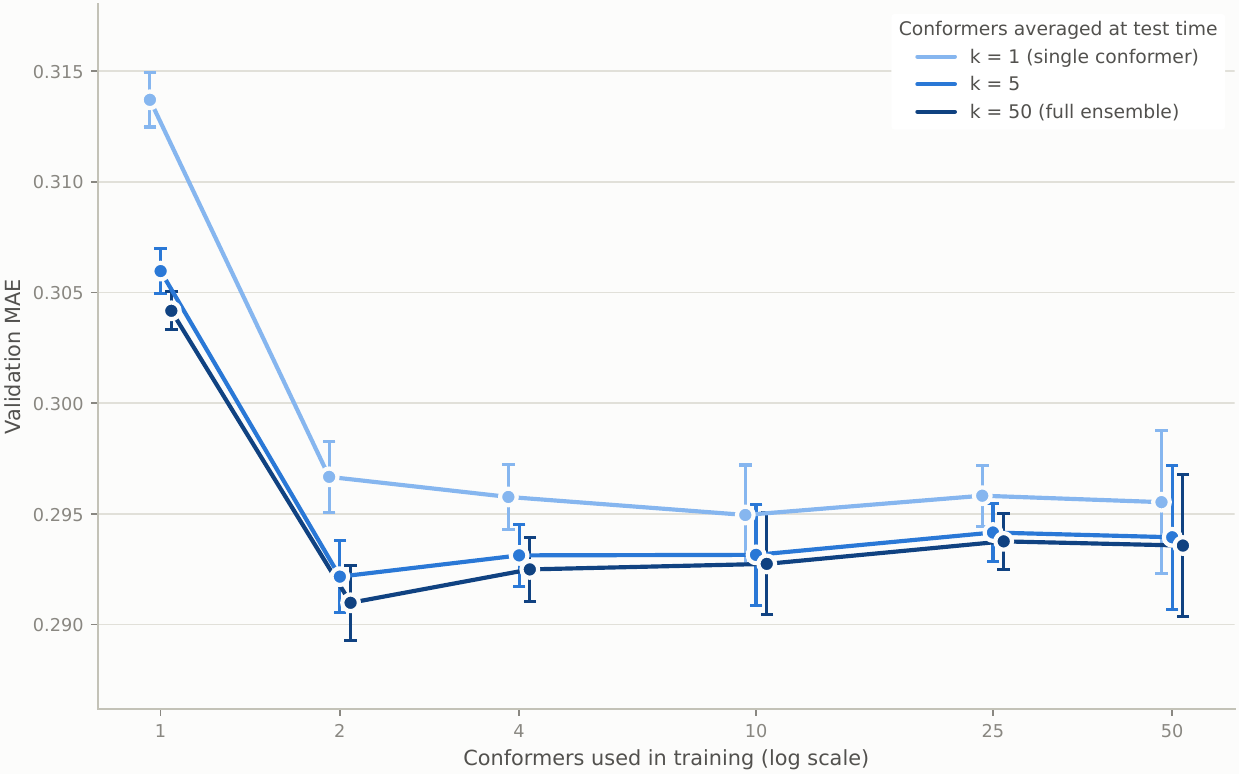}
    \caption{The validation error on the validation error (MAE) against the number of conformers used during training. The different curves represent different numbers of predictions from conformers averaged over during test time (test time augmentations). }
    \label{fig:conformer_training}
\end{figure}

Generation of conformers can be seen as a way of generating multiple 3D views of the same molecule. Previous work~\citep{zhou2023unimol} generates multiple conformers, specifically 11, as a type of augmentation for training. In this section we investigate the impact on performance these parameters have. To avoid leakage of data,  we use the same data for the train and validation split as the hyperparameter tuning. First, we investigate the impact of the number of conformers available during training and test time. Some of the targets have only 7 datapoints in the validation set, so instead of averaging the over the tasks, we present error over the datapoints. For this experiment we generate up to 50 conformers (from 10) which equivalates a new conformer seen each epoch. In addition to using conformers during training, we also investigate the impact of ensembling the predictions over conformers. We average over 10 conformers for each run. The results are shown in Figure \ref{fig:conformer_training}. 
In the figure, we see a large drop in validation MAE when going from training on 1 to 2 conformers, while the run to run variance dominates higher conformer numbers. This shows the importance of having multiple (>1) conformer available during training. Generating many conformers is a good trade-off for many ADMET tasks as cost of generating multiple conformers is comparatively cheap compared to the training time. In addition to training conformers, the figure reveals moderate gains from averaging multiple model predictions from conformers. This can be thought of as a way of performing test-time augmentations or ensemble averaging. Curiously, the figure reveals a weak trend of increasing error when ensembling conformer predictions when training on more conformers, although not statistically significant. We hypothesize, that the variance of the model's predictions over conformers is reduced when trained with more conformers, which could negatively impact the ensembles prediction.

\begin{figure}
    \centering
    \includegraphics[width=0.49\linewidth]{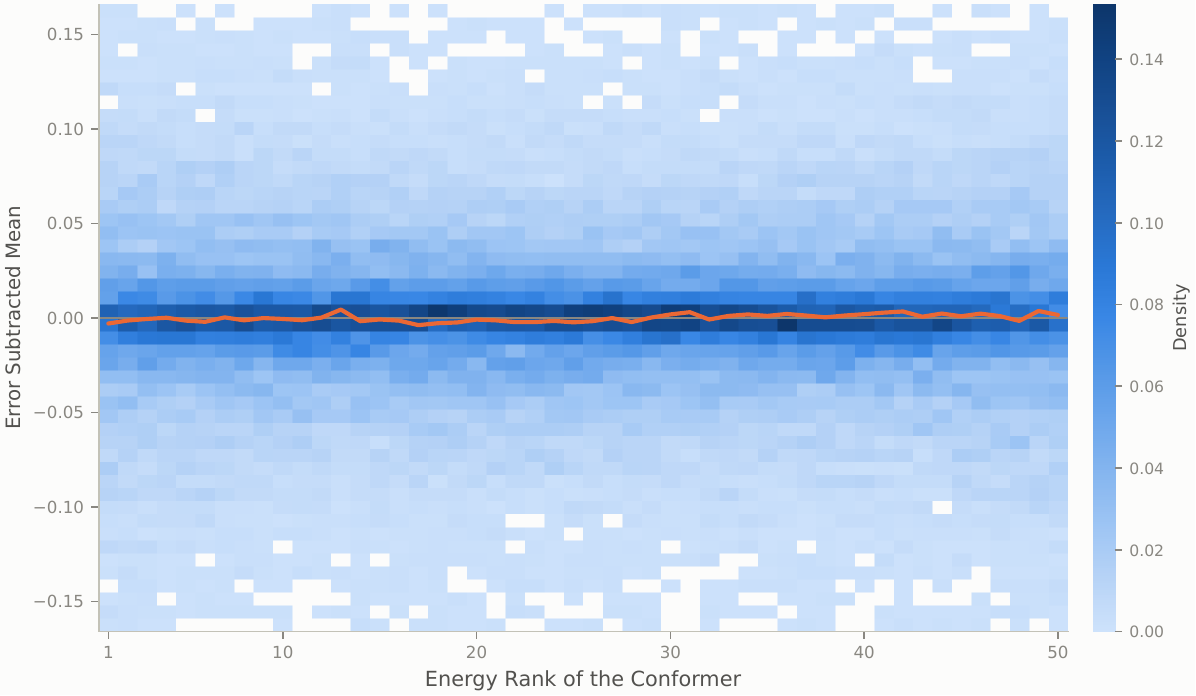}
    \includegraphics[width=0.49\linewidth]{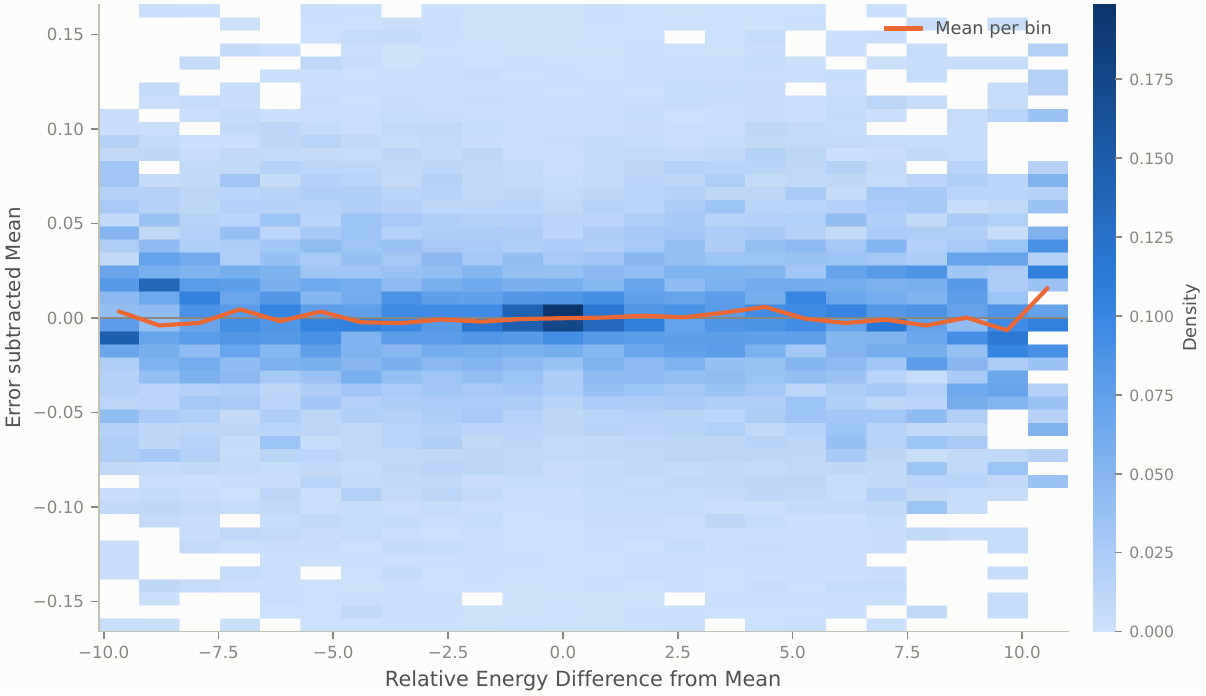}
    \caption{Histogram of the prediction error of validation datapoints versus the energy of the conformer. Left plot is the relative energy of the conformer from the mean, right plot is the rank of the conformer. For none of the plots a strong correlation between error and conformer energy was observed.}
    \label{fig:conformer_energy}
\end{figure}

We also investigate the impact of the conformer generation quality. Lower potential energy of conformers are typically associated with higher quality as they are more probable. To avoid a co-founding generation artifacts from the relaxation process of MMFF94 and using it to judge the same energies, we instead use the interatomic potential model UMA-s-1p2~\citep{wood2026umafamilyuniversalmodels} model to predict energy. The results are shown in Figure \ref{fig:conformer_energy}. Surprisingly, the figures shows no correlation between the error of the conformer prediction versus the conformer quality. Only for conformers of very high energy compared to the mean $(\sim10\%)$ we see some small increase in error.

\subsubsection{ADMET Parameters for Atom-JEPA}
The optimal hyperparameters obtained from hyperparameter tuning can be found in \cref{tab:finetune_hparams}.
\begin{table*}[t]
    \centering
    \small
    \setlength{\tabcolsep}{5pt}
    \renewcommand{\arraystretch}{1.08}

    \caption{
        Hyperparameters used for Atom-JEPA fine-tuning on Biogen ADME, ChEMBL-MT,
     ExpansionRX, and ADMET-TDC. All three start from the same Uni-Mol-pretrained context
        encoder; encoder architecture is unchanged from
        Table~\ref{tab:pretraining_hparams}. Every task
        gets its own MLP head on top of a shared trunk that takes the atom-pooled representation.
    }
    \label{tab:finetune_hparams}

    \begin{tabular}{lccc}
        \toprule
        \textbf{Hyperparameter}
        & \textbf{Biogen ADME}
        & \textbf{ChEMBL-MT}
        & \textbf{ExpansionRX} \\
        \midrule

        \multicolumn{4}{l}{\textbf{Prediction Head}} \\
        \addlinespace[2pt]

        Number of tasks
        & 6 & 25 & 9 \\
 Trunk Architecture& \multicolumn{3}{c}{Linear $(768,256)$ $\rightarrow$ SiLU $\rightarrow$ Dropout}\\
 Per Task Head Architecture& \multicolumn{3}{c}{Linear $(256,64)$ $\rightarrow$ SiLU $\rightarrow$ Linear $(64,1)$}\\

        Shared Trunk dropout& \multicolumn{3}{c}{.1}\\

        Pooling readout
        & \multicolumn{3}{c}{Mean}\\

        Higher-order invariants
        & \multicolumn{3}{c}{Yes}\\

        \midrule
        \multicolumn{4}{l}{\textbf{Conformers}} \\
        \addlinespace[2pt]

        Conformers generated per molecule
        & \multicolumn{3}{c}{10}\\

        Conformers used in training
        & \multicolumn{3}{c}{10}\\

        \midrule
        \multicolumn{4}{l}{\textbf{Objective}} \\
        \addlinespace[2pt]

        Per-task loss
        & \multicolumn{3}{c}{Huber}\\

        Label standardization
        & \multicolumn{3}{c}{\makecell{Per-task train\\mean/std}}\\

        Log-transformed targets
        &         None& None& All but LogD\\

        \midrule
        \multicolumn{4}{l}{\textbf{Regularization}} \\

        L2-SP coefficient $\lambda$
        & $1\times10^{-5}$ &0&$1\times10^{-5}$ \\

        Shared Trunk weight decay
        & \multicolumn{3}{c}{.5 }\\

        MTL log-$\sigma$ weight decay
        & \multicolumn{3}{c}{.1 }\\

        Encoder stochastic depth
        & \multicolumn{3}{c}{$.05$}\\

        Encoder normalization layers
        & \multicolumn{3}{c}{Eval mode }\\

        \midrule
        \multicolumn{4}{l}{\textbf{Optimization}} \\
        \addlinespace[2pt]

        Optimizer
        & \multicolumn{3}{c}{AdamW }\\

        Learning-rate schedule
        & \multicolumn{3}{c}{Cosine + linear warmup}\\

        Encoder learning rate
        &         \multicolumn{3}{c}{$5\times10^{-5}$ }\\

        Head learning-rate multiplier
        & \multicolumn{3}{c}{10 }\\

        MTL log-$\sigma$ learning rate
        & \multicolumn{3}{c}{$5\times10^{-5}$ }\\

        Minimum learning rate
        & \multicolumn{3}{c}{$1\times10^{-7}$ }\\

        Warmup epochs
        & \multicolumn{3}{c}{10 }\\

        Encoder-only warmup epochs
        & \multicolumn{3}{c}{10}\\

        Frozen-encoder epochs
        & 10 &0&10\\

        Batch size
        & \multicolumn{3}{c}{8 }\\

        Number of epochs
        & \multicolumn{3}{c}{50 }\\

        Gradient clipping norm
        & \multicolumn{3}{c}{10 }\\

        Weight EMA decay
        & \multicolumn{3}{c}{.99 }\\

        \bottomrule
    \end{tabular}%
\end{table*}

\newpage
\subsection{QM9 Finetuning}
\label{app:finetune-qm9}
Table~\ref{tab:qm9_finetuning_hparams} summarizes the hyperparameters used for QM9 finetuning. We use three target-specific readouts operating on the encoder's final atom representations following SchNet/SchNetPack-style atom-wise and electronic spatial-extent constructions~\citep{schutt2018schnet} and a PaiNN-style gated equivariant dipole construction~\citep{schütt2021equivariantmessagepassingprediction}, adapted from the GotenNet implementation~\citep{gotennet}. Let $C$ denote the encoder feature width and
$\mathbf{r}_i$ the position of atom $i$. All readouts use sum aggregation
of atom-wise contributions.

\textbf{Dipole moment ($\mu$).}
Two gated equivariant blocks combine the last layer scalar ($\ell=0$) and vector ($\ell=1$) features, with SiLU activations, to predict an atomic scalar $q_i$ and dipole vector $\mathbf{d}_i$. The molecular dipole magnitude is then
\[
    \hat{\mu}
    = \left\|\sum_i
      \left(\mathbf{d}_i + q_i\mathbf{r}_i\right)\right\|_2.
\]

\textbf{Electronic spatial extent ($R^2$).}
A two-layer atomwise MLP with dimensions $C \to C \to 1$ and
shifted-softplus activation predicts scalar weights $a_i$. The
molecular prediction is
\[
    \hat{R}^2
    = \sum_i a_i
      \left\|\mathbf{r}_i-\mathbf{r}_{\mathrm{CM}}\right\|_2^2,
\]
where $\mathbf{r}_{\mathrm{CM}}$ is the molecular center of mass.

\textbf{Remaining targets.}
A two-layer atomwise MLP with dimensions $C \to C \to 1$ and SiLU
activation maps scalar atom features to scalar contributions, which
are summed to obtain the molecular prediction. For $U$, $U_0$, $H$,
and $G$, summed atomic reference energies are subtracted during
training and restored for evaluation, following standard practice.
No targets are standardized.

\begin{table*}[htbp]
    \centering
    \small
    \setlength{\tabcolsep}{5pt}
    \renewcommand{\arraystretch}{1.08}

    \caption{
        Hyperparameters used for QM9 fine-tuning.
        All targets share the same optimization settings, with target-specific
        readout architectures for $\mu$ and $R^2$. Encoder architecture hyperparameters are
        inherited from the pretrained model.
    }
    \label{tab:qm9_finetuning_hparams}

    \resizebox{0.8\textwidth}{!}{%
    \begin{tabular}{lc}
        \toprule
        \textbf{Hyperparameter}
        & \textbf{All targets} \\
        \midrule

        \multicolumn{2}{l}{\textbf{Optimization}} \\
        \addlinespace[2pt]

        Optimizer
        & AdamW \\

        Adam $\epsilon$
        & $10^{-7}$ \\

        Maximum training epochs
        & 1,000 \\

        Batch size
        & 32 \\

        Encoder base learning rate
        & $10^{-4}$ \\

        Readout base learning rate
        & $10^{-4}$ \\

        Weight decay
        & 0 \\

        Learning-rate schedule
        & Linear warmup + ReduceLROnPlateau \\

        Global learning-rate warmup (optimizer steps)
        & 10,000 \\

        Learning-rate reduction factor
        & .8 \\

        Learning-rate reduction patience (epochs)
        & 15 \\

        Minimum scheduler learning rate
        & $10^{-7}$ \\

        Initial encoder freezing (epochs)
        & 10 \\

        Encoder ramp after unfreezing (epochs)
        & 10 \\

        Gradient clipping norm
        & 5 \\

        \midrule
        \multicolumn{2}{l}{\textbf{Objective and Model Selection}} \\
        \addlinespace[2pt]

        Training loss
        & MSE \\

        Scheduler and early-stopping metric
        & Validation MSE \\

        Early-stopping patience (epochs)
        & 150 \\

        Checkpoint-selection metric
        & Validation MAE \\

        \midrule
        \multicolumn{2}{l}{\textbf{Reproducibility}} \\
        \addlinespace[2pt]

        Training seed
        & 42 \\

        Split seed
        & 1 \\

        \bottomrule
    \end{tabular}%
    }
\end{table*}

\subsection{Matbench Finetuning}
\label{app:finetune-matbench}
We use a common training configuration for the five small-to-medium tasks (phonons, dielectric, Log GVRH, Log KVRH and Perovskites), and a second configuration for the three large tasks (MP Gap, MP E Form, and MP Is Metal) as summarized in Table~\ref{tab:matbench_finetuning_hparams}. MP Is Metal is the only classification task. 

All tasks use an atom-wise readout consisting of a two-layer MLP with SiLU activation with one output for regression or two logits for classification. The final layer has a bias only for classification, and no dropout is used. Atom-wise outputs are mean-pooled, except for phonons, which uses max pooling. Classification probabilities are obtained by applying softmax after pooling.

We standardize targets for all regression tasks and not for classification. Training uses separate encoder and readout learning rates, with linear warmup followed by cosine decay. After the initial encoder-freezing phase, the encoder learning rate is ramped over ten epochs for small-to-medium tasks and 1 epoch for large tasks. We do not use L2-SP regularization for Matbench. 

For each official Matbench fold, we reserve 10\% of the training fold for validation using split seed 42 and set the PyTorch random seed to 0. Regression normalization statistics are computed from the training fold. We select EMA weights by the lowest validation MAE for regression or highest validation F1 for classification. Scratch baselines use the same architecture and settings, but initialize the encoder randomly and train it from the first epoch without freezing.

\begin{table*}[htbp]
    \centering
    \small
    \setlength{\tabcolsep}{5pt}
    \renewcommand{\arraystretch}{1.08}

    \caption{
        Hyperparameters used for Matbench fine-tuning.
        Small-to-medium tasks comprise phonons, dielectric, Log GVRH, Log KVRH and Perovskites; large tasks comprise
        MP Gap, MP E Form, and MP Is Metal.
        Classification settings apply only to MP Is Metal.
        Encoder freezing applies only to pretrained runs;
        encoders trained from scratch are trainable from the first epoch.
    }
    \label{tab:matbench_finetuning_hparams}

    \resizebox{\textwidth}{!}{%
    \begin{tabular}{lcc}
        \toprule
        \textbf{Hyperparameter}
        & \textbf{Small-to-medium tasks}
        & \textbf{Large tasks} \\
        \midrule

        \multicolumn{3}{l}{\textbf{Optimization}} \\
        \addlinespace[2pt]

        Optimizer
        & AdamW & AdamW \\

        Adam coefficients $(\beta_1,\beta_2)$
        & $(.9,.95)$ & $(.9,.95)$ \\

        Adam $\epsilon$
        & $10^{-8}$ & $10^{-8}$ \\

        Number of epochs
        & 100 & 12 \\

        Training batch size
        & 4 & 32 \\

        Evaluation batch size
        & 32 & 32 \\

        Encoder base learning rate
        & $5\times10^{-5}$ & $1.5\times10^{-4}$ \\

        Readout learning-rate multiplier
        & 10 & 10 \\

        Readout base learning rate
        & $5\times10^{-4}$ & $1.5\times10^{-3}$ \\

        Learning-rate schedule
        & Linear warmup + cosine decay
        & Linear warmup + cosine decay \\

        Global learning-rate warmup (epochs)
        & 10 & 2 \\

        Initial warmup factor
        & .1 & .5 \\

        Encoder learning-rate floor
        & $10^{-7}$ & $10^{-7}$ \\

        Readout learning-rate floor
        & $10^{-6}$ & $10^{-6}$ \\

        Initial encoder freezing (epochs)
        & 10 & 2 \\

        Encoder ramp after unfreezing (epochs)
        & 10 & 1 \\

        Encoder / readout weight decay
        & $0 / 0$ & $0 / 0$ \\

        Gradient clipping norm
        & 10 & 10 \\

        Model EMA decay
        & .99 & .99 \\

        \midrule
        \multicolumn{3}{l}{\textbf{Objectives and Aggregation}} \\
        \addlinespace[2pt]

        Regression training loss
        & MAE
        & Huber \\

        Classification training loss
        & ---
        & Weighted cross-entropy \\

        Aggregation of atomwise outputs
        & \makecell{Mean\\max for phonons}
        & Mean \\
        \bottomrule
    \end{tabular}%
    }
\end{table*}

\section{Tukey HSD Plots}
\subsection{Biogen}
\begin{figure}[H]
    \centering
    \includegraphics[width=0.99\linewidth]{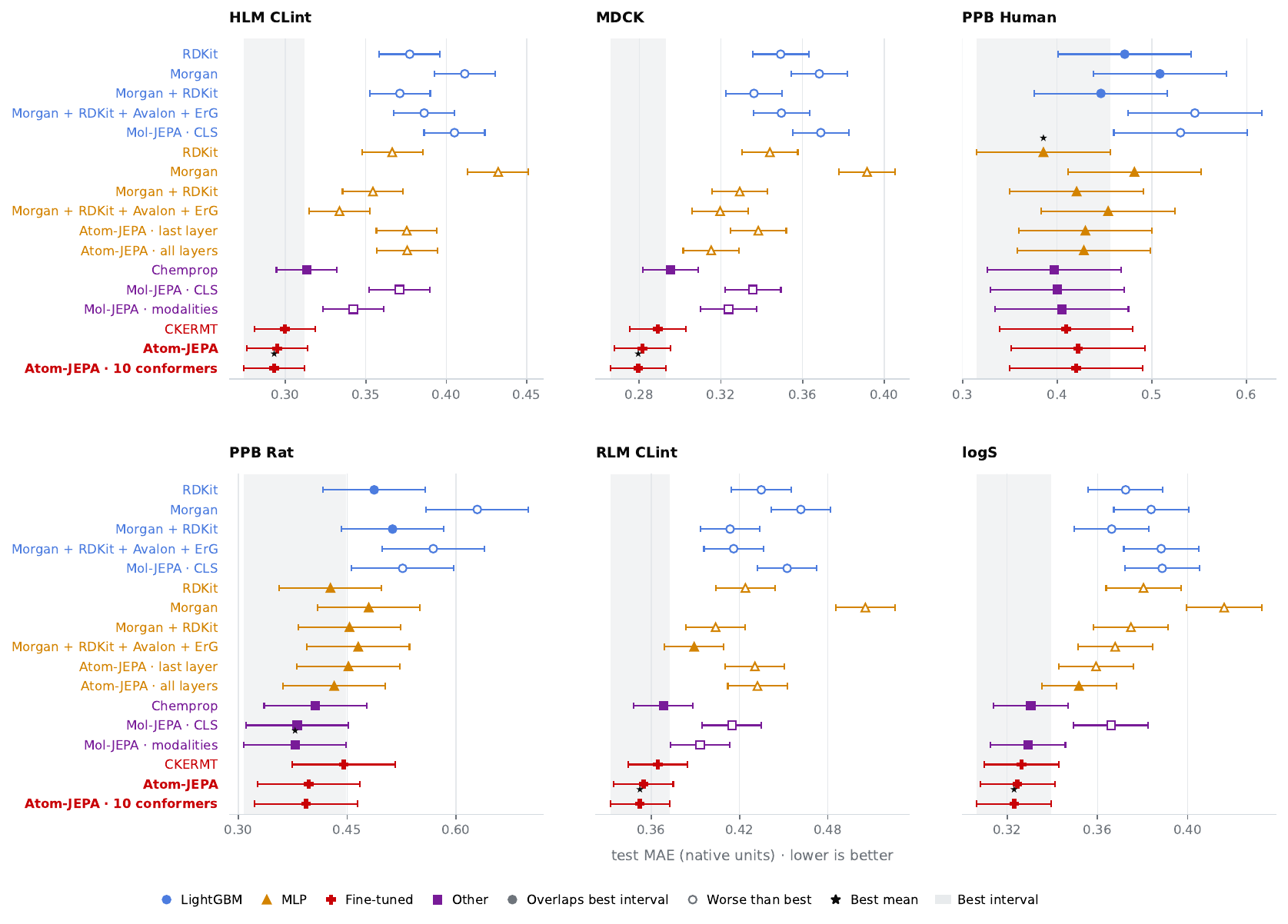}
    \caption{Tukey HSD grid plot across the endpoints of Biogen Scaffold test data over 5 seeds. The analysis was conducted with a blocked ANOVA test followed by Tukey HSD using the Statsmodels Python package. }
    \label{fig:tukey_biogen}
\end{figure}
\newpage

\subsection{ExpansionRX}
\begin{figure}[H]
    \centering
    \includegraphics[width=0.99\linewidth]{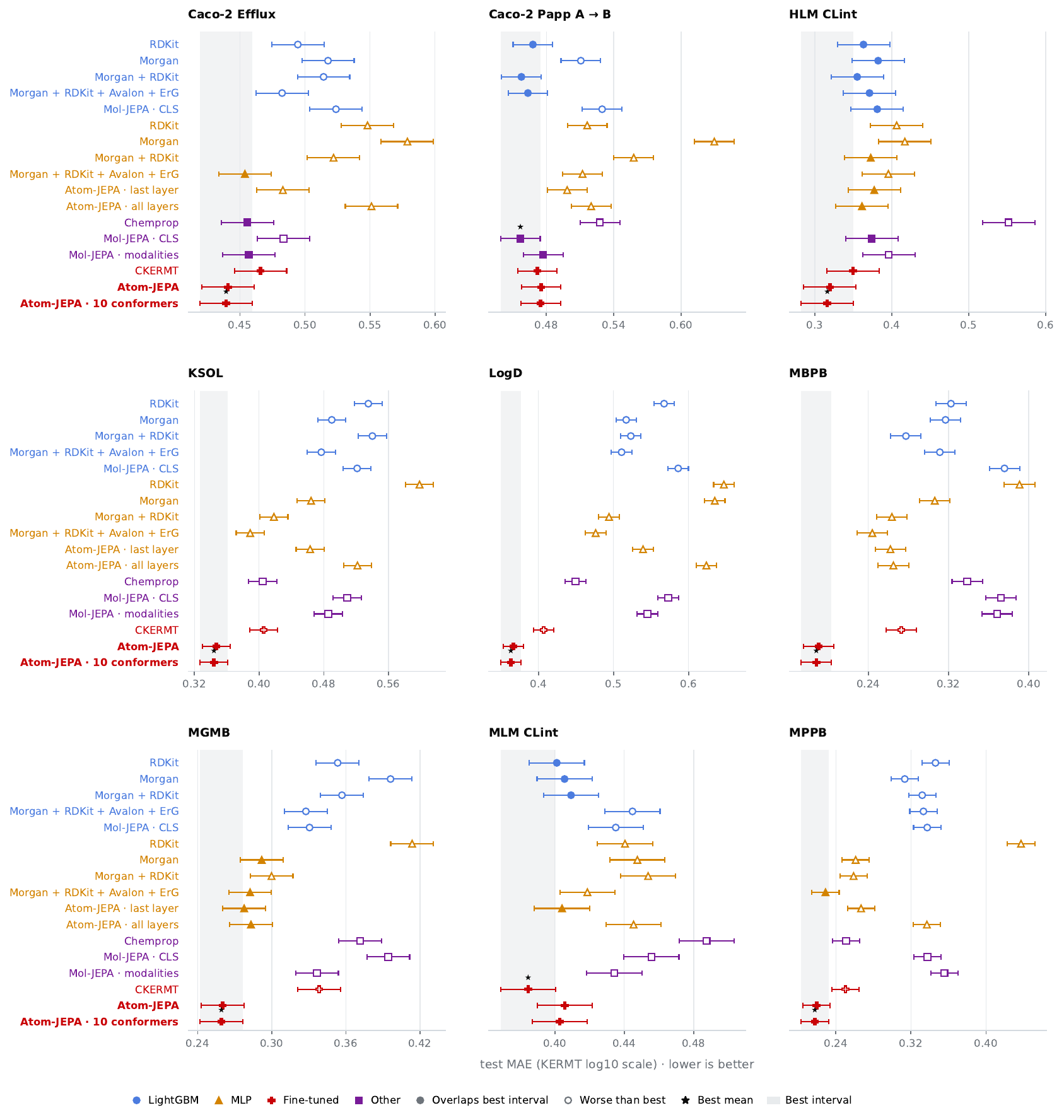}
    \caption{Tukey HSD grid plot across the endpoints of ExpansionRX temporal test data over 4 seeds. The analysis was conducted with a one-way ANOVA test followed by Tukey HSD using the Statsmodels Python package.}
    \label{fig:tukey_expansionrx}
\end{figure}
\newpage

\subsection{ChEMBL-MT}
\begin{figure}[H]
    \centering
    \includegraphics[width=0.99\linewidth]{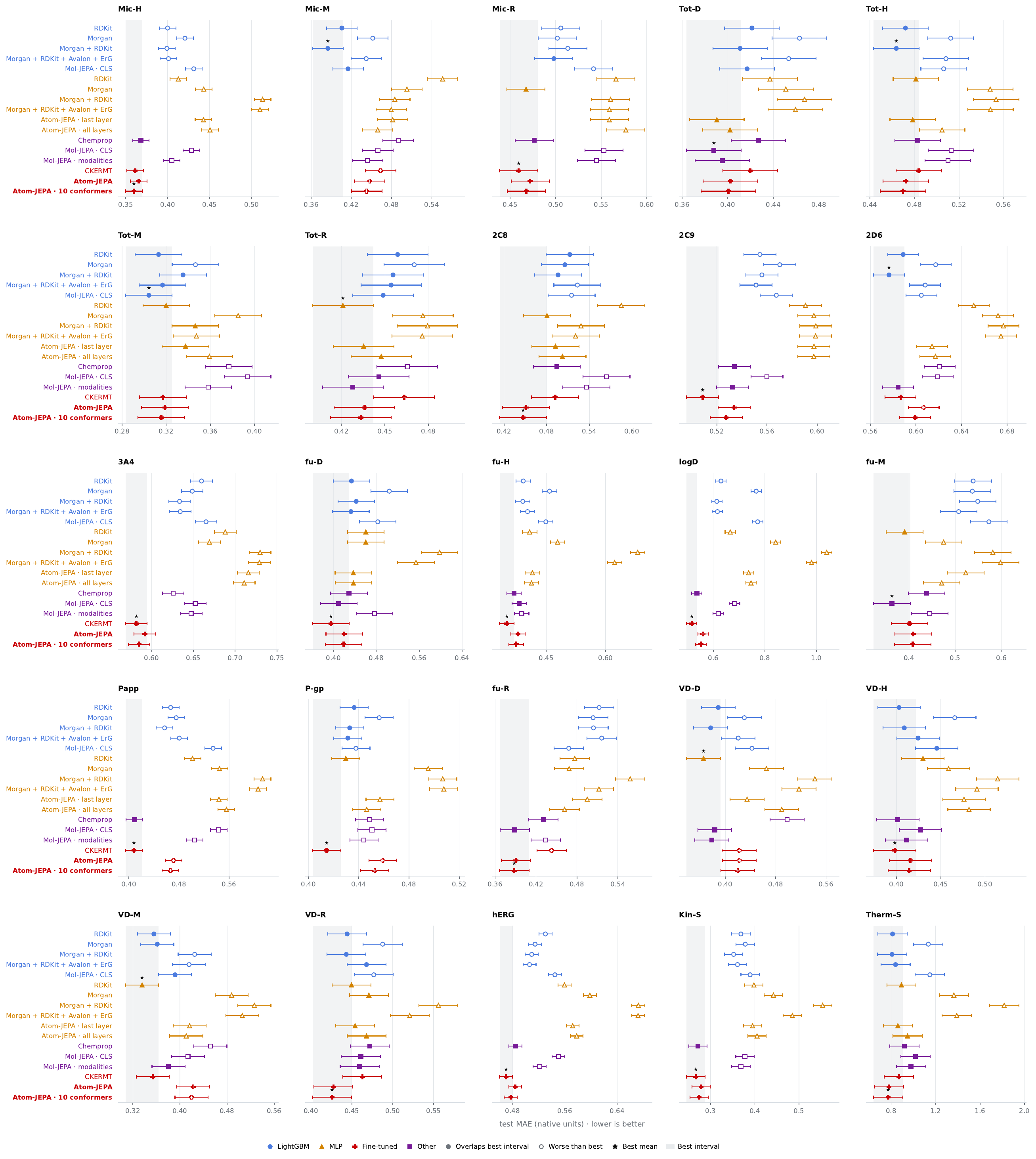}
    \caption{Tukey HSD grid plot across the endpoints of ChEMBL-MT test data over 4 seeds. The analysis was conducted with a blocked ANOVA test followed by Tukey HSD using the Statsmodels Python package.}
    \label{fig:tukey_chembl}
\end{figure}

\end{document}